\documentclass[a4paper,12pt,oneside,times,print,index]{Classes/PhDThesisPSnPDF}

\input{Preamble/preamble}
\title{Real-Time Monocular Scene Analysis Using Joint Deep Learning for UAV  \texorpdfstring{\\ \LaTeX2e}{LaTeX2e}}

\subtitle{Using the Latex thesis template}

\author{Yara AlaaEldin Abdelmottaleb}

\dept{DIBRIS}

\university{University of Genova}

\degreetitle{Doctor of Philosophy}

\subject{LaTeX} \keywords{{LaTeX} {PhD Thesis} {Robotics and Intelligent Machines} {University of Genova}}

\ifdefineAbstract
\fi

\ifdefineChapter
\fi

\begin{document}


\thispagestyle{empty}
\begin{figure}[h!]
 \centering
 \includegraphics[scale=0.2]{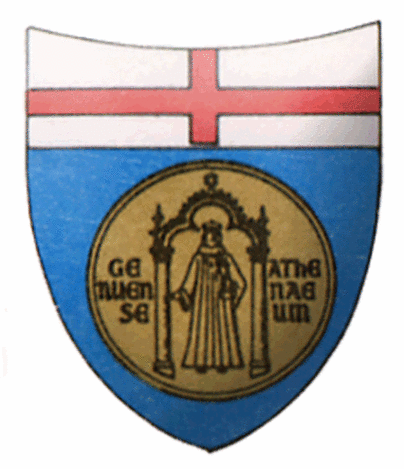} 
 		  \vspace{-0.5em}
	\begin{center} 
		\large {
		{\text{University of Genova}}}\\
		  \vspace{2em}
		  {\huge \text{DRIM}}\\
               \vspace{1.5em}
            \large {\text{Ph.D. Program of National Interest in Robotics and Intelligent Machines}\\
               \vspace{0.5em}
       		{\text{Administrative Headquarters: Università di Genova}}}\\

	\end{center}
\end{figure}

\begin{center} 
	

		\LARGE
		\textbf{Real-Time Monocular Scene Analysis for UAV in Outdoor Environments} \\
\end{center}

 	\begin{center} 
		by \\
		\vspace{0.5em}
		\textbf{Yara AlaaEldin Abdelmottaleb}\\
		\vspace{1em}

	\vspace{0.5cm}	
		\normalsize
		Thesis submitted for the degree of \textit{Doctor of Philosophy} ($38^\circ$ cycle) \\
		\normalsize
		December 2025\\ 
	\end{center}
	\vspace{1.5em}


	\noindent {Francesca Odone} \hfill  {Supervisor}	\\
	\noindent {Antonio Sgorbissa}	\hfill  {Head of the PhD program}	
	\vspace{1em} \\
		\noindent {\textbf{\textit{Thesis Jury:}}}			
	\\
	\noindent {Raffaella Lanzarotti, \textit{Università degli Studi di Milano}}	\hfill  {External examiner}	
	\\
	\noindent {Alessandra Sciutti, \textit{IIT}}	\hfill  {External examiner}
	\\
			\noindent {Nicoletta Noceti, \textit{Università degli Studi di Genova}} \hfill  {Internal examiner}

\begin{figure}[h!]
 \centering
 \includegraphics[scale=0.5]{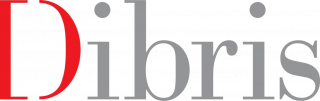}
\begin{center} 
		\small{Dibris\\}
	\end{center}
\end{figure}


\frontmatter



\begin{dedication} 

To all the children of Gaza.
\end{dedication}


\begin{declaration}

I hereby declare that except where specific reference is made to the work of 
others, the contents of this dissertation are original and have not been 
submitted in whole or in part for consideration for any other degree or 
qualification in this, or any other university. This dissertation is my own 
work and contains nothing which is the outcome of work done in collaboration 
with others, except as specified in the text and Acknowledgements. This 
dissertation contains fewer than 65,000 words including appendices, 
bibliography, footnotes, tables and equations and has fewer than 150 figures.


\end{declaration}


\begin{acknowledgements}

In the name of Allah, the merciful.
I would like to thank my parents for their continuous support during my study. I would like to give a special thanking to my supervisor prof. Odone for her support, guidance, and advising throughout the three years of my PhD. I would like also to thank the PhDRIM board for all their efforts to assist us in all the steps from the beginning to the end. I thank all my family members and my friends in Egypt and in Italy for their encouragements and standing by my side throughout the journey. In the end, I thank Allah for giving me the health and well-being necessary to complete this thesis.

\end{acknowledgements}

\begin{abstract}
Understanding the geometric and semantic properties of the scene is crucial in autonomous navigation and particularly challenging in the case of Unmanned Aerial Vehicle (UAV). Such information may be obtained by estimating depth and semantic segmentation maps of the surrounding environment, and for practicality, the procedure must be performed as fast as possible. In this thesis, we leverage monocular cameras on aerial robots to predict depth and semantic maps in low-altitude unstructured environments. We propose a joint deep-learning architecture, named Co-SemDepth, that can perform the two tasks accurately and rapidly, and validate its effectiveness on a variety of datasets.

The training of neural networks requires an abundance of annotated data, and in the UAV field, the availability of such data is limited due to the specificity of the domain and the burden of the annotation process. Simulation engines allow us to collect annotated data automatically with minimal effort. 
We introduce a new synthetic dataset in this thesis, \textit{TopAir}\footnote{Dataset is publicly available:\url{https://huggingface.co/datasets/yaraalaa0/TopAir}} that contains images captured with a nadir view in outdoor environments at different altitudes, helping to fill the gap of the scarcity of annotated datasets in the aerial field. 

While using synthetic data for the training is convenient, it raises issues when shifting to the real domain for testing. We conduct an extensive analytical study to assess the effect of several factors 
on the synthetic-to-real generalization in depth estimation and semantic segmentation. Co-SemDepth and TaskPrompter models are used for comparison in this study. 

The results reveal a superior generalization performance for Co-SemDepth in depth estimation and for TaskPrompter in semantic segmentation. Also, our analysis allows us to determine which training datasets lead to a better generalization for depth estimation and semantic segmentation.
Using few-shot learning generally improved the generalization outcomes, and a visualization of the 3D semantic maps using the predictions is presented.

Moreover, to help attenuate the gap between the synthetic and real domains, image style transfer techniques are explored on aerial images to convert from the synthetic style to the realistic style. Cycle-GAN and Diffusion models are employed. The results reveal that diffusion models are better in the synthetic to real style transfer.

In the end, we focus on the marine domain and address its challenges. Co-SemDepth is trained on a collected synthetic marine data, called \textit{MidSea}, and tested on both synthetic and real data. In addition, self-supervised approaches are tried to enhance the results and cope with the limited available annotated data. The results reveal good generalization performance of Co-SemDepth trained from scratch when tested on real data from the SMD dataset, which contains simple marine scenarios, while further enhancement is needed on the MIT Sea Grant dataset, which contains more challenging scenarios.

\end{abstract}


\tableofcontents

\listoffigures

\listoftables


\printnomenclature

\mainmatter
\chapter{Introduction}  

\ifpdf
    \graphicspath{{Chapter1/Figures/Raster/}{Chapter1/Figures/PDF/}{Chapter1/Figures/}}
\else
    \graphicspath{{Chapter1/Figures/Vector/}{Chapter1/Figures/}}
\fi


\section{Scene Analysis for UAV}
The applications of unmanned aerial vehicles, also known as UAVs, are rapidly expanding across various fields, including environmental exploration, national security, package delivery, firefighting, and many more. The tasks included 
vary depending on the mission. 
Such tasks can include object detection, tracking, classification, depth estimation, and semantic segmentation. 
We focus on depth estimation and semantic segmentation. As it commonly happens in autonomous navigation, sensors are adopted to estimate scene depth and semantic information. Unlike ground autonomous vehicles, many types of UAVs, including drones, have limited computational capability and allowed carried weight. Thus, not all types of sensors can be mounted on them. For instance, sensors like LiDAR and RADAR that are usually adopted for estimating the depth of near or far objects cannot be adopted for lightweight UAVs as they are heavy and power-consuming. Also, while LiDAR point clouds contain accurate depth information, it is difficult to extract rich semantic information from them. 
To associate such depth points to their semantic meaning, an additional step of calibration between LiDAR and RGB cameras has to be done to estimate their relative transformation~\cite{velas2014calibration} and associate the points in the point cloud to their corresponding pixels in the image frame. However, such calibration is never fully accurate, and this leads to errors in the semantic association.
Other depth sensors like stereo cameras are not common and may not be appropriate for UAVs since the small baseline distance between their two internal cameras, compared to the large distance between the stereo camera and the scene, produces inaccurate depth estimates~\cite{olson2010wide}. 
Therefore, there is a particular need to achieve depth estimation for UAV applications using only monocular cameras, as they are cheap, light, and small in size. Current state-of-the-art methods use deep learning architectures for Monocular Depth Estimation (MDE)~\cite{manydepth, bhat2021adabins, monodepth}. 


%

Extracting semantic information from the perceived scene can be achieved using multiple computer vision techniques: object detection, object classification, and semantic segmentation. In object detection and classification the goal is to predict the class of the objects appearing in the image and their corresponding bounding boxes in 2D pixel space or in 3D space. In semantic segmentation the goal is to assign a class label to every pixel in the image belonging to a predefined set of object categories.
There are no known sensors that can achieve semantic segmentation directly. Instead, this task has to be realized by the semantic analysis of the RGB frames received from monocular video cameras. Current state-of-the-art methods~\cite{zhao2018icnet, romera2017erfnet, xie2021segformer, yu2018bisenet, wang2018understanding} are using deep neural networks to learn the semantic features of the input image and decode the required output. 

\begin{figure*}%
    \centering
    {\includegraphics[width=0.8\linewidth]{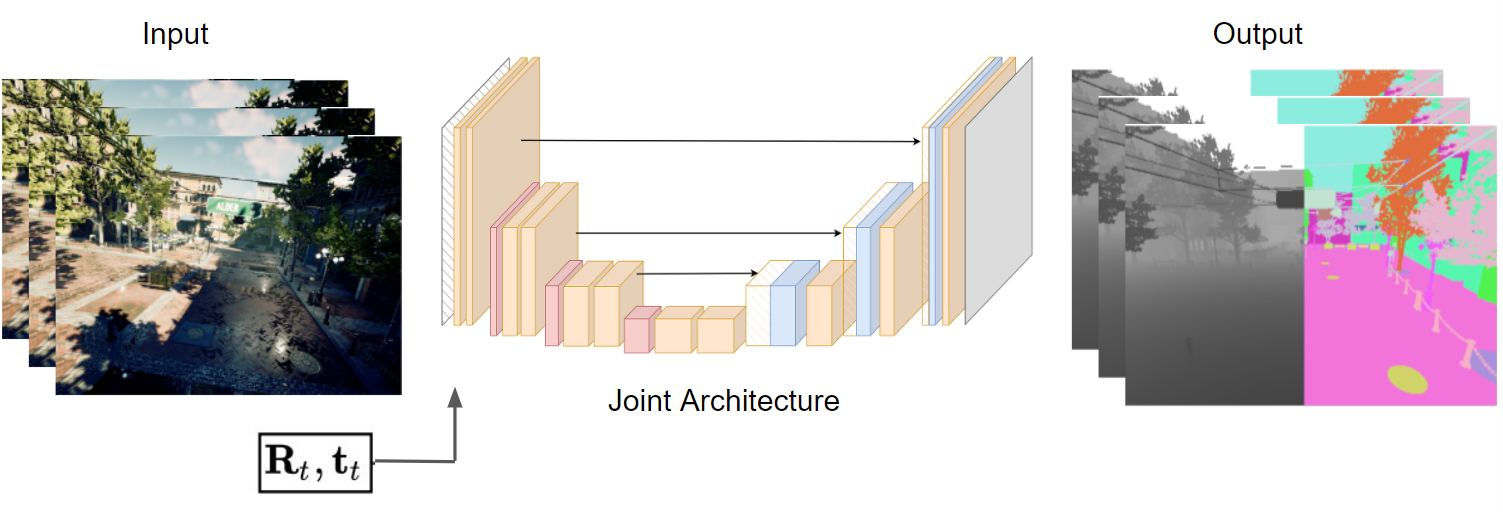} }
    \caption{\centering Our objective of designing a joint architecture for estimating both depth and semantic segmentation maps given as input video frames and camera pose data.}
    \label{fig:joint2}%
\end{figure*}


In this thesis, we leverage monocular cameras on aerial vehicles for obtaining two types of necessary information for scene understanding, {\em depth estimation} and {\em semantic segmentation}:
\begin{itemize}
    \item  In monocular depth estimation (MDE), the goal is to predict the depth of each pixel in each RGB frame captured by a video camera. Such depth expresses the distance (in meters) of the points in the world, projected in the pixels, with respect to the camera frame. 
    \item In semantic segmentation, the goal, instead, is to predict the semantic class of each pixel in the input RGB frames. This semantic class belongs to a set of predefined semantic classes of interest.
\end{itemize}

The two are complementary since depth estimation expresses the geometric properties of the scene while semantic segmentation expresses the semantic properties.

To address both challenges at once, we propose a joint deep architecture based on U-Net for achieving the two tasks accurately and in real-time, see Figure ~\ref{fig:joint2} for a broad overview. 
Using a {\em joint architecture} helps in saving computational time compared to performing each task separately, as well as saving GPU memory by having fewer model parameters. 
Also, joint-learning can help in sharing learned features between the two tasks, and this can, in turn, benefit both of them. 

As for the architecture design, while currently there is a trend in using foundation models for achieving these tasks (see, for instance, DepthAnything~\cite{depth_anything_v2}, SegmentAnything~\cite{segmentanything} and Any-to-Any~\cite{anytoany}), such models are more concerned with zero-shot prediction which is not our primary goal, and they are huge in size and not convenient for hardware deployment where memory and speed are critical factors. With current consumer hardware we can rely on at most 16 GB of primary memory and 16 GB of storage. To show these limitations, Table~\ref{table_specs} reports the memory and power specifications of commonly used microcontrollers and single-board computers (SBCs) in robotic hardware.

\begin{table}[h!]
\begin{center}
\caption{\centering Specifications of commonly used micro-controllers and single board computers in robotics}
\label{table_specs}
\resizebox{0.6\linewidth}{!}{
\begin{tabular}{ c | c | c | c | c }
\hline
 Board  & GPU? & Memory & Storage & Power \\
\hline
Arduino Uno~\cite{arduino_uno} & No & 32KB & - & 5V\\
 Arduino Portenta~\cite{arduino_port} & No & 2MB & 16MB & 5V\\
 Rapberry Pi~\cite{raspberry}  & No & 8GB & 8GB & 15W\\
 Jetson Nano~\cite{jetson}  & Yes & 4GB & 16GB & 5W-10W\\
 Jetson TX2~\cite{jetson}  & Yes & 4/8GB & 16/32GB &  7.5W-15W\\
Jetson Xavier NX~\cite{jetson} & Yes & 8/16GB & 16GB & 10W-20W\\
Jetson AGX Xavier~\cite{jetson} & Yes & 32/64GB & 32GB & 10W-30W\\
\hline
\end{tabular}

}
\end{center}
\end{table}

\section{Collection of Synthetic Datasets}
For the training of neural networks, different types of training are possible: supervised~\cite{monodepth2, supervised}, unsupervised~\cite{huang2019unsupervised, unsupervised}, or semi-supervised~\cite{semisupervised, semisupervised2}. Supervised training learns from input data (in our case, RGB images) and their labels. The goal of the network is to learn the mapping function from the input data to their corresponding labels to be able to predict the correct output for unseen input. Unsupervised training learns from unlabelled data. The network tries in this case to learn the pattern and the features from the data without any labels. 
Semi-supervised learning benefits from both labelled and unlabelled data. 
We hereby focus our attention on supervised learning to train our proposed architecture.

The training of a supervised deep network requires an abundance of data that should contain RGB frames along with their corresponding annotation, in our case, both depth maps and semantic segmentation maps. In general, and especially in the UAV field, annotated real datasets are limited and small in size due to the huge effort required for carrying out the annotation process. For example, labeling a single semantic urban image in Cityscapes dataset~\cite{cordts2016cityscapes} can take up to 60 minutes~\cite{diffumask}. In addition, not all types and variants of the outdoor environments are represented in the currently available real datasets. As a consequence, it is convenient to use simulation engines like AirSim~\cite{airsim} and CARLA~\cite{carla} for the collection of automatically annotated synthetic datasets in a variety of environments, and changing daytime and weather conditions to compensate for the lack in the available real datasets. Such synthetic data can be collected with large amounts and almost zero effort in the annotation process, and then, this data can be used for the training of supervised networks. 
We collect a synthetic aerial dataset \textit{TopAir} using the AirSim simulator that is captured from a nadir (top) view in different environments and at various altitudes. Our dataset is annotated with depth and semantic segmentation maps, and it contains camera pose information. 

\section{Synthetic-to-Real}
\begin{figure}%
    \centering
    \begin{subfigure}[b]{\textwidth}
  \centering
  \includegraphics[width=0.7\linewidth, height = 5.4cm]{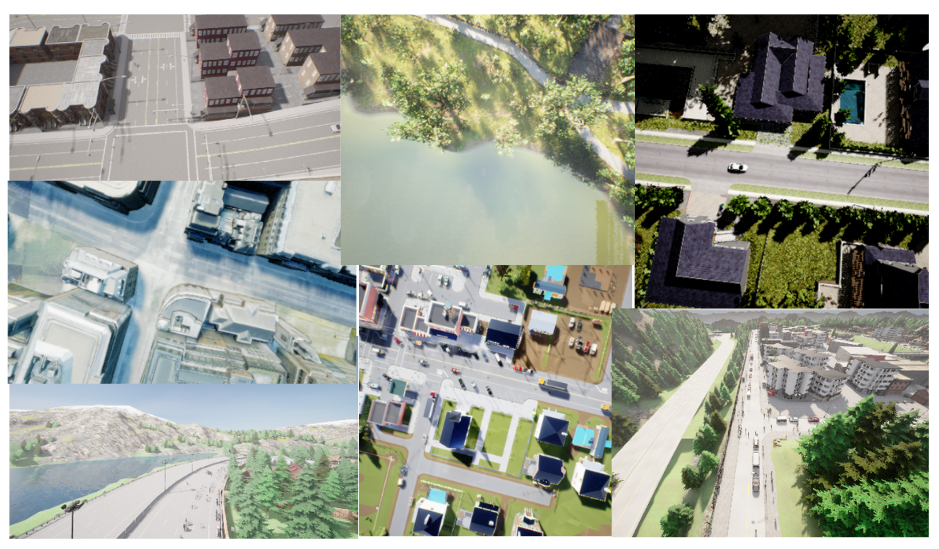}
  \caption{Synthetic images}
  \label{fig:sample_synth}
\end{subfigure}

\begin{subfigure}[b]{\textwidth}
  \centering
  \includegraphics[width=0.7\linewidth, height = 5.4cm]{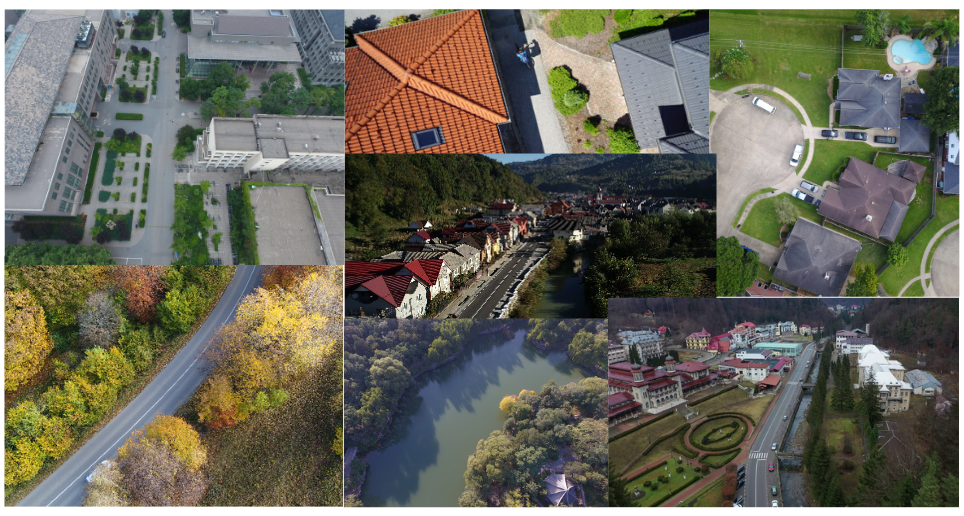}
  \caption{Real images}
  \label{fig:sample_real}
\end{subfigure}
    \caption{\centering Sample images captured in simulation (a) and real-world (b) outdoor environments. It can be observed the differences in color, texture, lighting, and semantic detail between the two domains.}%
    \label{fig:synth_real_diff}%
\end{figure}

While the usage of synthetic datasets for training neural networks is convenient, this brings to the surface the issue of \textit{synthetic-to-real domain generalization} where the neural network is trained on only or mostly synthetic data, and it is expected to perform well when tested on real data. The difference in the appearance of objects between the synthetic and the real world generally causes a drop in the neural network performance on real data. 
For example, in Figure~\ref{fig:synth_real_diff}, some of the variations between synthetic and real-world images are demonstrated. While there is a recent advancement in the 3D graphics of simulation engines, it can be observed that the lighting, texture, and semantic detail of the synthetic scenes still lag behind the real world. 
Most of the works in the literature addressing this problem were developed in the automotive field due to the latest trend in autonomous driving~\cite{loiseau2024reliability, xiao2022transfer, zheng2018t2net, chen2019learning}, leaving the aerial field under-explored~\cite{skyscenes, ruralsynth}. Motivated by this, we investigate the problem of synthetic-to-real domain generalization in UAV monocular depth estimation and semantic segmentation by conducting an extensive set of experiments adopting a variety of synthetic and real aerial datasets that are publicly available and using joint deep architectures. 

In particular, we evaluate the following factors:
\begin{itemize}
    \item How changing the synthetic dataset used for training changes the model performance on the real data.
    \item How changing the model architecture can affect the synthetic-to-real generalization performance.
    \item Whether adding a small number of real data to the training would enhance the synthetic-to-real generalization output.
\end{itemize}

The analysis we carry out involves our joint architecture Co-SemDepth (a light, small network, more convenient for hardware deployment) as well as TaskPrompter (a big transformer-based network, more suitable for offline testing)~\cite{j8}. 
This allows us to assess the impact of architectural properties in the reported analysis. In addition, we try out some of the image style transfer techniques to attenuate the gap between the synthetic and real images.



\section{Application to Marine Environments}

An additional objective of our thesis is to apply the developed pipeline to hazardous outdoor scenarios; we consider, in particular, the marine environment. Such an environment is specifically addressed due to the additional challenges related to the sea (like water reflections and waves) and the limited datasets available in the maritime domain. There are many applications for Unmanned Surface Vehicles (USV) in the sea, including rescue, border surveillance, environmental monitoring, as well as maintenance of offshore systems~\cite{marapp}. 

First, synthetic marine data was collected using UnrealEngine5~\cite{unrealengine}, and we call the resulting dataset \textit{MidSea}. Then, we apply the proposed Co-SemDepth architecture on MidSea and analyze the results obtained both quantitatively and qualitatively. In addition, we evaluate the synthetic-to-real performance by training the model on synthetic data and testing it on real, publicly available data. The analysis allows us to discuss the specifics of the marine environments and the potential and limits of our proposed solutions.

\section{Problem Statement \& Assumptions}

In this section, the specifications of the problem, system assumptions, and the objectives are summarized.

\textbf{Problem Statement:} We are considering the problem of a UAV flying in outdoor unstructured environments (for instance, desert, forest, wild nature, and sea) and performing scene analysis of the video frames received from a monocular camera mounted on it. 
We focus our attention on depth estimation and semantic segmentation with the aim of carrying out these tasks in real-time. In most of the applications, the scene analysis has to be carried out in real-time due to the criticality of instantaneous decisions in UAVs. 
To address the scene analysis using deep neural networks, a sufficient amount of data has to be available for the training of the networks. 
However, a limited amount of data is available in the aerial field, and it is hard to find annotated real datasets. 
By using simulation engines, a huge amount of data can be collected and automatically and precisely annotated. Nevertheless, the synthetic-to-real domain gap problem remains to be addressed. The content, colors and style of collected synthetic images are different from real ones. We try to address this gap by varying the data used for training the network and by varying the architecture of the network. 
Also, image style transfer techniques are explored.
One type of outdoor environments, the marine, has very limited data (synthetic and real), and it has challenges like transparent water, estimating the shoreline, and waves and sea dynamics that need attention. For this reason, we dedicate a separate chapter for it.

\begin{figure}%
    \centering
    {\includegraphics[width=9cm, height=5cm]{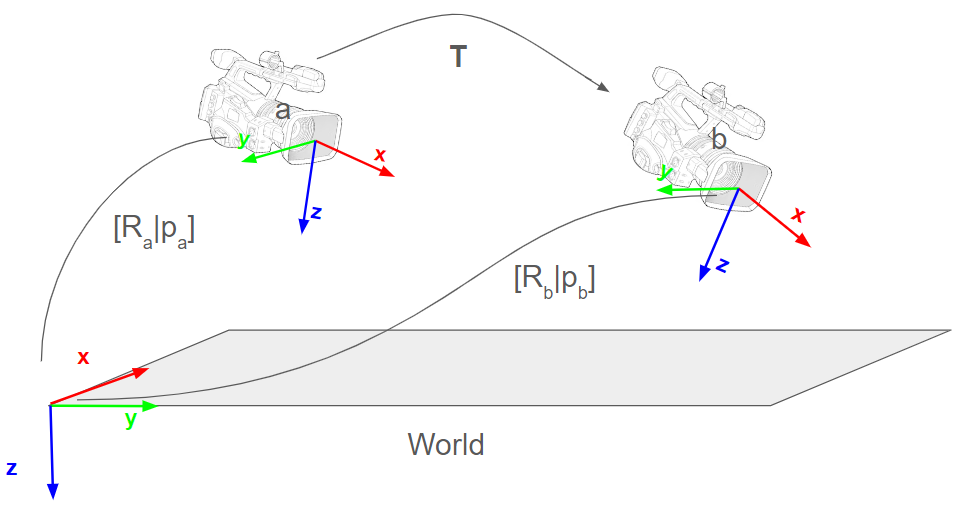} }
    \caption{\centering Illustration of the camera moving from one frame $a$ to the next $b$}%
    \label{fig:camera_transf}%
\end{figure}

\textbf{System Assumptions:} In this thesis, we are considering a monocular camera is rigidly attached to a UAV. The camera intrinsic parameters are assumed to be known and constant. The intrinsic parameters describe the internal geometrical and optical characteristics of a camera, specifically used for mapping 3D world points to 2D image pixels. They include focal length $f$, the principal point $c$, and the sensor skew. The camera frame rate is relatively high, resulting in overlapping regions between consecutive frames. The UAV moves freely (6 Degrees of Freedom) in space and records the video frames as well as the camera position (using an IMU sensor for example) at each time step. As illustrated in Figure~\ref{fig:camera_transf}, using the camera position \(p = [x, y, z]\) and orientation \(R = [q_w, q_x, q_y, q_z]\) at each time step we can compute the motion transformation matrix \({}^{a}T_{b}^{}\) from one frame \(a\) to the next \(b\), as follows:  
\begin{equation}\label{eq:trans}
t = R_{a}^{-1}(p_b - p_a)
\end{equation}
\begin{equation}\label{eq:rot}
r_{3\times3} = R_{a}^{-1}R_b
\end{equation}
\begin{equation}\label{eq:transf}
{}^{a}T_{b}^{} = \begin{bmatrix}
r & t\\
0 & 1
\end{bmatrix}
\end{equation}

\textbf{Objectives:}
Our first objective is to design a network, denoted by a function \(F\), that takes at each time step the current frame \(I_{t}\), previous \(n\) frames \(I_{seq} = [I_{t-1}, I_{t-2}, ..., I_{t-n}]\) and camera motion transformations \(T_{seq} = [T_{t-1}, T_{t-2}, ..., T_{t-n}]\) and outputs an estimated depth map \(\hat{D_t}\) and semantic segmentation map \(\hat{S_t}\) corresponding to the current frame:

\begin{equation}\label{eq:fn}
    (\hat{D_t}, \hat{S_t}) = F(I_t, I_{seq}, T_{seq}).
\end{equation}

Our second objective is to collect synthetic annotated data from a nadir view due to the scarcity of this type of data in the aerial field. 

Our third objective is to train the network, developed in the first stage, on only or mostly synthetic data \(Data_{synth}\) and make it generalize well to unseen real data \(Data_{real}\) at test time. 

The final objective is to focus on the marine environments by applying the developed pipeline on both synthetic and real marine data.

\section{Contributions}


In summary, the main contributions of this work are the following:
\begin{itemize}
    \item We propose a lightweight joint architecture, Co-SemDepth, for performing monocular depth estimation and semantic segmentation on aerial data~\cite{cosemdepth}.
    \item We provide a benchmark for our method and other state-of-the-art methods in semantic segmentation and depth estimation, and highlight the advantages of using our joint architecture with regard to speed and memory efficiency.
    \item We introduce \textit{TopAir}: an aerial synthetic dataset collected with a nadir view in various outdoor unstructured environments with annotations of depth, semantic segmentation, and camera location.
    \item We conduct a comparative analysis of the synthetic-to-real generalization between different synthetic data used for training and using different architectures~\cite{synrealpaper}.
    \item We demonstrate the positive effect of adding a small percentage of real data (few-shot learning) to the training of the networks on their synthetic-to-real generalization.
    \item We explore image style transfer techniques (Cycle-GAN and Diffusion models) to convert from the synthetic to the realistic style of input images
    \item We conduct an analysis on the application of Co-SemDepth in the marine environments 
\end{itemize}

\section{Thesis Structure}

The rest of the thesis is structured as follows. In Chapter~\ref{chap:related}, the reviewed research papers related to the field are presented and analyzed. In Chapter~\ref{chap:cosemdepth}, the Co-SemDepth joint architecture is proposed and the experiments done for validating it are presented. In Chapter~\ref{chap:syntoreal}, we show the procedure followed to collect the TopAir dataset. In addition, the experiments carried out on the synthetic-to-real domain shift are analyzed. In Chapter~\ref{chap:style}, image style transfer techniques are explored for the sake of possible improvements. In Chapter~\ref{chap:marine}, the approaches followed in application to the marine domain are discussed, and the experiments done in the maritime domain are presented. 
Finally in Chapter~\ref{chap:conclusion}, the conclusions of this PhD thesis are presented with a focus on the achieved results.

\chapter{Related Work}  
\label{chap:related}

\ifpdf
    \graphicspath{{Chapter2/Figures/Raster/}{Chapter2/Figures/PDF/}{Chapter2/Figures/}}
\else
    \graphicspath{{Chapter2/Figures/Vector/}{Chapter2/Figures/}}
\fi

\section*{Summary}

In this chapter, the research found in the literature addressing the monocular depth estimation, semantic segmentation, and joint architectures are presented. The problem of monocular depth estimation in deep learning can be tackled with supervised methods (require large amount of annotated data) or self-supervised (supervision signals are drawn from stereo image pairs or video sequences). 
Semantic Segmentation is mostly achieved using supervised methods that can accept either single images or video frames as input. 
Methods applied in the marine domain can benefit from additional modules for horizon line estimation and water-land boundary detection. 
Joint architectures are also investigated for the sake of memory and time efficiency. Whether an FCN-based, a UNet-based, or a transformer-based architecture is deployed for joint learning, a feature encoder is usually shared among the vision tasks and the dedicated decoders are separated, with the possibility of adding cross-task attention between the decoders.
In addition, we discuss the synthetic and real datasets found in both the aerial and marine domains. There exist a number of annotated datasets in the aerial field. However, datasets that contain both depth and semantic segmentation annotations (that can be used for training joint networks) are limited. 
Few datasets were found in the marine field, and most of them do not contain annotations of depth and segmentation. 
In the end, research techniques found for tackling the gap between the synthetic and real domains are discussed. The methods include image-style transfer, image-mixing, and few-shot prompting. 

\section{Scene Analysis using UAVs}
Applications of UAVs are countless and they are expanding across diverse fields. In military and public safety, they are used for surveillance, attack, border patrol, firefighting, and search and rescue. In commerce, they are used for package delivery, infrastructure inspection, traffic monitoring, and surveying for various industries~\cite{uavapp2}. In agriculture, they can be used for crops irrigation, applying fertilizers, and mapping and monitoring the fields~\cite{uavapp}.
Most of the mentioned applications require heavy vision tasks like object detection and tracking, classification, depth estimation, optical flow estimation, motion detection, semantic segmentation, etc. Such tasks work on a pixel level and require a large amount of computational resources to be achieved. Therefore, there is a crucial need in the aerial field to perform such tasks using memory and time-efficient approaches. One of these approaches is the use of multi-tasking deep joint architectures. Such architectures can share parts among the different vision tasks, leading to effectiveness in the utilization of computational resources and lower inference time. 
Unfortunately, much of the work in outdoor scene analysis has been driven by advancements in the automotive field, leaving aerial scene understanding comparatively under-investigated. 
Besides, data-hungry deep networks require an abundance of annotated datasets for training. Such annotated data are hard to be found in the real domain due to the burdensome annotation process. At the same time, recent advancements in computer graphics have remarkably improved the realistic appearance of simulation environments. Hence, there is a recent shift in the scientific community towards training neural networks on synthetic data that can be obtained in big amounts with minimal effort and automatically annotated by the simulation engine~\cite{scicomm1, scicomm2, scicomm3, scicomm4}.

\section{Monocular Depth Estimation}
\begin{figure}%
    \centering
    \includegraphics[width=\linewidth]{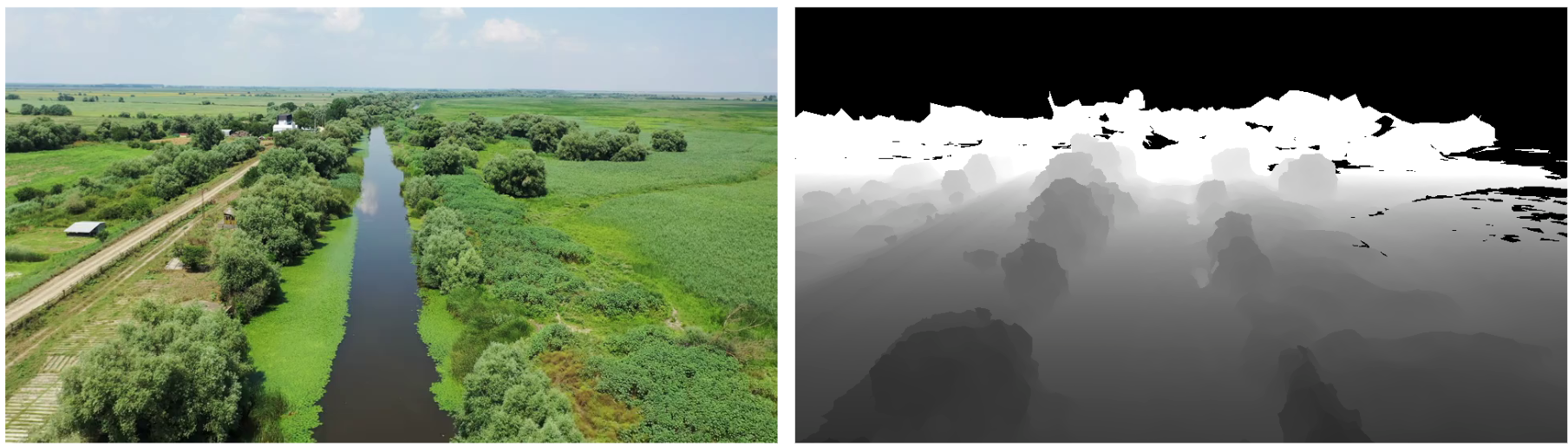}
    \caption{\centering Output depth map using SfM method~\cite{meshroom} resulting in holes and unknown regions in black}%
    \label{fig:sfm}%
\end{figure}

After investigating multiple works in the literature addressing the task of monocular depth estimation, it was found that this task can be performed using either classical methods or deep learning methods. While classical methods require minimal number of data and deep learning methods require an abundance of data for training, the accuracy and performance gain achieved using deep learning methods make them a better candidate in many cases for performing monocular depth estimation~\cite{d3}. One of the well-known classical methods for depth estimation is Structure from Motion (SfM). SfM focuses on the detection and tracking of feature points along given video frames and, then, using factorization to predict their 3D positions in space. This, however, generates sparse depth maps that contain 3D information for only the feature points selected. While interpolation techniques can be used to obtain a continuous depth map, SfM is often limited by the availability of feature correspondences that can be difficult to be captured in low-textured image regions, repetitive patterns, occlusions, and lighting changes. This leads to noise and missing parts in the output depth map~\cite{d15}, see Figure~\ref{fig:sfm} for an example. 

Instead of matching features across frames and geometric triangulation, \textbf{deep learning methods} can handle the above challenges by employing priors learned from a variety of training datasets. The architectures used for monocular depth estimation can rely on single images or image sequences as input. \begin{itemize}
    \item In \textbf{single-image} based architectures, the input to the neural network is a single image and the output is its predicted depth map. Some of the works found in this category are: the GAN-based architecture presented in~\cite{d2}, the transformer-based AdaBins~\cite{d8} that approached the depth estimation as a classification problem (bins of depth ranges) rather than regression, the U-Net based architectures proposed in~\cite{d16, d17, d18}, and the standard Fully Convolutional based networks (FCN) presented in~\cite{d19, d20, d21}. However, image-based methods suffer from the scale-ambiguity problem, where the scale of objects cannot be inferred accurately using a single image, and this leads to errors in the metric depth estimation. An attention-based network proposed in~\cite{d7} was used to solve such a problem by embedding the camera parameters as additional input to the network to reason over the physical size of objects and learn scale priors. Nevertheless, the network is huge and requires a big amount of data for training. 
    
    \item \textbf{image-sequence} based methods overcome the scale-ambiguity problem by making use of the temporal relation between input video frames. This helps to better understand the scale of the objects appearing in the images and leads to more accurate depth predictions. Some of the works used ConvLSTM or ConvGRU to extract temporal relation between input video frames~\cite{d1, d4, d6, d9}, some used an additional pose estimation network to infer camera poses from input video frames and help the original depth estimation network~\cite{d3,d5,d15}, some used a modified U-Net architecture for capturing spatio-temporal relation~\cite{d12}, and others used GAN or GCN~\cite{d13,d14,d11}. In their paper, ManyDepth~\cite{manydepth} shows the superiority of using multiple frames by comparing the predicted depth map from single frame input versus multiple frame input using the same model.
        
\end{itemize}

Whether depth estimation is done using single images or multiple frames, the training of the network can be achieved in a supervised or unsupervised fashion:

\begin{itemize}
    \item In \textbf{supervised methods}~\cite{li2017two,laina2016deeper,eigen2014depth,cs2018depthnet,m4depth}, depth maps ground truth should be provided to the network, where the network tries to learn the mapping function from images to depth maps by minimizing the cost between predictions and the ground truth. The main challenge for these methods is the scarcity of available annotated datasets that cover different scenarios and environments. Some of the most used datasets in the literature are KITTI~\cite{kitti} for autonomous driving applications, NYU-Depth-v2~\cite{nyu} for indoor scenes, and the synthetic MidAir ~\cite{midair} and TartanAir ~\cite{tartanair} datasets for aerial outdoor scenes. Recently, there exist foundation models, like DepthAnyhting~\cite{depth_anything, depth_anything_v2}, that are concerned with zero-shot depth estimation on any input image thanks to the huge amounts of data used for training such models.

    \item \textbf{Self-supervised methods} remove the limitation of requiring ground truth depth maps. They can be trained to predict the depth using as a supervision signal nearby video frames~\cite{casser2019depth,manydepth, masoumian2023gcndepth} or synchronized stereo image pairs ~\cite{huang2019unsupervised,monodepth}. Methods using synchronized stereo pairs require datasets that contain left and right stereo images and are trained to minimize image reconstruction losses. The methods using video frames require the scenes to be static, and there should be overlapping regions among the nearby frames to allow a pose estimation network to predict the pose transformation correctly. These requirements limit the application of such methods on dynamic scenes or videos of low frame rate, where there is little or no overlapping between the consecutive frames.
\end{itemize}

In~\cite{lu2024self}, they particularly handle images that contain reflective surfaces (water) and reformulate the depth estimation self-supervised problem to rely only on intra-frame features instead of inter-frame features as in stereo and adjacent-frames based methods. The reflection image in the water is regarded as another view and, thus, allows the application of multi-frame based methods on a single image. A segmentation network is used to detect the water mask region, a depth U-Net-based network is used to predict the depth map, and a photometric re-projection loss~\cite{jiang2020dipe} is employed. The challenge of weak texture in images received from USV is addressed in~\cite{marine_depth2} by the use of multi-scale feature fusion

\textbf{Challenges:} Few MDE methods mentioned their inference time, and no benchmarking was found in the literature comparing the inference time of different MDE methods. Also, very few methods were benchmarked on low-altitude aerial datasets~\cite{m4depth,miclea2021monocular}. Foundation models like DepthAnything~\cite{depth_anything_v2} are generally large in size ($>20M$ params) and may not be suitable for hardware deployment.


\section{Semantic Segmentation}
Semantic segmentation is widely addressed in the literature, in particular in the autonomous driving field~\cite{zhao2018icnet, romera2017erfnet, xie2021segformer, yu2018bisenet, wang2018understanding}  using the CityScapes~\cite{cordts2016cityscapes} dataset for benchmarking.

Semantic segmentation networks are normally trained in a supervised fashion, except for few cases~\cite{ss1,ss2}. A variety of architectures were found in the literature that can accept either single images or video frames as input. Unlike depth estimation, image-based semantic segmentation is not affected by the scale-ambiguity problem. This is because the prediction of the semantic class does not depend on the scale of the objects. Some state-of-the-art examples of \textbf{image-based methods} are ICNet~\cite{s15}, ERFNet~\cite{s3}, PSPNet~\cite{s16}, SegFormer~\cite{s4}, BiSeNet~\cite{s5}, MANet~\cite{s1}, Image Cascade Network~\cite{s2}, and RefineNet~\cite{s6}.

In contrast, \textbf{video-based methods} in semantic segmentation~\cite{s7, s8, s9, s10, s11, s12, s13, s14} focus on speeding up the semantic segmentation computation on individual frames by propagating the semantic segmentation prediction done on previous frames to successive video frames. Such propagation can be realized using light optical flow networks or interpolation techniques leveraging the overlapping regions between successive frames. This is done to speed-up inference time and to ensure temporal consistency among predicted semantic maps. However, the mIoU reported using video-based methods was usually lower than that of image-based methods~\cite{s9}.

A segmentation network, WaSR, is proposed in~\cite{wasr} for Unmanned Surface Vehicles in marine environments. The architecture is composed of a contracting path (encoder) and an expansive path (decoder). The encoder is an adaptation of the low-to-mid level parts of DeepLab2~\cite{deeplab2}. The decoder, instead, is a fusion of visual and inertial information. In particular, the horizon line is estimated by the camera-IMU projection~\cite{camera_imu}, and a binary mask distinguishing pixels below and above the horizon is constructed. Such an IMU mask is fused with the encoder features and it serves as a prior estimation of the water edge location and improves the water segmentation in the output. In~\cite{segmarine2}, they further enhance the segmentation of marine scenes by not only considering the boundary line between water and the shore, but also the boundaries between water and water obstacles. This is achieved by implementing a boundary feature extraction stream that takes as input the GT boundary image, and the extracted features are fused with the semantic feature stream during training time to strengthen its performance.

Recently, the foundation model SegmentAnything~\cite{segmentanything} has gained significant attention due to its notable results in zero-shot semantic segmentation. It is trained on a huge amount of data covering several domains and objects. However, SegmentAnything is dedicated for mesh segmentation, not class-based segmentation.

\textbf{Challenges:} Few semantic segmentation methods~\cite{nedevschi2021weakly, aeroscapes, zheng2023deep} were benchmarked on low-altitude aerial datasets. 


\section{Joint Vision Architectures}
The idea of a joint or multitasking architecture has been tackled in multiple works in the literature for various vision tasks. In~\cite{xu2020multi}, a joint architecture is implemented for image segmentation and classification, while in~\cite{qin2018joint}, they developed a joint network for motion estimation and segmentation. 

The main advantages of using joint architectures compared to single dedicated architectures are the efficiency in computational time and GPU memory (for hardware deployment, memory and speed are critical factors) as well as the possibility of enhancement in the accuracy of the predictions of each task, owing to the shared learning process. We hereby mention some of the works explored in the domain of joint depth estimation and semantic segmentation. 

Multiple works~\cite{j1,j2,j3, j4} used a shared encoder part in the joint architecture and two separate decoders dedicated for semantic segmentation and depth estimation. In~\cite{j9}, a multi-tasking transformer (Swin transformer) was developed where the encoder is shared, and each task has its own separate decoder. The encoder and the decoders have pyramidal structures with skip connections between the encoder and decoder levels. The architecture was tested on 6 tasks (depth, semantic segmentation, surface normals, edge detection, keypoints, and reshading). The results obtained show significant improvements compared to the single-task transformer. However, the model has a lot of parameters (231 million), and thus requires a big GPU memory. 


A Spatial-Channel Multi-task Prompting framework based on vision transformers was proposed in~\cite{j8}. With the help of task prompts and attention modules, the network learns task-generic, task-specific representations and cross-task interaction in the same network without using multiple networks. It produced state-of-the-art results on NYUD-V2~\cite{nyu} and PASCAL~\cite{pascal} datasets. However, the architecture appears somewhat complicated, and no information is provided about its inference time.

A modular design was proposed in~\cite{j4} where depth estimation and semantic segmentation were estimated separately using any chosen open-source methods, and then both predictions enter a joint refinement network (CNN) to produce better estimations. A similar idea was proposed in~\cite{j7} where multiple off-the-shelf networks were used to predict various vision tasks. The initial predictions of different tasks were then combined together through a distillation unit to make refined predictions. The predictions were distilled on different scales and then were aggregated to make the final predictions. While such methods help in enhancing the output accuracy compared to the initial predictions, the main disadvantage is the required additional computational time for the refinement phase. %

\section{Aerial Datasets}
There are various available datasets in the aerial domain. Such datasets can be captured in indoor, outdoor, urban, or non-urban environments. They can be synthetic or real, and they can have different view angles: forward-view or top-view (also called nadir-view). We hereby list the available datasets that are annotated with ground truth depth maps and semantic segmentation maps. In Table~\ref{table_datasets}, we provide a summary of the main characteristics of the datasets. \\

\noindent \textbf{Aerial Datasets for depth and semantic segmentation:}
\begin{itemize}
    \item \textit{MidAir}~\cite{midair}: a synthetic dataset collected using the AirSim simulator. It contains around 420K frames captured in different trajectories in 2 outdoor non-urban natural environments in different weather and light conditions. Its semantic segmentation annotation has 14 classes: \{animals, trees, dirt ground, ground vegetation, rocky ground, boulders, empty, water plane, man-made construction, road, train track, road sign, and others\}.
    \item \textit{SynDrone}~\cite{syndrone}: a synthetic dataset collected using Carla simulator at various heights. It contains around 72K frames in 8 different environments. It has annotation of 28 semantic classes: \{building, fence, pedestrian, pole, roadline , road, sidewalk, vegetation, cars, wall, traffic sign, sky, bridge, railtrack, guardrail, traffic light, water, terrain, rider, bicycle, motorcycle, bus, truck, others\}
    \item \textit{SkyScenes}~\cite{skyscenes}: a synthetic dataset collected using Carla simulator. It contains around 33.6K frames, and it covers a wide variety of pitch angles, and heights in 8 different environments (same towns as in SynDrone). It has an annotation of 28 semantic classes: \{building, fence, pedestrian, pole, roadline , road, sidewalk, vegetation, cars, wall, traffic sign, sky, bridge, railtrack, guardrail, traffic light, water, terrain, rider, bicycle, motorcycle, bus, truck, others\}.
    \item \textit{VALID}~\cite{valid}: a synthetic dataset collected using AirSim simulator. It contains around 6.7K frames collected in 6 different environments, and it has 30 semantic segmentation classes.
    \item \textit{WildUAV}~\cite{wilduav}: a real dataset containing around 1.5K frames captured in an outdoor non-urban environment. Its semantic segmentaton annotation contains 16 labels: \{sky, deciduous tree, coniferous tree, fallen trees, dirt ground, ground vegetation, rocks, water plane, building, fence, road, sidewalk, static car, moving car, people, and empty\}.
    \item \textit{DroneScape}~\cite{dronescape}: a real dataset collected in 8 different outdoor urban and non-urban environments. It contains around $>10K$ frames and it has 8 semantic segmentation classes: \{Land, Forest, Residential, Road, Little-objects, Water, Sky, and Hill\}.
    \item \textit{UAVid}~\cite{uavid}: a real dataset containing around 410 frames captured in an outdoor urban environment in China. The ground truth depth maps are provided for only some frames of the test set. Its semantic segmentation annotation has 8 classes: \{building, road, static car, tree, low vegetation, human, moving car, background clutter\}.
\end{itemize}

\begin{figure}[h!]%
    \centering
    
\begin{subfigure}[b]{\textwidth}
  \centering
  \includegraphics[width=0.93\linewidth, height=3.5cm]{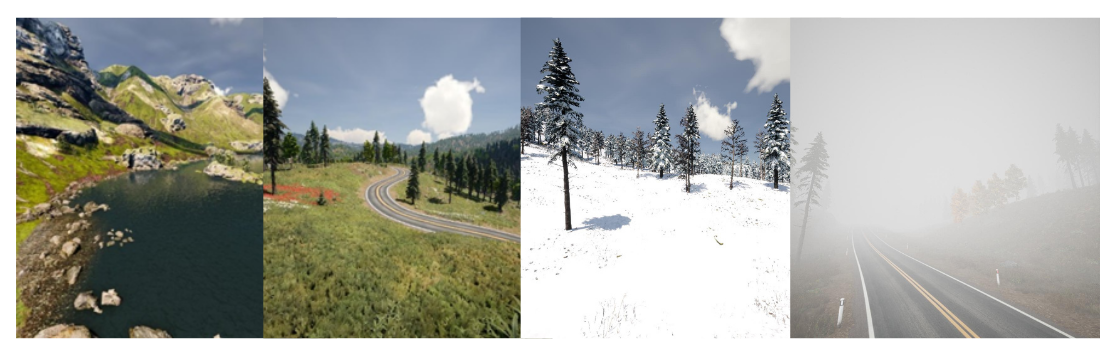}
  \caption{MidAir~\cite{midair}}
  \label{fig:midair}
\end{subfigure}
\begin{subfigure}[b]{\textwidth}
  \centering
  \includegraphics[width=0.93\linewidth, height=3.5cm]{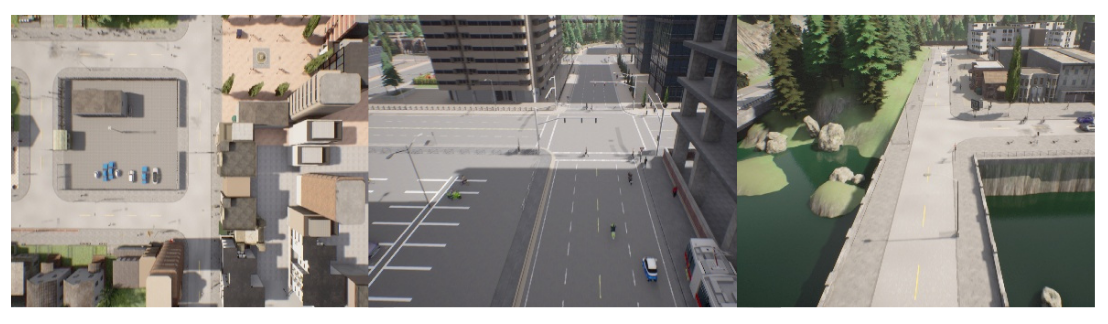}
  \caption{SynDrone~\cite{syndrone}}
  \label{fig:syndrone}
\end{subfigure}
\begin{subfigure}[b]{\textwidth}
  \centering
  \includegraphics[width=0.93\linewidth, height=3.5cm]{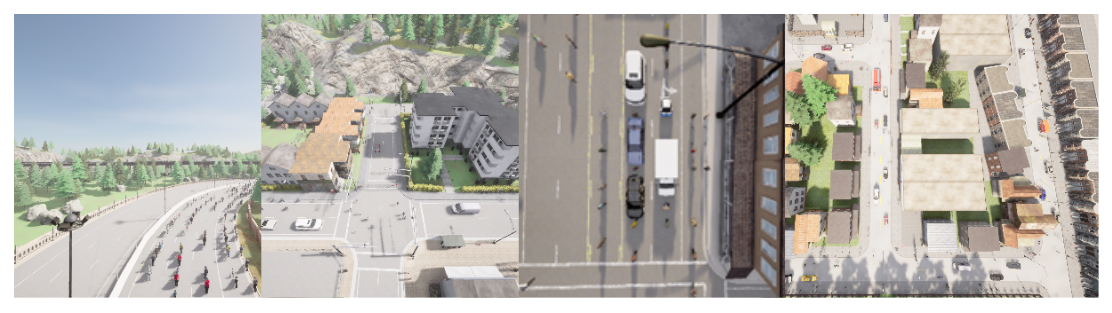}
  \caption{SkyScenes~\cite{skyscenes}}
  \label{fig:skyscenes}
\end{subfigure}
\begin{subfigure}[b]{\textwidth}
  \centering
  \includegraphics[width=0.93\linewidth, height=3.5cm]{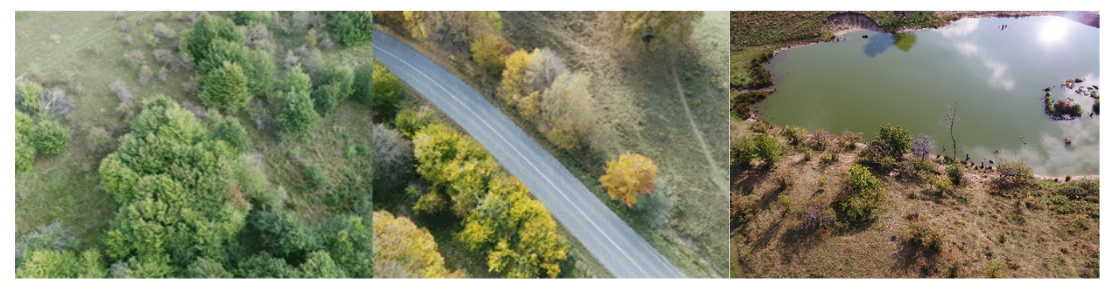}
  \caption{WildUAV~\cite{wilduav}}
  \label{fig:wilduav}
\end{subfigure}
\begin{subfigure}[b]{\textwidth}
  \centering
  \includegraphics[width=0.93\linewidth, height=3.5cm]{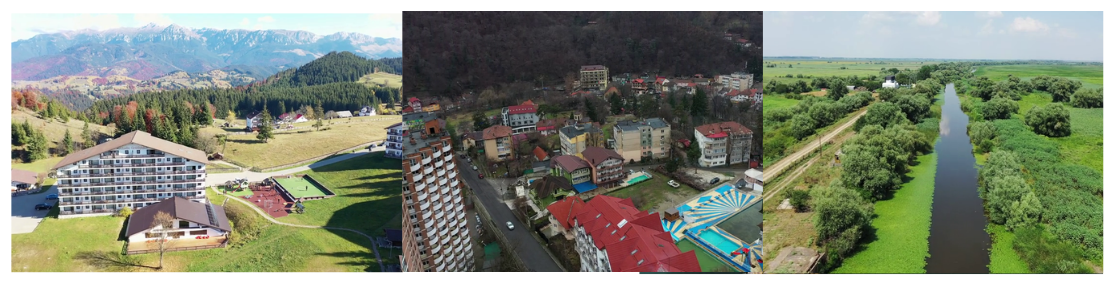}
  \caption{DroneScapes~\cite{dronescape}}
  \label{fig:dronescapes}
\end{subfigure}

    \caption{\centering Sample images from different aerial datasets that contain depth and semantic segmentation annotation}%
    \label{fig:sample1}%
\end{figure}

Image samples from the datasets are shown in Figure~\ref{fig:sample1}. \\

\noindent \textbf{Aerial Datasets for depth estimation:}
\begin{itemize}
    \item \textit{TartanAir}~\cite{tartanair}: a synthetic dataset collected using the AirSim simulator. It contains more than 1 million frames collected in indoor and outdoor environments with various view angles. While this dataset contains semantic segmentation annotation, it is mesh segmentation, not class segmentation. 
    \item \textit{ESPADA}~\cite{espada}: a synthetic dataset collected using the AirSim simulator in 49 different scenes. It contains around 80K frames. 
    \item \textit{UrbanScene3D}~\cite{urbanscene}: a dataset containing both synthetic and real images. The synthetic data were collected using the AirSim simulator. It contains around 128K frames collected in urban environments. As in Tartan Air, the semantic segmentation in this dataset is mesh segmentation, not class segmentation. 
    \item \textit{UseGeo3D}~\cite{usegeo}: a real dataset containing 829 frames collected in semi-urban environments.
\end{itemize}

\begin{figure}[h!]%
    \centering

\begin{subfigure}[b]{\textwidth}
  \centering
  \includegraphics[width=0.93\linewidth, height=3.2cm]{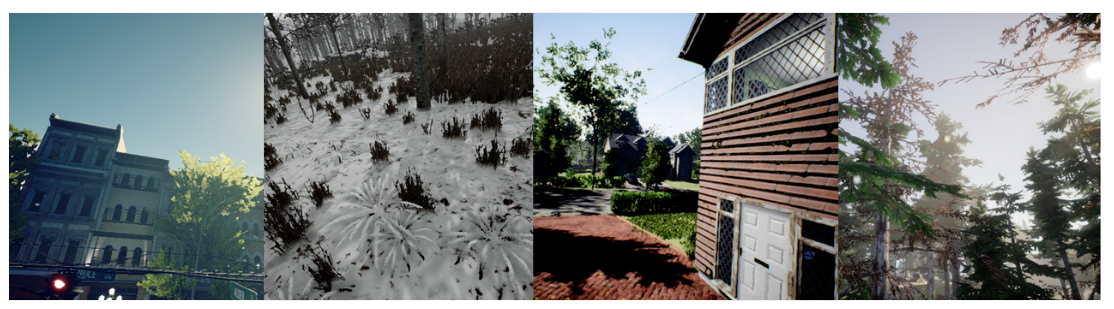}
  \caption{TartanAir~\cite{tartanair}}
  \label{fig:syndrone}
\end{subfigure}
\begin{subfigure}[b]{\textwidth}
  \centering
  \includegraphics[width=0.93\linewidth, height=3.2cm]{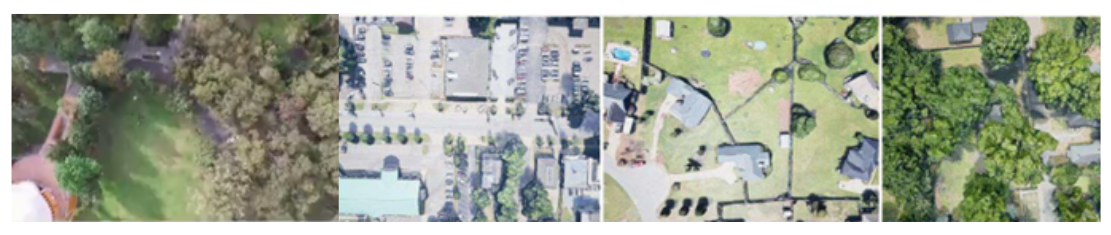}
  \caption{ESPADA~\cite{espada}}
  \label{fig:midair}
\end{subfigure}
\begin{subfigure}[b]{\textwidth}
  \centering
  \includegraphics[width=0.93\linewidth, height=3.2cm]{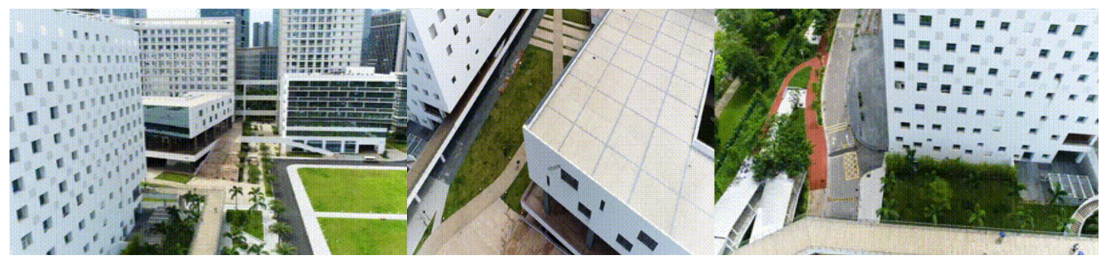}
  \caption{UrbanScene3D~\cite{urbanscene}}
  \label{fig:dronescapes}
\end{subfigure}
\begin{subfigure}[b]{\textwidth}
  \centering
  \includegraphics[width=0.93\linewidth, height=3.2cm]{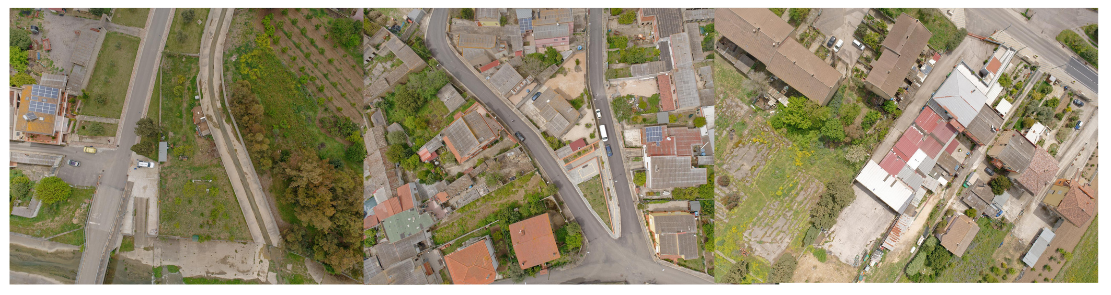}
  \caption{UseGeo~\cite{usegeo}}
  \label{fig:skyscenes}
\end{subfigure}

    \caption{\centering Sample images from different aerial datasets that contain only depth annotation}%
    \label{fig:sample2}%
\end{figure}

Image samples from the datasets can be found in Figure~\ref{fig:sample2}. \\

\noindent \textbf{Aerial Datasets for semantic segmentation:}
\begin{itemize}
    \item \textit{SynthAer}~\cite{synthaer}: a synthetic dataset collected using the modeling tool Blender~\cite{blender} in an urban environment at 30-50 meters height. It contains 765 frames and 8 semantic labels: \{Building, Vehicle, Road, Footpath, Sky, Tree, Vegetation, Wall\}
    \item \textit{Aeroscapes}~\cite{aeroscapes}: a real dataset collected in urban and rural environments with various altitudes and view angles. It is composed of 3269 frames and its semantic segmentation annotation has 12 classes: \{ background, person, bike, car, drone, boat, animal, obstacle, construction, vegetation, road, sky \} 
    \item \textit{Ruralscapes}~\cite{ruralscapes}: a real dataset collected in semi-urban environments. It contains 1127 frames and its semantic segmentation annotation has 12 classes: \{forest, residential, land, sky, hill, road, church, fence, water, person, car, haystack\}
    \item \textit{VDD}~\cite{vdd}: a real dataset collected in urban and semi-urban environments. It contains 400 images and its semantic segmentation annotation has 7 classes: \{wall, roof, road, water, vehicle, vegetation, others\}
    \item \textit{UDD}~\cite{udd}: a real dataset collected in urban and semi-urban environments (same environments in VDD). It contains 200 images and its semantic segmentation annotation has 6 classes: \{facade, road, vegetation, vehicle, roof, others\}
    \item \textit{FSI, Flood Satellite Imagery}~\cite{ninja2}: a real dataset collected in semi-urban environments where some of them were flooded. It contains 261 images, and we are only interested in the images without flood. Its semantic segmentation annotation contains 25 labels: \{background, grass, trees, roof, vehicle, chimney, secondary structure, swimming pool, power lines, window, road, flooded, forest, street light, water, garbage bins, trampoline, satellite antenna, parking area, solar panel, under construction, boat, sports complex, industrial site, water tank.\}
    \item \textit{ICG}~\cite{icg}: a real dataset collected at low-medium heights in semi-urban environments. It contains 600 images with semantic segmentation annotation of 20 classes: \{tree, rocks, dog, fence, grass, water, car, fence-pole, other vegetation, paved area, bicycle, window, dirt, pool, roof, door ,gravel, person, wall, obstacle\}
    
\end{itemize}

\begin{figure}[h!]%
    \centering

\begin{subfigure}[b]{\textwidth}
  \centering
  \includegraphics[width=0.93\linewidth, height=3.5cm]{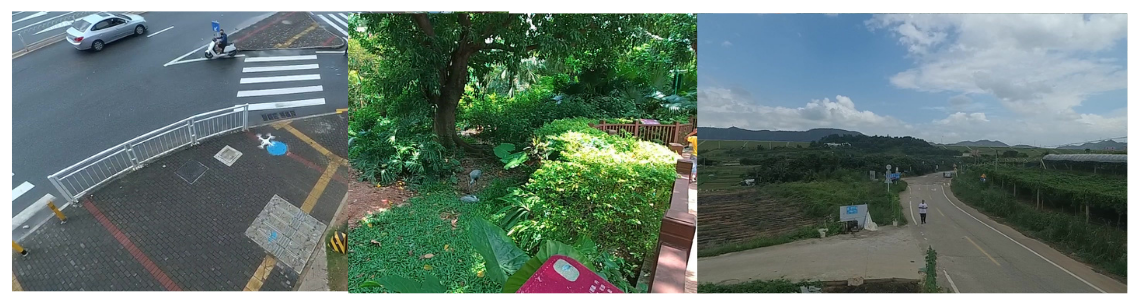}
  \caption{AeroScapes~\cite{aeroscapes}}
  \label{fig:aeroscapes}
\end{subfigure}
\begin{subfigure}[b]{\textwidth}
  \centering
  \includegraphics[width=0.93\linewidth, height=3.5cm]{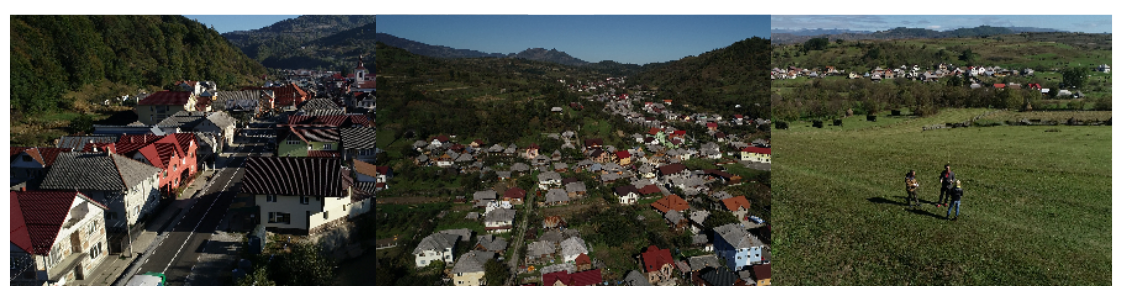}
  \caption{RuralScapes~\cite{ruralscapes}}
  \label{fig:ruralscapes}
\end{subfigure}
\begin{subfigure}[b]{\textwidth}
  \centering
  \includegraphics[width=0.93\linewidth, height=3.5cm]{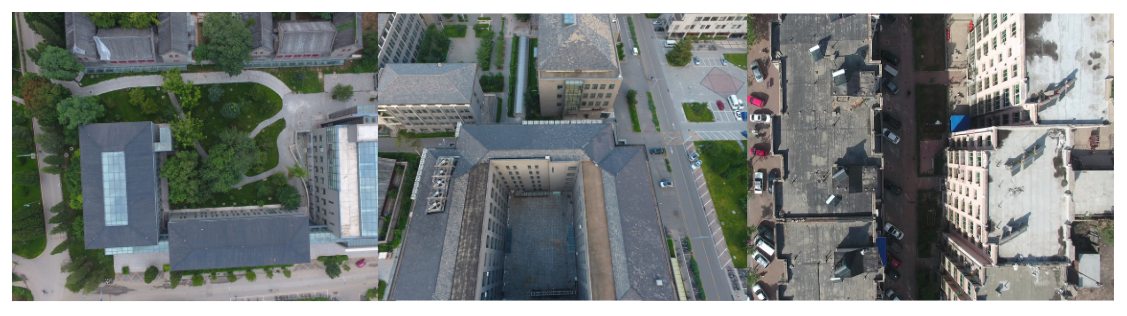}
  \caption{UDD~\cite{udd}}
  \label{fig:udd}
\end{subfigure}
\begin{subfigure}[b]{\textwidth}
  \centering
  \includegraphics[width=0.93\linewidth, height=3.2cm]{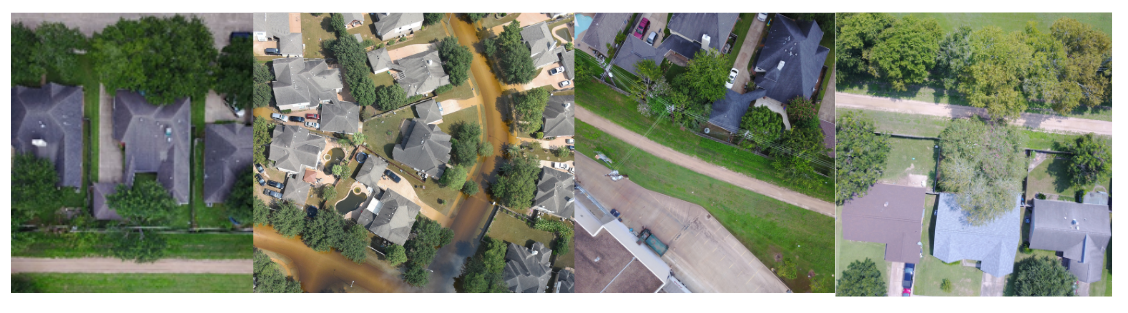}
  \caption{FSI~\cite{ninja2}}
  \label{fig:fsi}
\end{subfigure}
\begin{subfigure}[b]{\textwidth}
  \centering
  \includegraphics[width=0.93\linewidth, height=3.5cm]{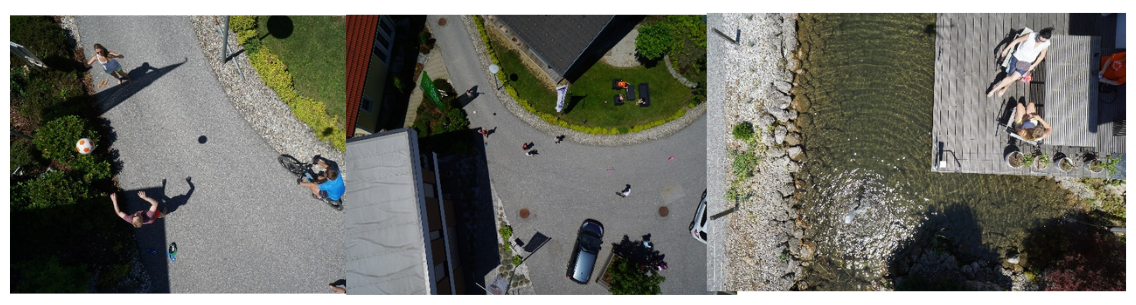}
  \caption{ICG~\cite{icg}}
  \label{fig:icg}
\end{subfigure}

    \caption{\centering Sample images from different aerial datasets that contain only semantic segmentation annotation}%
    \label{fig:sample3}%
\end{figure}

Image samples from the datasets can be found in Figure~\ref{fig:sample3}.

\begin{sidewaystable}
\begin{center}
\caption{\centering Available aerial outdoor datasets and their main characteristics. D means depth maps, S means semantic segmentation maps, and Loc means camera location information. top-view means pitch angle close to $\ang{90}$. The top part of the table lists datasets with both depth and semantic annotations. The middle part contains datasets with only depth annotation, and the bottom part contains datasets with only semantic annotation.}
\label{table_datasets}
\begin{tabular}{ c | c | c | c | c | c | c | c | c }
\hline
 Dataset & Type & img/vid & Annotation & View & Size & Environment & Height & \#Classes \\
\hline
\textit{MidAir} & synth & vid & D+S+Loc & fwd & $420K$ & rural & low-med & $14$ \\
\textit{SynDrone} & synth & vid & D+S+Loc & fwd + top & $72K$ & urb+semi-urb & low-med-high $(20,50,80m)$ & $28$ \\
\textit{SkyScenes} & synth & vid & D+S+Loc & fwd + top & $6.7K$ & urb+semi-urb & low-med $(15,35,60m)$ & $28$ \\
\textit{VALID} & synth & vid & D+S & top & $6.7K$ & urb+semi-urb & med-high $(20,50,100m)$ & $30$ \\
\textit{WildUAV} & real & vid & D+S+Loc & top & $1.5K$ & rural & low-med $(50m)$ & $16$ \\
\textit{Dronescape} & real & vid & D+S+Loc & fwd & $>10K$ & urb+rural & high & $8$ \\
\textit{UAVid} & real & vid & D+S & top & $410$ & urban & high & $8$ \\
\hline
\hline
\textit{TartanAir} & synth & vid & D+Loc & fwd+top & $1M$ & rural+urb & low-med & - \\
\textit{ESPADA} & synth & vid & D+Loc & top & $80K$ & urb+rural & med-high$(30-100m)$ & - \\

\textit{UrbanScene3D} & synth+real & vid & D+Loc & fwd+top & $128K$ & urb & low-med & - \\
\textit{UseGeo3D} & real & vid & D+Loc & top & $829$ & semi-urb & high$(80m)$ & - \\
\hline
\hline
\textit{SynthAer} & synth & vid & S & fwd+top & $765$ & semi-urb & med & $8$ \\
\textit{Aeroscapes} & real & vid & S & fwd+top & $3.3K$ & urb & low-med $(5-50m)$ & $12$ \\
\textit{RuralScapes} & real & vid & S & fwd & $1.1K$ & semi-urb & med-high & $12$ \\
\textit{VDD}& real & img & S & top & $400$ & urb & med-high $(50-120m)$ & $7$ \\
\textit{UDD} & real & img & S & top & $200$ & urb & med-high $(60-100m)$ & $6$ \\
\textit{FSI} & real & img & S & top & $261$ & semi-urb & med-high & $25$ \\
\textit{ICG} & real & img & S & top & $600$ & semi-urb & low-med $(5-30m)$ & $20$ \\

\hline
\hline
\textbf{TopAir} & synth & vid & D+S+Loc & top  & $10K$ & rural+urb & low-med-high$(5-100m)$ & $9$ \\
\hline
\end{tabular}
\end{center}
\end{sidewaystable}



\section{Maritime Datasets}
Different from aerial datasets, the datasets available in the maritime domain are limited. 

\noindent \textbf{Marine datasets annotated for depth estimation or semantic segmentation:}
\begin{itemize}
    \item MassMIND~\cite{massmind}: contains 2916 diverse Long Wave InfraRed (LWIR) images captured around the Boston Harbor in Massachusetts, USA, in various weathers, seasons, and time of the day. Each image is annotated with pixel-level semantic segmentation across 7 classes: Sky, Water, Bridge, Obstacles, Living Obstacles, Background, Self. Resolution is 640$\times$512. 
    \item MaSTr1325~\cite{mastr}: consists of 1325 diverse images captured in the Gulf of Koper, Slovenia, with a real USV. It includes various weather conditions and times of day. Each image is per-pixel semantically segmented with one of four labels: Obstacles and Environment, Water, Sky, or Unknown. In addition, GPS and IMU data are provided and time-synchronized with the images. Resolution is 512$\times$384.
    \item MODD2~\cite{modd2}: an extended version of MaSTr1325 dataset containing 11675  stereo frames with a resolution of 1278$\times$958 pixels. The dataset was recorded over a period of several months to ensure diversity.
    \item MODS~\cite{MODS}: A dataset of 81k images acquired in the Slovanian coast. Their dataset annotations were performed in a very strict procedure to ensure quality and accuracy. The segmentation classes are 3: Sky, Water, and Others. Most of the obstacles present in the dataset are in the range of 15-20 meters away from the USV.
    \item MIT\footnote{\href{https://seagrant.mit.edu/auvlab-datasets-marine-perception-1/}{Available on this link}}: a multi-modal sensor dataset for mobile robotics research in the marine domain. It is composed of videos acquired from optical and IR cameras, point clouds acquired from 3D Lidar, RADAR images, and IMU data. The data was captured during various missions at sea with several obstacles and weather conditions. Point clouds can be used as ground truth for depth estimation tasks.
\end{itemize}

\begin{figure}[h!]%
    \centering
    
\begin{subfigure}[b]{\textwidth}
  \centering
  \includegraphics[width=\linewidth, height=3.3cm]{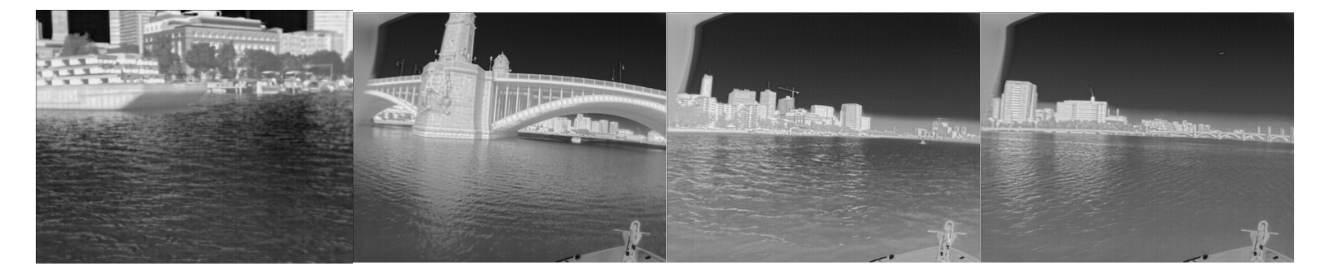}
  \caption{MassMIND~\cite{massmind}}
  \label{fig:massmind}
\end{subfigure}
\begin{subfigure}[b]{\textwidth}
  \centering
  \includegraphics[width=\linewidth, height=3.3cm]{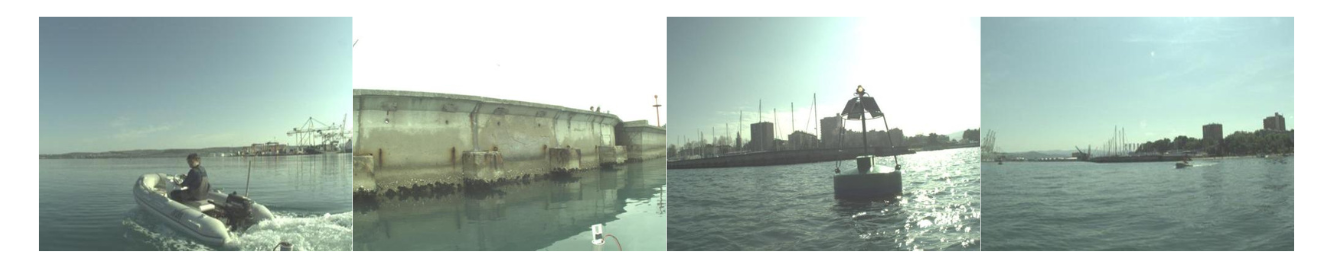}
  \caption{MaSTr1325~\cite{mastr}}
  \label{fig:mastr}
\end{subfigure}
\begin{subfigure}[b]{\textwidth}
  \centering
  \includegraphics[width=\linewidth, height=3.3cm]{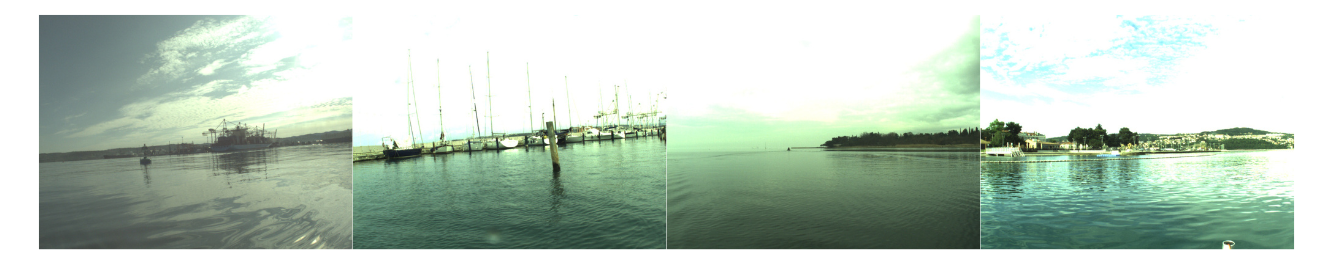}
  \caption{MODD2~\cite{modd2}}
  \label{fig:modd2}
\end{subfigure}
\begin{subfigure}[b]{\textwidth}
  \centering
  \includegraphics[width=\linewidth, height=3.3cm]{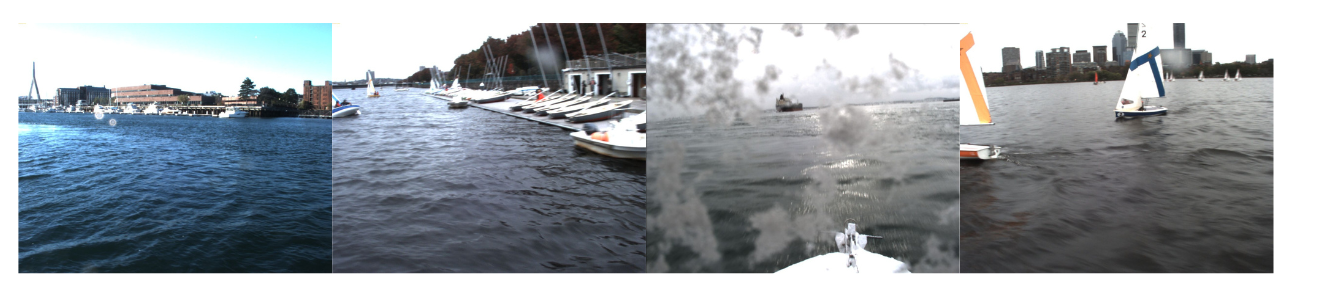}
  \caption{MIT}
  \label{fig:mit}
\end{subfigure}

    \caption{\centering Sample images from different marine datasets that are annotated with either depth or semantic segmentation}%
    \label{fig:sample_marine1}%
\end{figure}

See Figure~\ref{fig:sample_marine1} for samples from the datasets.

There are other marine datasets without depth or semantic segmentation annotations that can be used only for qualitative evaluation.

\noindent \textbf{Non annotated Marine datasets:}
\begin{itemize}
    \item ABOships~\cite{aboships}: a dataset only annotated with bounding boxes belonging to one of 11 categories.
    The data is composed of 10k images acquired in the Aura River and the port of Turku in Finland. 
    \item Singapore Maritime Dataset (SMD)\footnote{\href{https://sites.google.com/site/dilipprasad/home/singapore-maritime-dataset}{Available on this link}}: a dataset of optical and IR video frames captured at various times of the day and weather conditions. Obstacles are annotated with bounding boxes.
    \item SeaShips~\cite{seaships}: It contains images of sea ships annotated with bounding boxes. The images are taken from the shore around Hengqin Island, Zhuhai city, China.
    \item MarDCT~\cite{mardct}: A dataset captured in Venice, Italy, from a top view. Images contain different boats, with each image containing only one boat in the center. 
    \item Marvel~\cite{marvel}: A large-sized dataset of marine vessels images. The images are collected from the internet, and each image contains only one vessel.
\end{itemize}

\begin{figure}[h!]%
    \centering
    
\begin{subfigure}[b]{\textwidth}
  \centering
  \includegraphics[width=\linewidth, height=3.3cm]{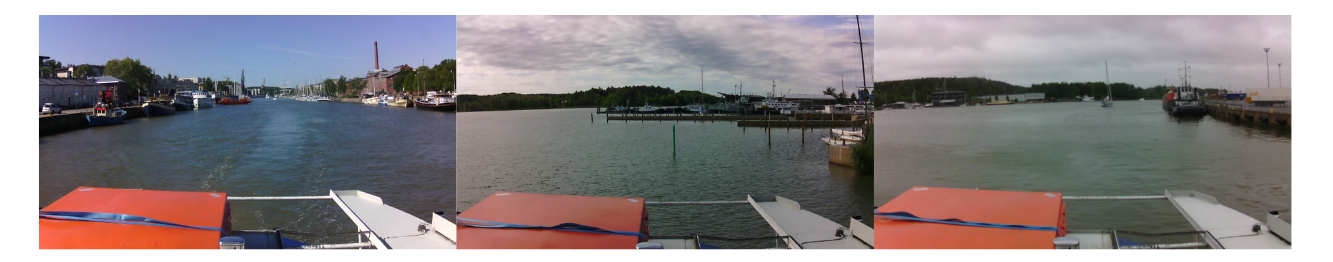}
  \caption{ABOships~\cite{aboships}}
  \label{fig:aboships}
\end{subfigure}
\begin{subfigure}[b]{\textwidth}
  \centering
  \includegraphics[width=\linewidth, height=3.3cm]{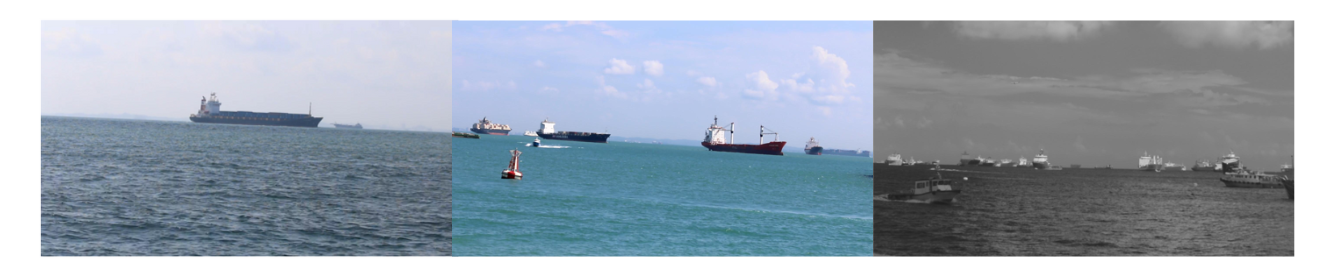}
  \caption{SMD~\cite{SMD}}
  \label{fig:smd}
\end{subfigure}
\begin{subfigure}[b]{\textwidth}
  \centering
  \includegraphics[width=\linewidth, height=3.3cm]{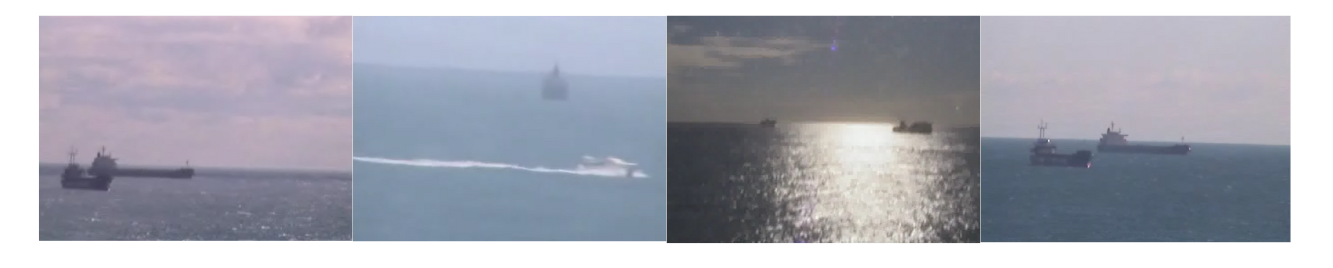}
  \caption{MarDCT~\cite{mardct}}
  \label{fig:mardct}
\end{subfigure}
\begin{subfigure}[b]{\textwidth}
  \centering
  \includegraphics[width=\linewidth, height=3.3cm]{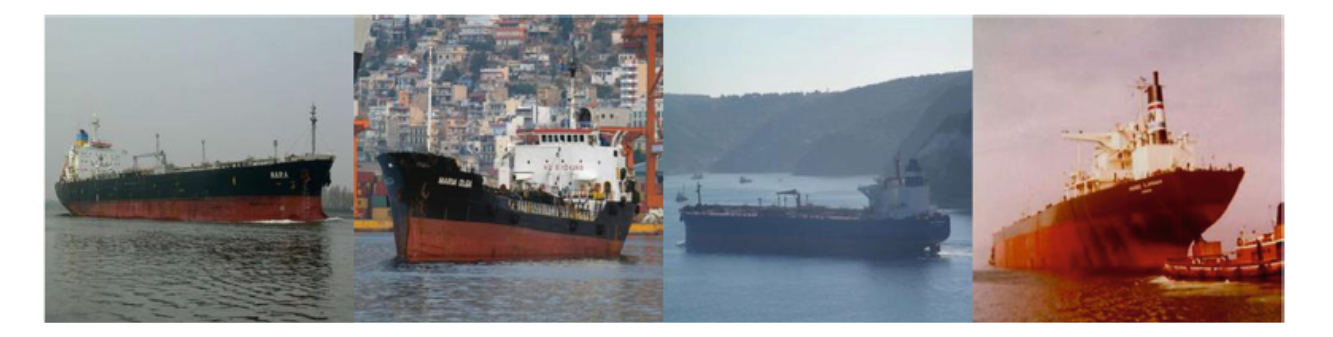}
  \caption{Marvel~\cite{marvel}}
  \label{fig:marvel}
\end{subfigure}

    \caption{\centering Sample images from different non-annotated marine datasets}%
    \label{fig:sample_marine2}%
\end{figure}

See Figure~\ref{fig:sample_marine2} for samples from the datasets.

\section{Synthetic-to-Real Domain Shift}
Many works addressed the synthetic-to-real domain shift problem in either monocular depth estimation or semantic segmentation. However, most of these works, driven by the current trend in the autonomous driving field, focused on the ground vehicle domain. Only few papers~\cite{skyscenes, syndrone, quantifying, ruralsynth, wilduav} investigated the synthetic-to-real domain shift in the aerial field, and they mostly focused on semantic segmentation. In the following, a summary of the reviewed works is presented.

\textbf{In the automotive field:}
The techniques used for improving the synthetic-to-real generalization in depth estimation or semantic segmentation were found to be either image mixing techniques~\cite{song2024transformer, chen2024transferring}, or image style transfer techniques~\cite{atapour2018real, xiao2022transfer, zheng2018t2net, chen2019learning}. In image mixing, the synthetic images used for training a neural network are mixed with parts or elements from the real images (for example, cars, persons, or buildings). This enhances a bit the realistic appearance of the synthetic images, and hence, improves the generalization to the real domain during testing. 

In image style transfer, generative models like GANs or Diffusion models are used to transform the appearance of the synthetic images to look more realistic, and consequently, when used to train a deep model, it enhances its generalization to the real domain. 

The widely-used benchmarks for the evaluation of synthetic-to-real domain shift in the automotive field are SYNTHIA-to-Cityscapes and SYNTHIA-to-Mapillary~\cite{synthia, cordts2016cityscapes, mapillary}.

\textbf{In the aerial field:}
The main works found addressing the synthetic-to-real problem~\cite{skyscenes, syndrone, quantifying, ruralsynth} focused on semantic segmentation. 

In~\cite{syndrone}, the DeepLabV3 network~\cite{deeplabv3} was trained on the SynDrone dataset and evaluated on the real datasets: UAVid, Aeroscapes, ICG, and UDD. In addition, the effect of changing training and testing towns or altitudes was demonstrated, and it was shown that even when using the same synthetic dataset, changing the environment between training and testing or changing the altitude leads to a drop in performance.

In~\cite{skyscenes}, 3 different semantic segmentation architectures were trained on synthetic datasets (SkyScenes and SynDrone), and the synthetic-to-real performance was evaluated on 3 real datasets: UAVid, Aeroscapes, and ICG. It was found that networks trained on SkyScenes generalize better to real datasets than those trained on SynDrone, due to the variability of heights and pitch angles in SkyScenes as well as the better representation of human and vehicle classes. It was also found that changing the height and pitch angles between the synthetic and the real datasets greatly affects the generalization accuracy. It was found as well that augmenting part of the testing real data with the synthetic data during training improves the model's generalization performance.

In~\cite{quantifying}, a study on the perceptual and structural complexity of datasets (SkyScenes and DroneScapes) was presented, and it was suggested that the gap between a synthetic and a real dataset can be quantified by measuring the difference between their corresponding perceptual complexity scores.

In~\cite{ruralsynth}, a synthetic dataset similar to RuralScapes~\cite{ruralscapes} was generated and used to train a semantic segmentation network using zero-shot and low-shot regimes. In low-shot, it was found that by only adding a small percentage $(1-5\%)$ of the real Ruralscapes to the training synthetic data, the evaluation accuracy on Ruralscapes was boosted. 

In~\cite{udareview}, they divide the UDA methods into three levels: on the input, on the feature representation, and on the output. Methods working at the input level include image-style transfer techniques that convert the style of the input images to the style of the target real domain before training on the input images. The core idea of adaptation at the feature-level is to force the feature extractor of the network to predict domain-invariant latent representations from the source and target domains. After that, the network classifier should be able to classify both the source and target representations correctly by relying solely on the supervision from the source. Adaptation at the output level can benefit from adversarial strategies applied to the output low-dimensional space spanned by the segmentation maps. A domain discriminator is trained to distinguish the source and target domains from the given predicted maps, and the segmentation network has to fool the discriminator by aligning the distribution of the predicted labels across the two domains. In~\cite{unsupervisedremote}, an unsupervised domain adaptation semantic segmentation network is proposed and tested on remote sensory images. The approach focuses on storing invariant features of the source and target domains using a memory module. 

For depth estimation, in~\cite{wilduav}, some experiments were performed where depth estimation networks were trained on selected images from the MidAir dataset~\cite{midair} and tested on the real WildUAV dataset~\cite{wilduav}.

\textbf{In the marine domain:}
The work in~\cite{marinesynthtoreal} addressed the synthetic-to-real domain shift in object detection YOLO~\cite{yolo} by adding a limited number of annotated real data to the synthetic training data, while a synthetic marine environment was proposed in~\cite{marinesynthetictoreal2} to support the research in this area.




\chapter{Proposed Co-SemDepth}
\label{chap:cosemdepth}

\ifpdf
    \graphicspath{{Chapter3/Figures/Raster/}{Chapter3/Figures/PDF/}{Chapter3/Figures/}}
\else
    \graphicspath{{Chapter3/Figures/Vector/}{Chapter3/Figures/}}
\fi
\section*{Summary}

In this chapter, we first give a brief overview of M4Depth, our backbone network for depth estimation, then describe our developed M4Semantic segmentation network. After that, we merge the two networks and shed light on our proposed joint Co-SemDepth architecture. 
To summarize, M4Depth~\cite{m4depth} is a network with a pyramidal structure that incorporates motion and video frames in the process of supervised training to enhance the depth estimation. M4Semantic is an adapted version of M4Depth modified to be dedicated to semantic segmentation. 
Co-SemDepth is designed by merging the encoder part in M4Depth and M4Semantic, and separating the dedicated decoders for the two tasks.


In addition, the setup used to conduct the experiments on Co-SemDepth is explained. Then, we discuss the experiments conducted using the developed joint architecture Co-SemDepth to validate its effectiveness and competency against other state-of-the-art methods. Results reveal that using Co-SemDepth is more time and memory-efficient than using single dedicated architectures. In addition, Co-SemDepth is notably faster than other joint architectures (and single ones), while being competent in the accuracy of both depth estimation and semantic segmentation. 

\section{Architecture Design}

\subsection{Backbone Depth Network}
The starting point of our work is the M4Depth network proposed in~\cite{m4depth}. Our motivations to choose such a network are that, to the best of our knowledge, it produces the current top results in monocular depth estimation on the synthetic MidAir~\cite{midair} aerial dataset, and its model weights and code are publicly available. In addition, the encoder-decoder modularity of their architecture makes it a convenient candidate to be transformed into a joint architecture. Moreover, M4Depth has a unique architecture that integrates camera motion data to enhance the depth estimation. We adopt the M4Depth architecture as it is, without changes, except for the number of layers we use 5 instead of 6 layers, and this will be justified in the Experiments section.

\begin{figure}[h!]%
    \centering
    {\includegraphics[width=\linewidth]{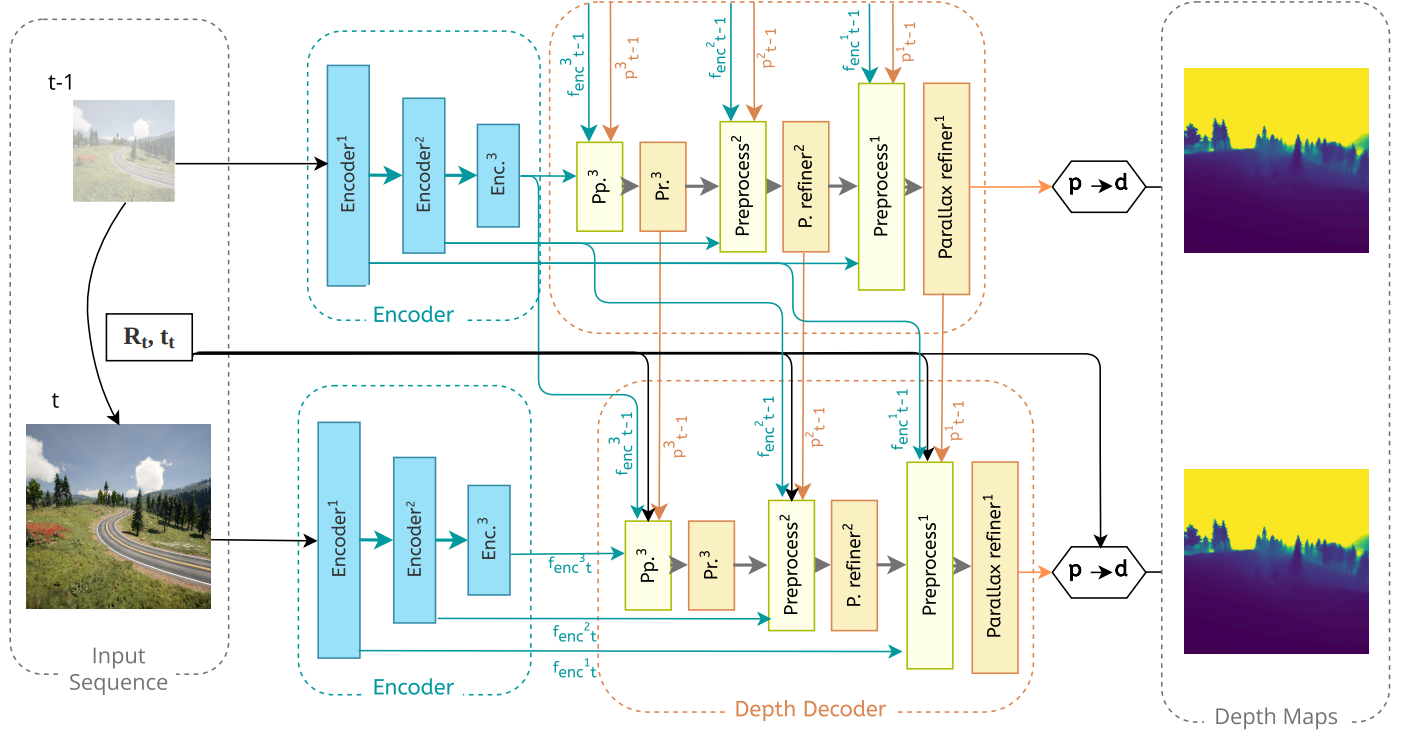} }
    \caption{\centering The architecture of M4Depth. It is fed by two consecutive frames and the camera motion. Each convolution layer is followed by a ReLU activation unit. The depth map is shown in the viridis scale. Details of the modules are given in Figure~\ref{fig:modules_depth} 
    }%
    \label{fig:m4depth}%
\end{figure}

The architecture of M4Depth~\cite{m4depth}, see Figure~\ref{fig:m4depth}, is an adaptation of the standard U-Net encoder-decoder network~\cite{unet} trained to predict parallax maps to be then transformed into depth maps. The authors define parallax as a function of perceived motion, thus it can be seen as a general form of stereo disparity for an unconstrained camera baseline. Similar to disparity, parallax can be related to the depth of points in space appearing in the image. They train the network to estimate parallax instead of depth directly to make it robust to unseen environments. This is because estimating depth from input images make the network tied to the training data distribution while estimating parallax is more robust and is less dependent on the training data distribution. One of the unique things about M4Depth is that it incorporates motion information in the process of supervised training and exploits such information for enhancing the depth prediction. 

The network takes as input a sequence of $n$ video frames (we choose \(n=3\)) and the camera transformation $T$ between every two consecutive frames. At each time step $t$, the encoder takes a new video frame $I_t$ and extracts image features at different scales using its pyramidal structure. Each encoder level is composed of two convolutional layers and a domain-invariant normalization layer (DINL)~\cite{dinl} to increase the network robustness to varied colors and luminosity conditions. 

Then, the decoder takes the feature maps at different resolutions obtained by the encoder at time $t$, the features extracted from the previous frame ($t-1$), the parallax map predicted at time $t-1$, and the camera motion transformation $T_t$ to predict the parallax map of the current frame $I_t$.  This parallax $\rho_t$ is then transformed into a depth map $d_t$ following the pixel-wise transformation proposed in Equation~\ref{eq:parallax}. 

\begin{equation}
\label{eq:parallax}
    \rho_{i_{t}j_{t}} = \frac{\sqrt{(f_x t_x - t_z i_V)^2+(f_yt_y-t_zj_V)^2}}{z_{i_{t}j_{t}}z_V + t_z}
\end{equation}

Where $z_{i_{t}j_{t}}$ is the depth of the point $P$ located in pixels $(i,j)$ at time $t$, $f_x$ and $f_y$ are the focal lengths of the camera, $t_x$, $t_y$, and $t_z$ are the camera translation between two consecutive frames, $i_V$ and $j_V$ are the image pixel coordinates of the point $P$ in the plane of a virtual camera $V$ whose origin is the same as the camera at time $t$ but with the orientation of the camera at time $t-1$.

\begin{figure}[h!]%
    \centering
    {\includegraphics[width=\linewidth]{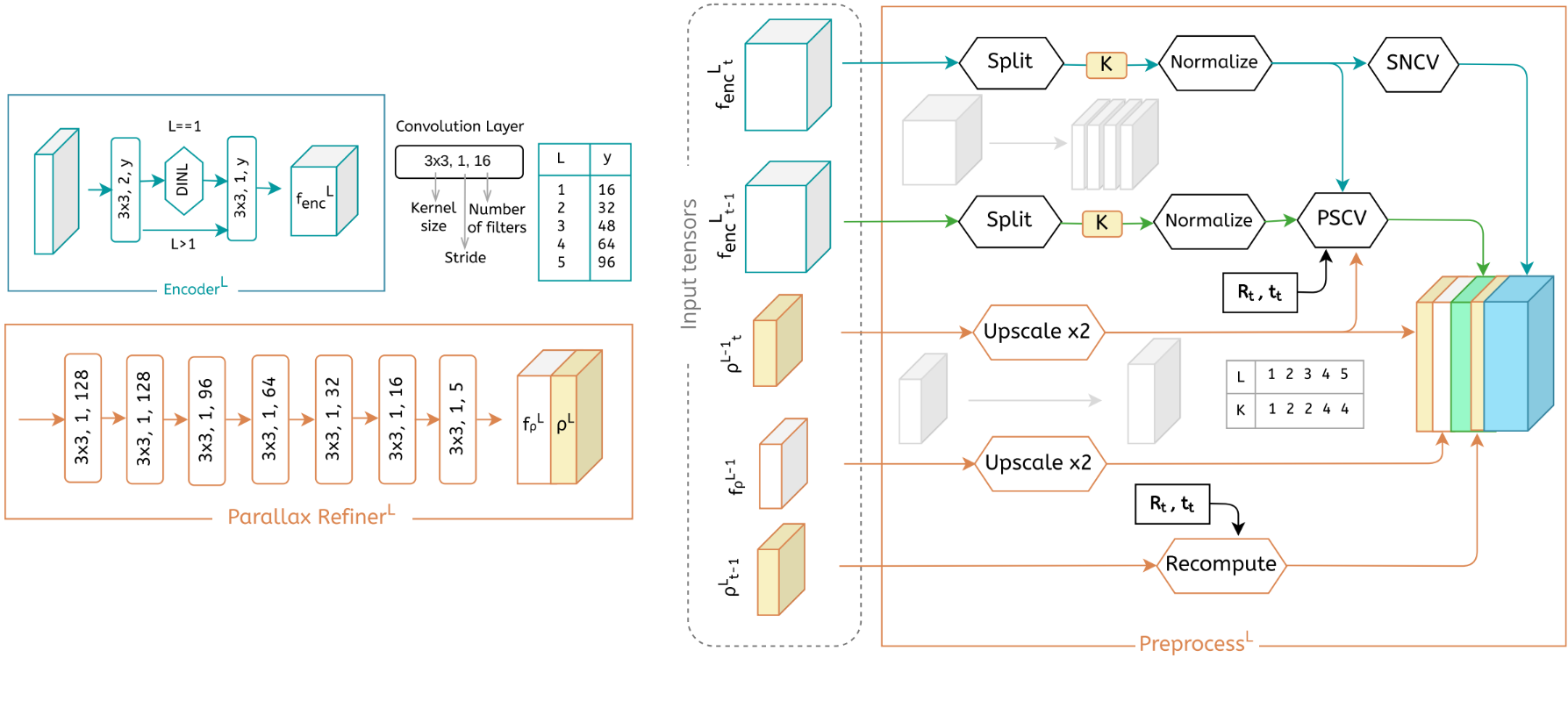} }
    \caption{\centering An illustration of the modules included in M4Depth. The preprocessing unit does not contain any learnable parameters. Feature vectors are subdivided into K sub-vectors using the split layer for subsequent parallel processing. The value of K at each level L is provided. For the encoder, the number of filters y at each level L is provided.}
    \label{fig:modules_depth}%
\end{figure}

Each level of the decoder is composed of a preprocessing unit and a parallax refiner. The preprocessing unit is responsible for preparing the input to the parallax refiner at this level, refer to Figure~\ref{fig:modules_depth} for an overview of the modules included in the preprocessing unit. Specifically, this unit performs the following operations:
\begin{itemize}
    \item \textbf{Upscaling:} It upscales the parallax map \(\rho^{L-1}\) and the parallax features \(f_{\rho}^{L-1}\) estimated from the parallax refiner of the previous level by a multiple of 2 to match the resolution of the current level.
    \item \textbf{Split and Normalize:} the split layer subdivides the feature 3D matrices into K submatrices to decouple the relative importance between them. The normalize layer normalizes the features of each submatrix, allowing to leverage the information embedded in them that have low magnitudes due to the Leaky ReLU activation.
    \item \textbf{Spatial Neighborhood Cost Volume (SNCV):} It describes the two-dimensional spatial autocorrelation of the feature map. Each pixel of the cost volume is assigned the cost of matching the feature vector located in the same location in the feature map with its neighboring feature vectors within a given range, where the cost of matching two vectors $x_1$ and $x_2$ of size N is defined as the correlation:
    \begin{equation}
        cost(x_1,x_2) = \frac{1}{N}x_1^Tx_2
    \end{equation}
    In this way, the network should be invariant to changes in the feature vectors if they lead to the same cost volume. Thus, this makes the network focus more on the spatial structure of the image rather than the values of the features themselves, and, consequently, the network becomes robust and generalizable.
    \item \textbf{Parallax Sweeping Cost Volume (PSCV):} It is computed from two consecutive feature maps $f_t$ and $f_{t-1}$ and a parallax estimate $\rho_t$ coming from the upscaled version of the estimated parallax map of the previous level. Each pixel in the cost volume is assigned the value of matching the feature vector in the same location in $f_t$ with the corresponding feature vector in $f_{t-1}$ after being reprojected to the current time $t$ by the use of $\rho_t$ and a range of values close to it. By searching through candidates in the range around $\rho_t$, it is possible to assess which parallax value at each pixel leads to the best feature matching, and thus it is more likely to be associated with this pixel. 
    \item \textbf{Recompute Layer:} It recomputes the parallax values estimated in the previous time step using the camera transformation matrix and the warping operation~\cite{imagewarp} to provide a first estimate of the parallax values at the current step. Such a first estimate serves as a hint to the parallax refiner. Refer to the Appendix for a detailed explanation of the warping operation.
\end{itemize}

The parallax refiner, ~\ref{fig:modules_depth}, is a stack of 7 convolutional layers responsible for giving an estimate of the parallax map at each level, given as input the preprocessed data generated by the preprocessing module.

 \textbf{Depth loss:} The network is trained in an end-to-end fashion, and a scale-invariant \begin{math}L_1\end{math} loss is used to compute the loss between the predicted depth map \(\hat{d_i}\) and the ground truth \(d_i\). The loss is computed at each decoder level and then accumulated through a weighted sum across all levels, refer to Equation~\ref{eq:lossd}:

 \begin{equation}\label{eq:lossd}
     L_{depth} = \sum_{l=1}^{M}\frac{1}{N_p^l}\sum_{d_i^l} 2^{l+1}|log(d_{i})-log(\hat{d_{i}})|
 \end{equation}

 where \(M\) is the number of decoder levels and \(N_{p}^l\) is the total number of pixels in the image at level \(l\).

\subsection{M4Semantic Network}

\begin{figure}[h!]%
    \centering
    {\includegraphics[width=0.9\linewidth]{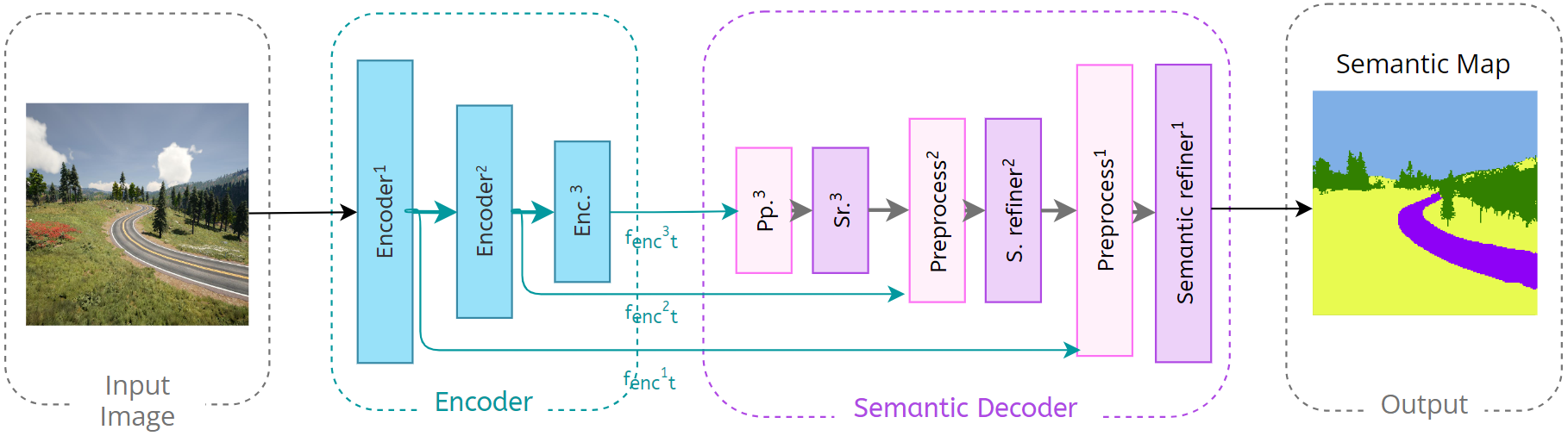} }
    \caption{\centering Our M4Semantic Architecture. It is composed of an encoder and a decoder module with a pyramidal structure. Each level of the decoder is composed of a preprocessing unit and a semantic refiner. 
    }%
    \label{fig:semantic}%
\end{figure}

Inspired by M4Depth, we propose a similar but simpler architecture for semantic segmentation depicted in Figure~\ref{fig:semantic}. Similar to M4Depth, our architecture is composed of an encoder and a decoder. The encoder is a stack of multiple levels and has a pyramidal structure where the resolution of the feature map is decreased while proceeding forward through the levels. The feature map predicted at each level is passed to its corresponding decoder level. Each decoder level is composed of a preprocessing unit and a semantic refiner in the place of the parallax refiner in M4Depth. The preprocessing unit prepares the input to the semantic refiner, and the semantic refiner at each level gives an estimate of the semantic segmentation map at a specific resolution. The resolution of the semantic map is scaled-up proceeding forward through the decoder levels.

\begin{figure}[h!]%
    \centering
    {\includegraphics[width=0.8\linewidth]{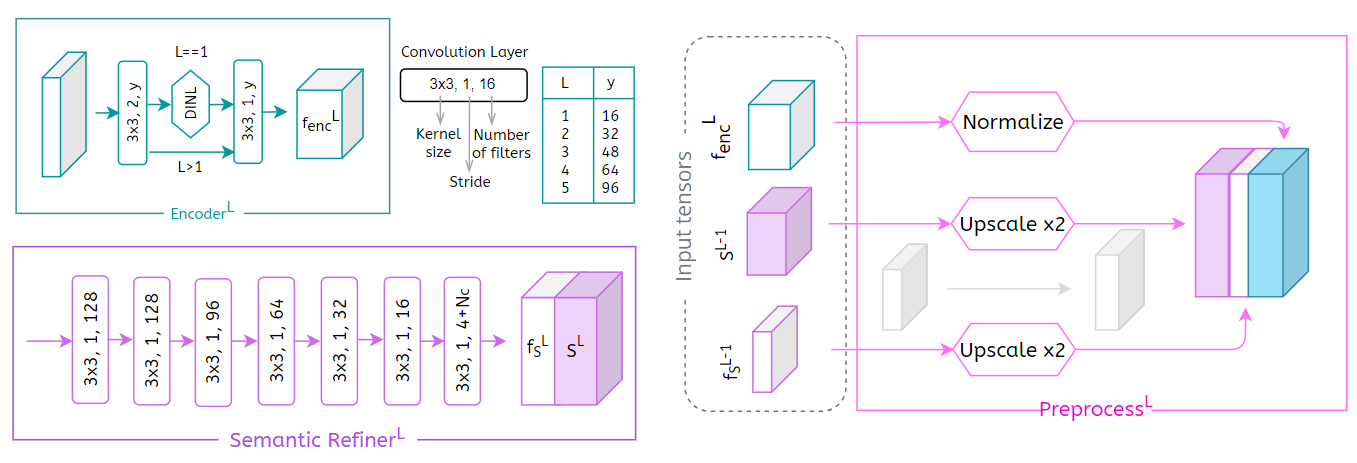} }
    \caption{\centering An illustration of the modules in our M4Semantic architecture. \(N_c\) is the number of semantic classes}%
    \label{fig:modules}%
\end{figure}

 In Figure~\ref{fig:modules} we show the modules used in our architecture. It can be noted that the modules in M4Semantic are less than the ones used in M4Depth. The encoder at each level is composed of 2 convolutional layers. In the first level, DINL~\cite{dinl} is added after the first convolution to increase the network's robustness to varied colors and luminosity conditions. ReLU activation is applied after each convolutional layer, and the resolution is decreased by a factor of 2 after each level. The pyramidal structure of the encoder helps in extracting both coarse and fine (global and local) features from the input image.

The preprocessing unit at each decoder level is a pure computational unit with no parameters to be trained. It performs two operations:
\begin{itemize}
    \item It upscales the semantic map \(S^{L-1}\) and the semantic features \(f_{S}^{L-1}\) estimated from the semantic refiner of the previous level by a multiple of 2 to match the resolution of the current level.
    \item It normalizes the feature map \(f_{enc}^L\) received from the encoder. This allows to leverage the information embedded in the feature map that has low magnitudes due to the Leaky ReLU activation applied after each layer of the encoder.
\end{itemize}


Similar to the parallax refiner, the semantic refiner at each level is composed of a stack of convolutional layers. The last convolutional layer has a depth of 4 (depth of the semantic features map) + N (the number of semantic classes). The output of the semantic refiner is a predicted semantic features map and an estimated semantic segmentation map. We apply Softmax activation on the predicted segmentation map to obtain a probability score for each class on every pixel.

Different from M4Depth, our M4Semantic architecture works on single images. We removed the time dependency in the semantic decoder because this produced 2 times faster results than the one with time dependency, with a slight drop in accuracy. Refer to the Architecture Study in section~\ref{ablation} for further clarification.   

\textbf{Semantic loss:} The standard categorical cross-entropy loss is used on the predicted semantic maps at each level. The ground truths are resized using Nearest Neighbor interpolation to match the resolution of the predicted semantic maps at intermediate levels. Then, the losses are aggregated through a weighted sum, as shown in Equation~\ref{eq:losss}.

\begin{equation}\label{eq:losss}
    L_{semantic} = \sum_{l=1}^{M}\frac{1}{N_p^l}\sum_{p_t^l} -log({p_{t}})
\end{equation}
where \(M\) is the number of decoder levels, \(N_{p}^l\) is the total number of pixels in the image at level \(l\), and \(p_{t}\) is the softmax score for the target class.

\subsection{Joint Co-SemDepth Network}
\begin{figure*}[h!]%
    \centering
    {\includegraphics[width=\linewidth]{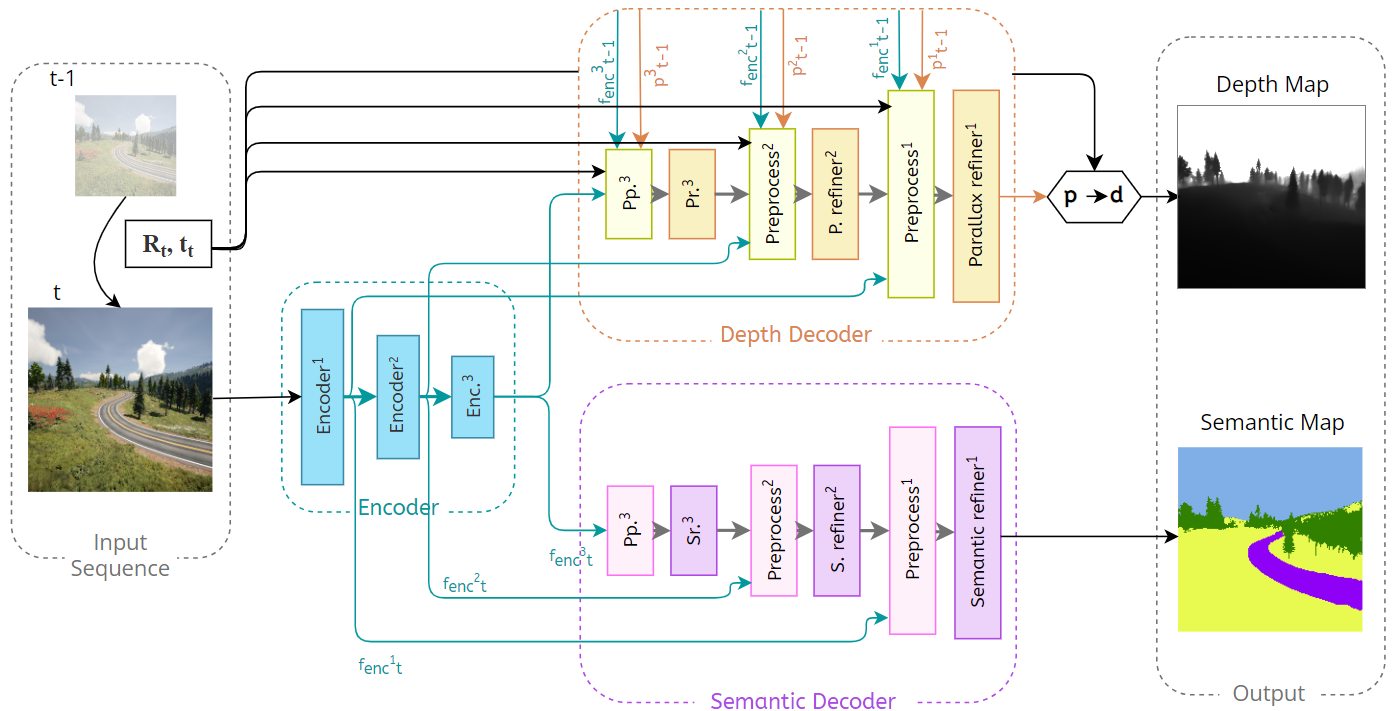} }
    \caption{\centering Our proposed joint architecture Co-SemDepth. It predicts both depth and semantic segmentation maps. It is composed of a shared encoder and two decoders. The encoder and the depth decoder are the same presented in~\cite{m4depth}. The semantic decoder makes use of the encoded feature maps to give an estimate of the semantic segmentation map. The depth and semantic maps get scaled up as they go forward through the successive levels of the decoders. The number of shown levels here is 3, while in our experiments, we use 5 levels. Depth map is shown in gray scale}%
    \label{fig:joint}%
\end{figure*}

To merge the two previously described networks, M4Depth and M4Semantic, we adopt a multi-tasking shared encoder architecture~\cite{cao2016exploiting, mousavian2016joint, nekrasov2019real, zhang2018joint, he2021sosd}. The depth estimation and semantic segmentation networks share the encoder part for feature extraction, but each of them has its own decoder for their corresponding map prediction. An overview of our joint architecture is in Figure~\ref{fig:joint}. 

\textbf{Loss Function:} Our joint network is trained in an end-to-end fashion. The loss function for our architecture is defined as the weighted summation of the depth and semantic segmentation losses: 
\begin{equation}\label{eq:loss_tot}
    L_{total} = L_{depth} + w * L_{semantic}
\end{equation}
where $w$ is a weighting factor whose value was set after experimentation to $0.15$. A discussion on the choice of $w$ is reported in Section~\ref{cosemdepth_implement}.

\section{Experiments Setup}

\subsection{Datasets}
It was stated before that there is a limited availability of annotated real datasets in the aerial domain compared to other application domains. In particular, they often lack appropriate ground truth for joint depth estimation and semantic segmentation tasks. For this reason, we conduct our experiments on the joint architecture using the synthetic MidAir~\cite{midair} dataset, and we use the real dataset AeroScapes~\cite{aeroscapes}, which contains only semantic segmentation annotation, for the validation of our semantic segmentation network. See Figure~\ref{fig:midair} and Figure~\ref{fig:aeroscapes} for sample images and annotations from both datasets.

\begin{figure*}[h!]%
    \centering
    {\includegraphics[width=0.7\linewidth]{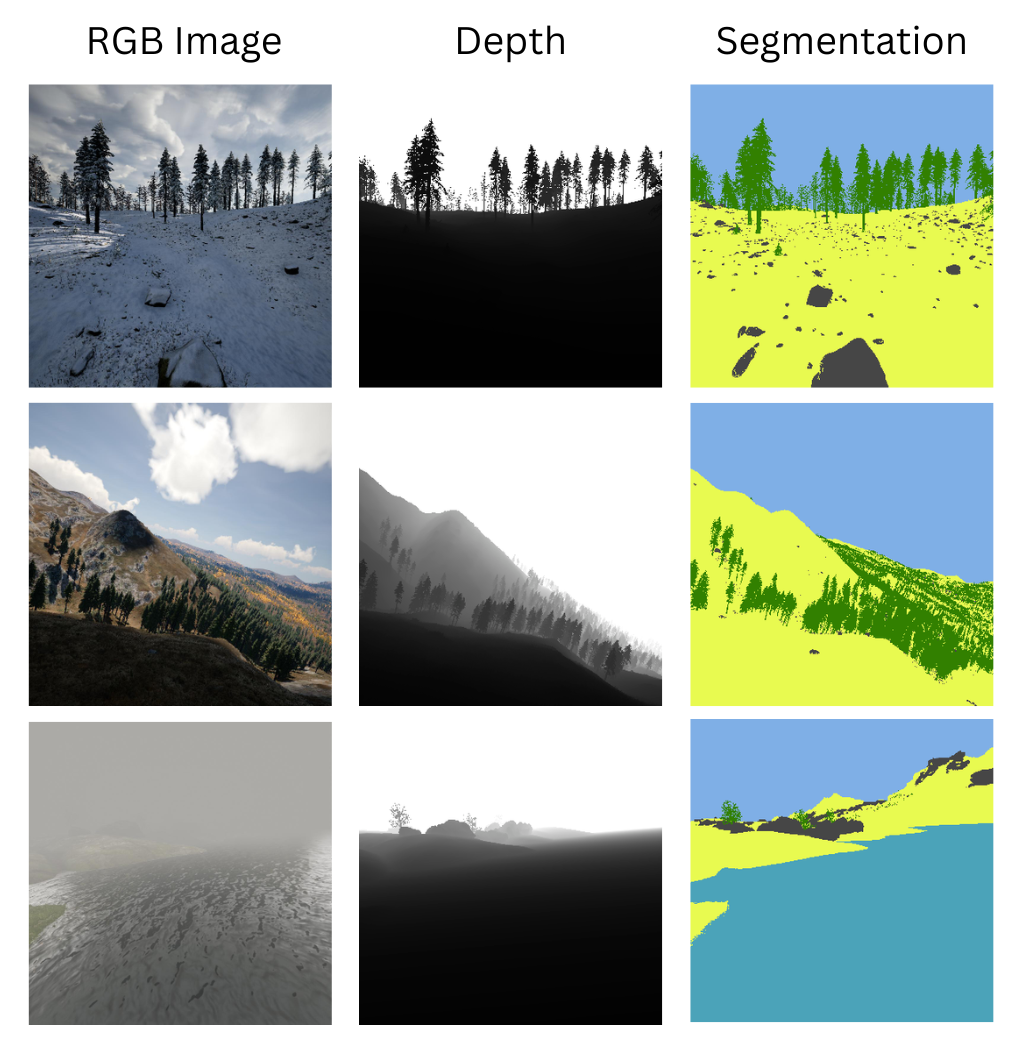} }
    \caption{\centering Sample images and annotations from the MidAir dataset}%
    \label{fig:midair}%
    
\end{figure*}
\begin{figure*}[h!]%
    \centering
    {\includegraphics[width=0.7\linewidth]{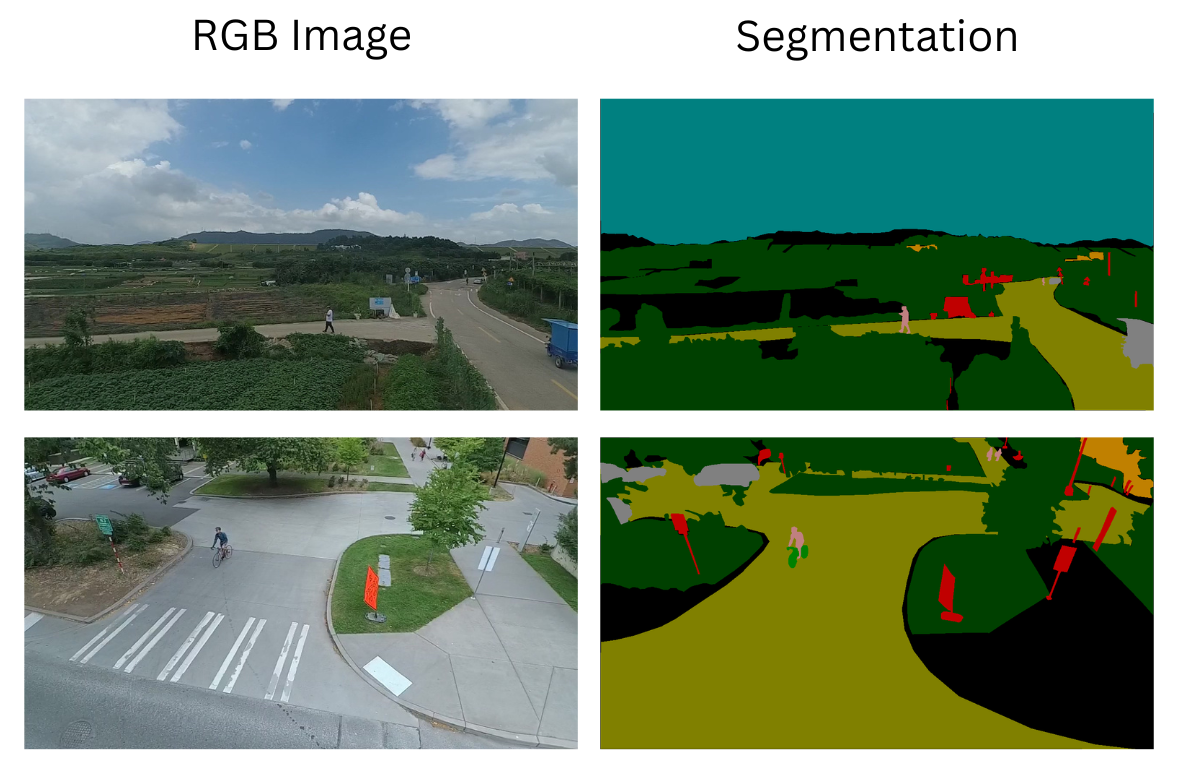} }
    \caption{\centering Sample images and annotations from the AeroScapes dataset}%
    \label{fig:aeroscapes}%
\end{figure*}

{\bf MidAir}~\cite{midair} is a synthetic dataset collected using the AirSim simulator~\cite{airsim}, consisting of 420K forward-view RGB video frames captured at low altitude in outdoor unstructured environments with various weather conditions. It contains annotations of depth maps, semantic segmentation, surface normals, stereo disparity, and camera locations. Hence, this dataset is suitable for training and testing our joint Co-SemDepth architecture.

We adopt the train-test split used in~\cite{m4depth}, but we select 8 trajectories that cover a variety of conditions to create validation data, and we provide the data split on our \hyperlink{https://github.com/}{Github page}. In the evaluation, the depth values are capped at 80.0 meters. We resize images to a resolution of 384x384. 
In the original semantic annotation of MidAir, there are 14 semantic classes: Sky, Animals, Trees, Dirt Ground, Ground Vegetation, Rocky Ground, Boulders, Empty, Water, Man-Made Construction, Road, Train Track, Road Sign, and Others. Since several classes are visually indistinguishable and some of them are very small, we map them to a smaller set of 7 semantic classes: Sky, Water, Land, Trees, Boulders, Road, and Others. Specifically, we considered Ground Vegetation, Rocky Ground, and Dirt Ground as {Land}, and Animals, Empty, Train Track, and Road Sign in {Others}.


{\bf Aeroscapes}~\cite{aeroscapes} is a real dataset collected using drones at low-mid altitude in various outdoor environments. It consists of 3,269 images with an 80\%-20\% train-test split and a resolution of 1280x720. 
This dataset contains only semantic segmentation annotation. For this reason, we can not use it for the training of our joint architecture. However, we use Aeroscapes for the training and testing of the individual M4Semantic network. 

\subsection{Implementation Details}
\label{cosemdepth_implement}
We adopt Adam optimizer with the default momentum parameters \((\beta_1 = 0.9, \beta_2 = 0.999)\) and a fixed learning rate of \(10^{-4}\). 
We apply image augmentation of random rotation, flipping, and changing color (contrast, brightness, hue, and saturation) during training, and we train with a batch size of 3 and a number of epochs $=60$. 
After training, we choose the checkpoint that produced the best validation results for evaluation on the test set.

Our workstation has 16GB RAM, an Intel Core i7 processor, and a single NVIDIA Quadro P5000 GPU card running CUDA11.4 with CuDNN 7.6.5 and Ubuntu OS. Due to its memory-limited resources, depth and semantic maps are predicted at a resolution equal to half the input resolution, and then Nearest Neighbour interpolation is applied on the output maps to scale up their resolution to the original size. As reported in~\cite{chen2018driving}, decreasing the image resolution can slightly decrease the accuracy; however, it gives the advantage of reducing the computational runtime and memory footprint. 


To quantitatively evaluate the depth prediction results, we consider the commonly used evaluation metrics in prior works~\cite{m4depth, manydepth, monodepth2}. These include the linear root mean square error (RMSE) (Equation~\ref{eq:rmse}), the absolute relative error (Equation~\ref{eq:abs}), and accuracy under a threshold (Equation~\ref{eq:delta}). 
\begin{equation}
\label{eq:rmse}
    RMSE = \sqrt{\frac{1}{N_{p}}\sum_{i=1}^{N_{p}} | \hat{d_i} - d_i | }
\end{equation}
\begin{equation}
\label{eq:abs}
    AbsRelErr = \sqrt{\frac{1}{N_{p}}\sum_{i=1}^{N_{p}} | \hat{d_i} - d_i | }
\end{equation}
%
\begin{equation}
\label{eq:delta}
    max(\frac{\hat{d_i}}{d_i},\frac{d_i}{\hat{d_i}}) = \delta < T
\end{equation}
%
where $\hat{d_i}$ is the estimated depth, $d_i$ is the ground truth, $N_p$ is the number of pixels in the image, and \(T\) is the set thresholds: \((1.25, 1.25^2, 1.25^3)\). 

For semantic segmentation, we adopt the commonly used mean Intersection over Union \(mIoU\) metric (Equation~\ref{eq:miou}). 
%
\begin{equation}
\label{eq:miou}
    mIoU = \frac{TP}{TP + FP + FN}
\end{equation}
%
where TP, FP, and FN are, respectively, the number of true positives, false positives, and false negatives at the pixel level.
The Inference Time (Inf. Time) is computed in milliseconds per frame (ms/f).

\begin{figure}[h!]%
    \centering
    {\includegraphics[width=\linewidth]{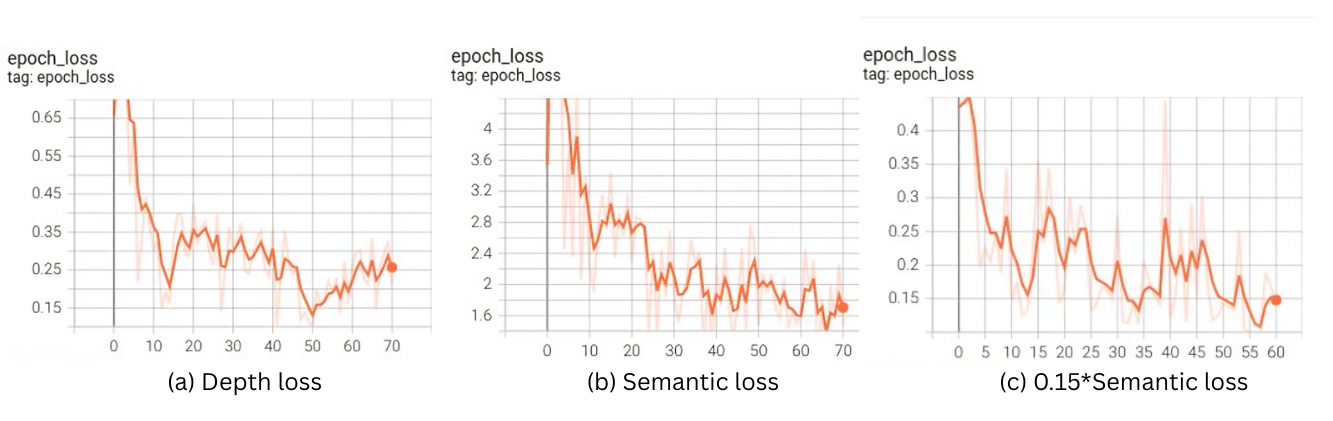} }
    \caption{\centering Loss curves showing the difference in the loss range of values for the two tasks during training. After multiplying the semantic loss (b) by a weighting factor of 0.15 (c), its range of values became comparable to the depth loss (a).}%
    \label{fig:loss}%
\end{figure}

In Figure~\ref{fig:loss}, the loss curves for M4Depth and M4Semantic networks are shown. The curves show the difference in the range of loss values between the two tasks due to the difference in the loss function used: L1 loss and categorical cross-entropy loss. We incorporate a weighting factor  \( w=0.15\) to force the loss values for semantic to lie within the same range of losses for depth, and thus ensure a comparable contribution for the two losses during training of the joint model.

\section{Results}
In this section, we discuss the experiments we conducted to validate the effectiveness of our joint architecture on various datasets. First, we evaluate the effectiveness of using our joint architecture compared to using the two single architectures. Then, we benchmark our model against other state-of-the-art single and joint methods. Finally, we make a comparative study of architectural design alternatives. Code is  available on our Github repository: \url{https://github.com/Malga-Vision/Co-SemDepth}

\subsection{Joint vs Single Architectures}
We conduct experiments to compare the performance of our joint architecture Co-SemDepth with the two single ones: M4Depth and M4Semantic. Each architecture is trained equally for 60 epochs. The results are reported in Table~\ref{table_joint_vs_single}. We can notice that the accuracy values (in terms of depth and semantic metrics) of the joint architecture are close to the single ones. Depth is a bit better using Co-SemDepth, and this can signify that adding the shared encoder that extracts features related to both depth estimation and semantic segmentation has helped to enhance the results of depth estimation, probably because it has enriched the features entering the parallax decoder part with features related to semantic segmentation. While for semantic segmentation, the results using M4Semantic ($mIoU= 76.8\%$) are a bit better than using the joint architecture $mIoU=75.44\%$. From this, we can say that the shared encoder in the joint architecture did not enhance the results of semantic segmentation; instead, using a dedicated encoder for extracting solely the semantic segmentation features led to a better accuracy. 

While it was found in other works~\cite{cao2016exploiting} that depth features can enhance the semantic segmentation accuracy, it was mentioned in~\cite{j7} that multi-tasking architectures can perform poorly compared to dedicated single ones. We thereby think that this enhancement or degradation can differ depending on the dataset used and its semantic classes. 

The inference time of the joint architecture (49.6 ms/f) is \emph{lower than} the sum of m4Semantic and m4Depth (\(44.9 + 9.8\) ms/f). Moreover, the number of parameters of the joint architecture (5.2 Million) is less than the sum of the two single ones (3.06 + 2.61 Million) by around 500K parameters. 

The above signifies that using our joint architecture Co-SemDepth is more effective in terms of computational time and memory footprint than using the two single architectures while achieving very close accuracies. The trade-off between accuracy and computational cost in the multi-tasking architectures was previously discussed in~\cite{j7}, and it can differ depending on the environment. The high inference time of the depth branch compared to the segmentation branch can be due to the added computations of parallax in the depth branch and the additional modules for computing the cost volumes SNCV and PSCV, which are not present in the segmentation branch. 

During inference, Co-SemDepth required only \emph{6.2GB of GPU memory} while running M4Depth and M4Semantic together required 14.6GB of GPU memory. This makes Co-SemDepth compatible to run on microcontrollers that have only 8GB RAM and that are widely used in robotics hardware due to their affordable cost.



\begin{table*}[h!]
\centering
\caption{\centering Evaluation of our joint vs single architectures for depth estimation and semantic segmentation on the MidAir dataset. }
\label{table_joint_vs_single}
\resizebox{\linewidth}{!}{
\begin{tabular}{ c || c | c | c | c | c | c | c | c | c }
\toprule
 \multirow{2}{*}{Architecture}& \multirow{2}{*}{Output}  & \multirow{2}{*}{Params(M)} & \multirow{2}{*}{Inf. Time (ms/f)}   & Semantic & \multicolumn{5}{|c}{Depth}\\ \cline{5-10}
  & & & & mIoU $\uparrow$ & RMSE $\downarrow$ & RelErr $\downarrow$ & \(\delta1\) $\uparrow$ &  \(\delta2\) $\uparrow$ &  \(\delta3\) $\uparrow$\\
\midrule
M4Depth & D & 3.06 & 44.9  & - & 6.99 & 0.109 & 92.0\% & 95.4\% & 97.0\%\\ 
\hline
M4Semantic & S & 2.61 & 9.8 & \textbf{76.80}\% & - &  - & - & - & -\\
\hline
\textbf{Co-SemDepth} & D+S & 5.2 & 49.6  & 75.44\% & \textbf{6.70} & \textbf{0.096} &  \textbf{92.3}\% & \textbf{95.7}\% & \textbf{97.2}\%\\
\bottomrule 
\end{tabular}
}

\end{table*}

\subsection{Benchmarking}
{\bf Benchmarking on MidAir: }

We compare the performance of Co-SemDepth with other open-source state-of-the-art methods. We compare it with both single and joint architecture methods. Table~\ref{table_baseline} summarizes the training parameters used for each method. 
For each method, we fix the input image size to 384x384 and the number of training epochs to 60 or a maximum of 80k iterations (for Segformer and TaskPrompter). Maximum depth is set to 80 meters.
Other parameters are kept as the default. 
\begin{itemize}
    \item For FCN, we implement FCN-32S~\cite{fcn32s} and we use two backbone networks; VGG16~\cite{vgg16} and MobileNetV2~\cite{mobilenetv2}.
    \item For SegFormer~\cite{xie2021segformer}, SegFormer-B0 is used. It should be noted that we had difficulties training and testing higher versions of SegFormer on our server due to their large size and complexity.
    \item For TaskPrompter~\cite{taskprompter}, the vision transformer Base model was set as the backbone with an embedding dimension of 384 and a number of channel heads equal to 8. All the other values were kept as the default.
    \item For DepthAnything~\cite{depth_anything_v2}, we use the small model of DepthAnything-V2 because it has the lowest number of parameters and to be compatible to run on our machine. The trained model dedicated to outdoors, "Outdoor Virtual KITTI2", was selected, and zero-shot prediction was performed.
\end{itemize}



\begin{table}[H]
\begin{center}
\caption{\centering Training parameters used for benchmarking the baseline methods on MidAir.}
\label{table_baseline}
\resizebox{0.8\linewidth}{!}{
\begin{tabular}{ c || c | c | c | c | c | c }
\hline
 Method  & optimizer & lr & sched wt decay & max iter & epochs & batch \\
\hline
 FCN  & Adam & $1\times10^{-4}$ & - & - & 60 &  4\\
 ERFNet  & Adam & $5\times10^{-4}$ & $1\times10^{-4}$ & - & 60 & 6\\
 SegFormer-B0  & Adam & $6\times10^{-5}$ & $1\times10^{-2}$ & 80K & - & 4\\
 RefineNet  & Adam & $1\times10^{-4}$ & $1\times10^{-4}$ & - & 60 & 8\\
TaskPrompter  & Adam & $1\times10^{-5}$ & $1\times10^{-6}$ & 80K & - & 3 \\
\hline

\end{tabular}
}
\end{center}

\end{table}

\begin{table*}[h!]
\centering
\setlength{\tabcolsep}{1mm}
\caption{\centering Benchmarking Co-SemDepth on MidAir against other state-of-the-art methods in both depth estimation (D) and semantic segmentation (S). The top part reports single depth estimation methods, the middle part for single segmentation methods, and the bottom part for joint methods. $^*$ means the depth metrics values were reported in~\cite{m4depth}. $^{@}$ means zero-shot prediction.}
\label{tab:table_benchmark}
\resizebox{\linewidth}{!}{
\begin{tabular}{ c || c | c | c | c | c | c | c | c | c }
\toprule
 \multirow{2}{*}{Method} & \multirow{2}{*}{Output} & \multirow{2}{*}{Params(M)} & \multirow{2}{*}{Inf. Time (ms/f)$\downarrow$}  & Semantic & \multicolumn{5}{|c}{Depth}\\ \cline{5-10}
  & & & & mIoU $\uparrow$ & RMSE $\downarrow$ & RelErr $\downarrow$ & \(\delta1\) $\uparrow$ &  \(\delta2\) $\uparrow$ &  \(\delta3\) $\uparrow$\\
\midrule
MonoDepth2$^*$ & D & 14.8 & 23.9 & - & 12.35 & 0.394 & 61.0\% & 75.1\% & 83.3\%\\ 
ST-CLSTM$^*$ & D & 15.04 & 35.3 & - & 13.69 & 0.404 & 75.1\% & 86.5\% & 91.1\%\\ 
ManyDepth$^*$ & D & 46.3 & 82.9 & - & 10.92 & 0.203 & 72.3\% & 87.6\% & 93.3\%\\ 
PWCDC-Net$^*$ & D & 9.4 & 25.8 & - & 8.35 & 0.095 & 88.7\% & 93.8\% & 96.2\%\\ 
DepthAnythingV2$^@$ & D & 24.8 & 75.2 & - & 33.37 & 0.640 & 12.3\% & 25.6\% & 39.7\%\\ 
\hline
FCN(VGG16)  & S & 14.7 & 58.3 & 72.93\% & - & - & - & - & - \\ 
FCN(MobileNetv2)  & S & 2.2 & 60.5 & 69.82\% & - & - & - & - & - \\ 
ERFNet & S & 2.07 & 19.1 & 77.40\% & - & - & - & - & - \\ 
SegFormer-B0 & S & 3.8 & 49.1 & 75.10\% & - & - & - & - & - \\ 
\hline
RefineNet & D+S & 3.0 & 74.2 & 72.70\% & 9.74 & 0.200 & 74.9\% &  89.0\% &  94.5\%\\ %
TaskPrompter & D+S &  126 & 120.6 &  \textbf{80.20\%} &  9.80 &  0.250 &  50.0\% &  78.3\% &   90.0\%\\

\textbf{Co-SemDepth} & D+S & \textbf{ 5.2} & \textbf{49.6}  & 75.44\% & \textbf{6.70} & \textbf{0.096} &  \textbf{92.3}\% &  \textbf{95.7}\% &  \textbf{97.2}\%\\ 
\bottomrule
\end{tabular}
}

\end{table*}


The benchmarking results are reported in Table~\ref{tab:table_benchmark}. From the table, we can clearly notice that our method outperforms the other joint networks in depth accuracy and inference time. While its semantic segmentation accuracy is lower than TaskPrompter, Co-SemDepth has a notably lower inference time and model size, making it more convenient for hardware deployment. The inference time of the network can be further enhanced by converting the model to ONNX\footnote{https://onnx.ai/onnx/intro/converters.html} or using tensor RT\footnote{https://docs.nvidia.com/deeplearning/tensorrt/latest/getting-started/quick-start-guide.html}.

Compared to the single depth estimation networks, Co-SemDepth could maintain its superior accuracy in depth estimation that was reported in~\cite{m4depth}. This indicates that transforming M4Depth to the joint Co-SemDepth did not have a negative effect on its depth estimation performance compared to the state-of-the-art. In addition, Co-SemDepth has a notably smaller number of parameters compared to the other single depth estimation methods. 

For the single semantic segmentation networks, Co-SemDepth has a competitive mIoU with the others, only slightly inferior to ERFNet, which is, in any case, a dedicated architecture, not a joint one. 

For further analysis of semantic segmentation performance, we compute the per-class IoU for each method. The per-class IoU evaluation can be found in Table~\ref{table_per_class}.
It can be noted that the high mIoU of TaskPrompter is mostly caused by its good detection of the class "Others" ($20\%$ better than other methods). This can be justified by the fact that the "Others" class contains a variety of objects (train track, road sign, animals, and others), and such variety requires a model of high capacity to learn it well (TaskPrompter has 126M parameters).

\begin{table*}[h]
\begin{center}
\caption{\centering Per-Class IoU Evaluation of Co-SemDepth architecture and other baseline methods on MidAir.}
\label{table_per_class}
\resizebox{\linewidth}{!}{
\begin{tabular}{ c || c | c | c | c | c | c | c || c }
\hline
 Method & Sky & Water & Trees & Land &  Boulders & Road & Others & \textbf{mIoU} \\
\hline
FCN(VGG16)~\cite{fcn} &  88.56\% &  83.12\% &  75.50\% &  82.30\% &  29.97\% &  88.04\% &  54.60\% &  72.93\% \\
FCN(MobileNetv2)~\cite{fcn} &  87.82\% &  82.42\% &  73.42\% &  81.28\% &  26.57\% &  84.64\% &  46.09\% &  69.82\% \\
ERFNet~\cite{romera2017erfnet} &  91.50\% &  87.64\% &  82.48\% &  85.10\% &  40.63\% &  90.90\% &  63.70\% &  77.42\% \\
SegFormer-B0~\cite{xie2021segformer} &  90.54\% &  88.58\% &  79.70\% &  83.33\% &  30.13\% &  92.19\% &  61.20\% &  75.10\% \\
\hline
RefineNet~\cite{nekrasov2019real} &  89.70\% &  81.60\% &  79.70\% &  82.20\% &  30.15\% &  91.36\% &  54.10\% &  72.69\% \\
TaskPrompter~\cite{taskprompter} &  90.60\% &  87.10\% &  79.10\% &  85.30\% &  41.00\% &  95.30\% &  83.20\% &  \textbf{80.23}\% \\
\textbf{Co-SemDepth} &  90.10\% &  86.40\% &  79.95\% &  82.60\% &  33.10\% &  94.60\% &  59.25\% &  75.44\%\\
\hline
\end{tabular}
}
\end{center}
\end{table*}

\begin{figure*}%
    \centering
    {\includegraphics[width=\linewidth]{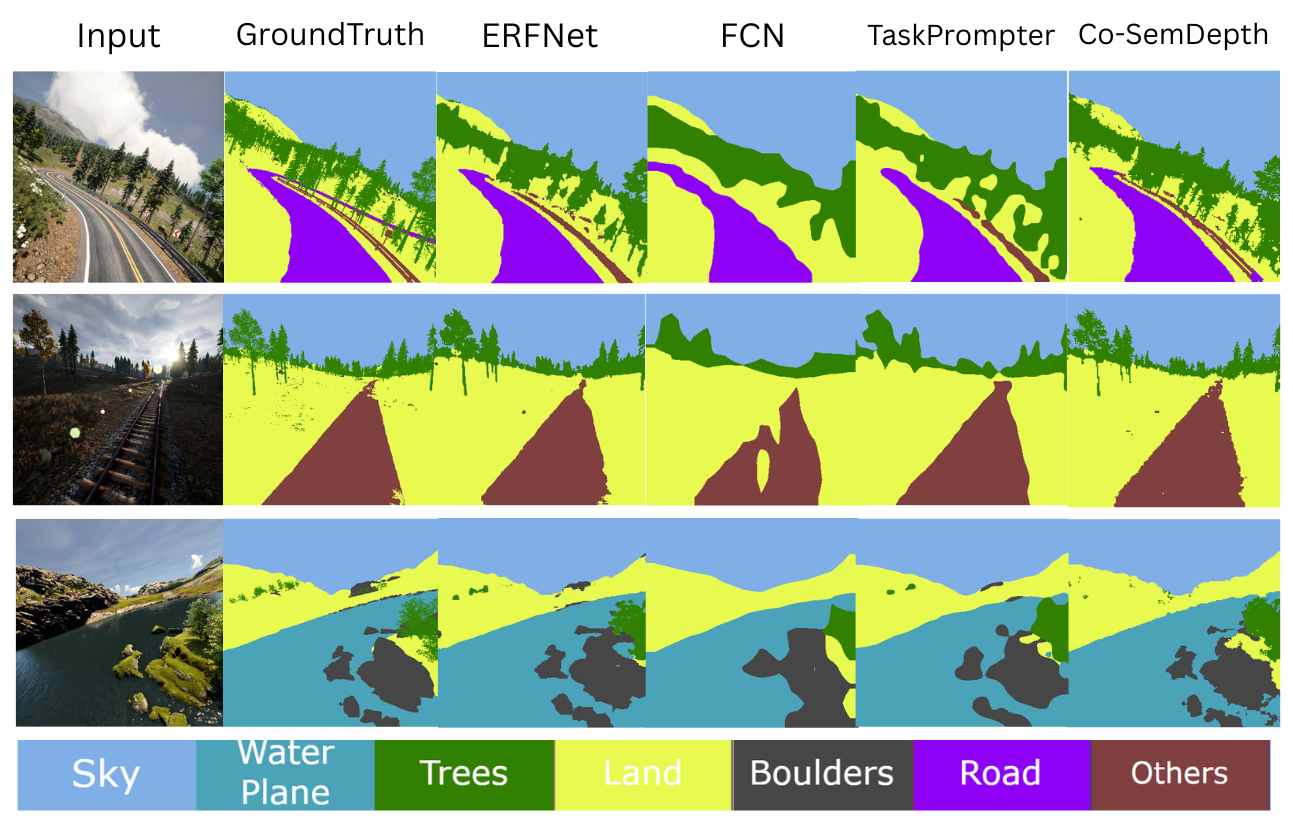} }
    \caption{\centering Qualitative evaluation of the semantic map predictions of ERFNet, FCN MobileNet, TaskPrompter, and Co-SemDepth respectively, on sample images from MidAir dataset.}%
    \label{fig:qualit}%
\end{figure*}

\textbf{Visualization:} The qualitative visualization of the depth map predictions of M4Depth and other methods on MidAir was already done in~\cite{m4depth}. Here, we do a qualitative visualization of the semantic map predictions of the different methods. The outputs can be found in Figure~\ref{fig:qualit}. We can notice that all networks could capture the overall semantic layout of the input images; the location of trees, roads, water, land, and sky. However, Co-SemDepth and ERFNet are remarkably better in capturing the details (notice the trees in the first row and the train track in the second row). Compared to Co-SemDepth, ERFNet and TaskPrompter are better in capturing some of the faraway objects (for example, the distant boulders and bushes in the third row). While TaskPrompter has a higher $mIoU$ than Co-SemDepth, Co-SemDepth is notably better at recognizing the details of the trees, road, and rocks than TaskPrompter. 


\textbf{Generalization:} We conduct some experiments for assessing the generalization capability of the network. The datasets TartanAir~\cite{tartanair} and WildUAV~\cite{wilduav} were used for testing. \textit{TartanAir} is a synthetic dataset captured using AirSim in a multitude of outdoor and indoor environments with depth and mesh segmentation annotations. It also includes camera pose information. For this reason, it could be used for testing Co-SemDepth generalization. However, the evaluation was done only qualitatively due to the absence of per-class segmentation in the dataset. The trajectories of the two environments: Gascola and SeasonForestWinter were used in the evaluation due to their similarity with the scenes of MidAir. On the other hand, \textit{WildUAV} is a real dataset captured in an outdoor rural environment. It contains both depth and per-class semantic segmentation annotations, and it includes camera pose information. Therefore, it could be used to test the generalization capability of Co-SemDepth both quantitatively and qualitatively. All the trajectories of WildUAV were used due to their similarity with MidAir environments.

\begin{figure}[h!]%
    \centering
    {\includegraphics[width=\linewidth]{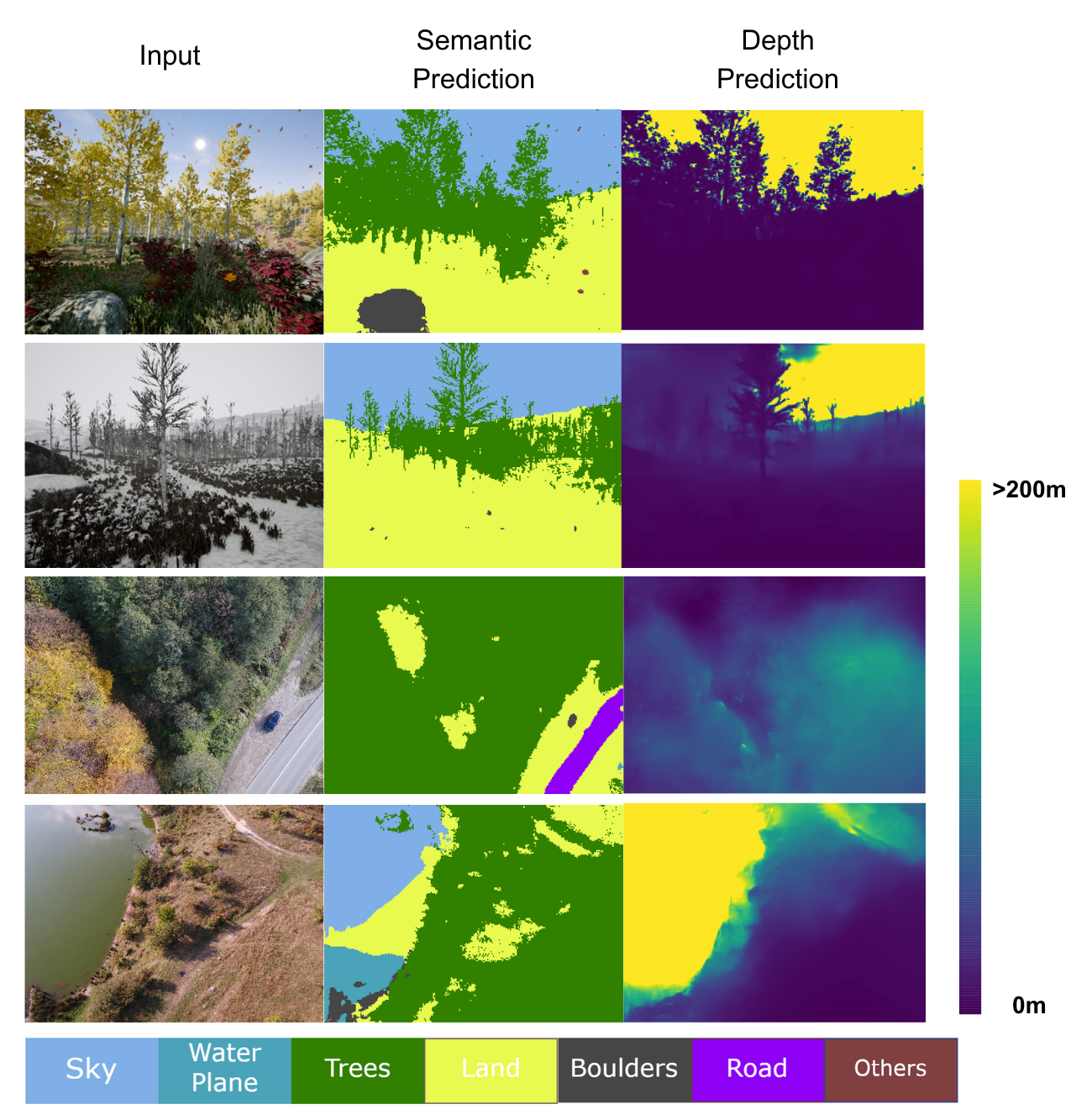} }
    \caption{\centering Qualitative evaluation of the zero-shot generalization performance of MidAir-trained Co-SemDepth on sample images from TartanAir (top and 2nd row) and WildUAV (3rd and bottom rows).}%
    \label{fig:generalqualit}%
\end{figure}

In Figure~\ref{fig:generalqualit}, a visualization of the prediction on sample images from the synthetic TartanAir~\cite{tartanair} and the real WildUAV~\cite{wilduav} datasets using Co-SemDepth (trained on MidAir) is shown. 

The semantic segmentation prediction of TartanAir images went well, and trees, land, rocks, and sky were correctly segmented. However, for depth estimation of the first image, there was no big difference in the depth values predicted for trees and land. It is apparent that the network was able to predict the depth of the sky correctly, but it was unable to distinguish the relative depth of the trees and land, and they were all predicted with very close values. For depth estimation of the second image, the network was able to distinguish between the depths of the close and far trees and the land. However, due to the strange color of the sky in the winter, the network predicted part of the sky wrongly with near depth values.

In WildUAV, in the third image, the semantic segmentation was overall correct except that the vehicle was detected as a rock. The depth estimation of the third image could distinguish between the relatively far distance of the road compared to the tree area. In the fourth image, there is an apparent confusion between Water and Sky in the semantic segmentation. This can be due to the visual similarity of these two classes. In addition, the training data contained only forward camera-view images where the sky is always present in the top part of the images, while in WildUAV, the images are captured from a nadir view where the sky is not present in the images. For this reason, it can be noticed that the network predicted the top part of the lake as Sky while the bottom part was predicted correctly as Water. The depth of the water area was predicted with a far distance in the depth output, as the water was treated wrongly as sky. 

Quantitatively, the generalization of the model is assessed on WildUAV. Co-SemDepth was trained on the synthetic MidAir+TopAir datasets and then tested on WildUAV. The 16 semantic classes of wildUAV were mapped to our set of 7 classes as the following: sky $\rightarrow$ sky, water $\rightarrow$ water, \{deciduous trees, coniferous trees, fallen trees\}$\rightarrow$ trees, \{dirt ground, ground vegetation\}$\rightarrow$land, rocks $\rightarrow$ rocks, \{road, sidewalk\}$\rightarrow$ road, \{static car, moving car, building, fence, people, empty\}$\rightarrow$ others. 

The depth estimation results can be found in Table~\ref{table_depth}, and the semantic segmentation results in Table~\ref{table_sem}. While the depth estimation results are acceptable, the semantic segmentation results are only acceptable for trees and land classes.

\begin{table}[h]
\begin{center}
\caption{\centering Depth estimation results on WildUAV obtained using Co-SemDepth tained on MidAir+TopAir}
\label{table_depth}
\resizebox{0.6\linewidth}{!}{
\begin{tabular}{ c | c | c | c | c }
\hline
 RMSE $\downarrow$ & AbsRelErr $\downarrow$ & \(\delta1\) $\uparrow$ &  \(\delta2\) $\uparrow$ &  \(\delta3\) $\uparrow$\\
\hline
14.9 & 0.303 & 58.7\% & 80.7\% & 91.5\%\\  
\hline
\end{tabular}
}
\end{center}
\end{table}

\begin{table}[h]
\begin{center}
\caption{\centering Semantic segmentation results on WildUAV obtained using Co-SemDepth tained on MidAir+TopAir}
\label{table_sem}
\resizebox{0.8\linewidth}{!}{
\begin{tabular}{ c | c | c | c | c | c | c | c }
\hline
 Sky & Water & Trees & Land & Rocks & Road  & Others & mIoU \\
\hline
 - &  0.9\% & 32.8\% & 34.8\% &  - &  14.9\% &  0\% &  16.7\% \\ 
\hline
\end{tabular}
}
\end{center}
\end{table}



{\bf Aeroscapes: }

As mentioned in the Experiments Setup, the AeroScapes dataset is used to evaluate the performance of the individual M4Semantic network because the dataset only contains semantic segmentation annotations. The M4Semantic results are reported in Table~\ref{table_aeroscapes} and compared with other methods. Our network was trained for 200 epochs with a batch size of 3 and a learning rate of \(10^{-4}\) for the first 70 epochs and then decreased to \(10^{-5}\). 

\begin{figure*}[h!]%
    \centering
    {\includegraphics[width=0.8\linewidth]{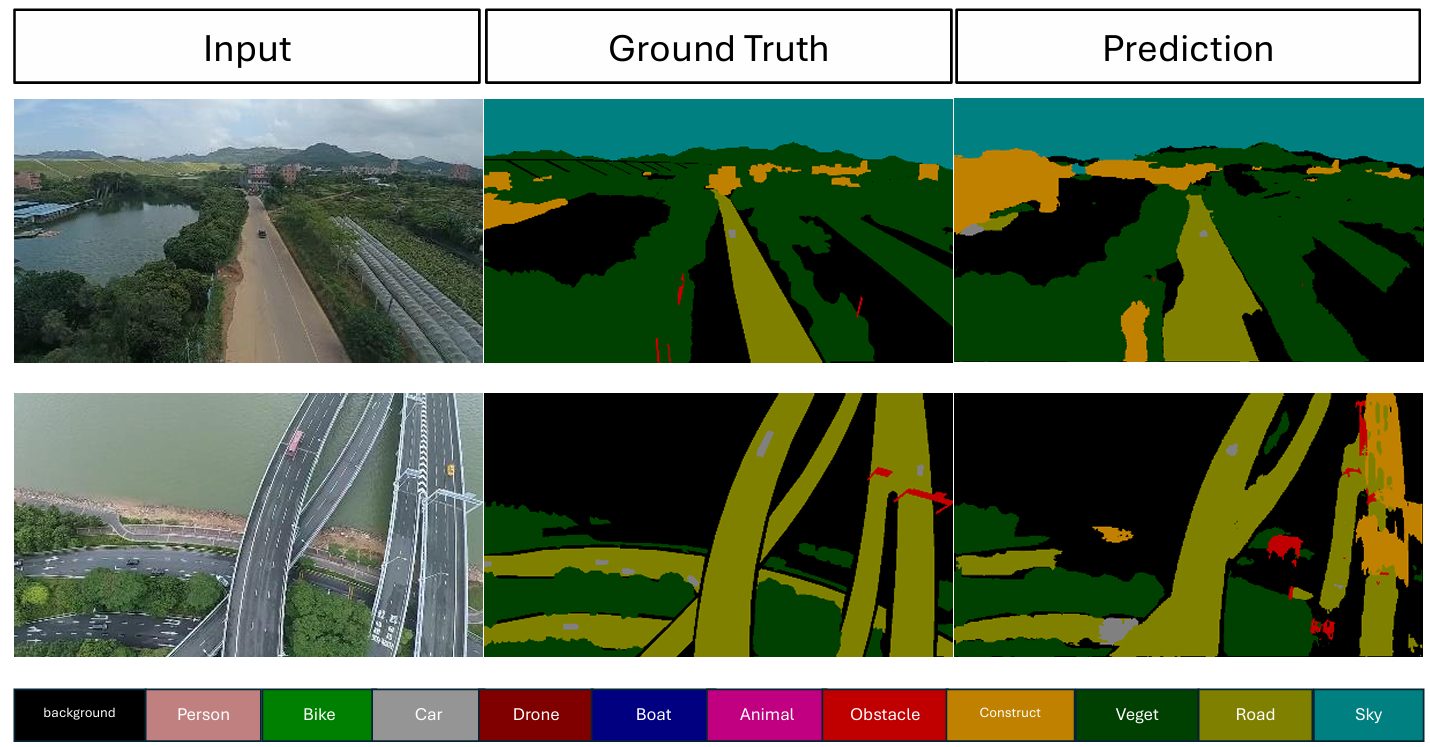} }
    \caption{\centering Visualization of the predicted semantic maps using M4Semantic on sample images from the Aeroscapes test data.}%
    \label{fig:qualitaero}%
\end{figure*}

M4Semantic was implemented on TensorFlow as one whole model that can be trained in an end-to-end fashion without separation between the weight files of the encoder and the decoder. This led to a more compact code but limited us from pretraining the encoder separately on Imagenet, as was done in the other methods. 
Nevertheless, we could produce a competitive mIoU compared to the others ($50.4\%$ compared to the best $52\%$), as can be seen in Table~\ref{table_aeroscapes}.
In Figure~\ref{fig:qualitaero}, a visualization of the output semantic segmentation maps predicted by M4Semantic on samples from the AeroScapes dataset is presented. It can be seen that the network can capture the layout of the input images and detect well the roads, trees, sky, and background areas.

\begin{table}[h]
\begin{center}
\caption{\centering Comparison of our M4Semantic architecture with other semantic segmentation methods benchmarked on Aeroscapes. P means pretrained on other datasets and S means trained from scratch}
\label{table_aeroscapes}
\resizebox{0.7\linewidth}{!}{
\begin{tabular}{ c || c | c | c | c }
\hline
 Method  & {Params(M)} & P/S & Open-Source & mIoU $\uparrow$ \\
\hline
FCN-8S~\cite{zheng2023deep} & 14.7 & P & No & 43.12\% \\
FCN-16S~\cite{zheng2023deep} & 14.7 & P & No & 44.52\% \\
FCN-32S~\cite{zheng2023deep} & 14.7 & P & No & 45.51\% \\
FCN-ImageNet-4S~\cite{aeroscapes} & 14.7 & P & No & 48.96\% \\
FCN-Cityscapes~\cite{aeroscapes} & 14.7 & P & No & 49.55\% \\
FCN-ADE20K~\cite{aeroscapes} & 14.7 & P & No & 51.62\% \\
FCN-PASCAL~\cite{aeroscapes} & 14.7 & P & No & 52.02\% \\
\textbf{M4Semantic (Ours)} & 2.61 & S & Yes & 50.40\%  \\
\hline

\hline
\end{tabular}
}
\end{center}
\end{table}


\subsection{Ablation experiments}
\label{ablation}
We conduct architecture study experiments on M4Semantic to highlight the importance of the addition or ablation of different modules. The results can be found in Table~\ref{table_ablation}. From the top part of the table, we can notice that using 5 levels produced the highest $mIoU$ ($76.8\%$) while keeping the inference time relatively low ($9.8$ ms per frame). 

In the bottom part, Original means M4Semantic (5 level). In Original+\{SNCV\}, we used the Spatial Neighbourhood Cost Volume module used in the decoder of M4Depth, see Figure~\ref{fig:modules_depth}, on the encoded feature maps instead of adding the normalized feature maps directly in the preprocessing unit. Such a module measures the two-dimensional spatial autocorrelation of the scene and improves the performance in depth estimation. However, using such a module in M4Semantic did not improve the performance in semantic segmentation and, moreover, it increased the inference time. So, we decided to discard it in semantic segmentation.

In Original+\{\(S^{t-1}\)\}, we test the addition of time dependency on previous frames in M4Semantic. At each decoder level, the semantic segmentation map predicted from the previous frame \(S^{t-1}\)\ is used along with the camera motion information and the ground truth depth map to warp it and give an initial prediction of the semantic map of the current frame. For more details about the warping operation, see the Appendix. We concatenate such a warped map with the outputs of the preprocessing module, Figure~\ref{fig:modules}, at each decoder level in order to act as an initial hint for the semantic segmentation map at the current time step. While such a technique achieved a higher mIoU $78.4\%$, the inference time increased due to the added warping computation. Also, the warping of the segmentation map requires the ground truth depth map, and this is not guaranteed to be available in reality. For these reasons, we decided to discard such a module in the semantic segmentation network.

Given this architecture study and the one done on the single depth estimation architecture~\cite{m4depth}, we choose the number of levels of Co-SemDepth to 5 levels, and we adopt the architecture depicted in Figure~\ref{fig:joint}.

\begin{table}[h]
\begin{center}
\caption{\centering Evaluation of our M4Semantic architecture on MidAir with the addition (+) or ablation (-) of different modules. The top part evaluates choosing a different number of levels. The bottom part was performed on M4Semantic (5level).}
\label{table_ablation}
\resizebox{0.7\linewidth}{!}{
\begin{tabular}{ c || c | c }
\hline
 Architecture &   Inf. Time (ms/f) $\downarrow$  & mIoU $\uparrow$ \\
\hline
M4Semantic (4level) & 9 & 75.1\% \\ 
M4Semantic (5level) & 9.8 & \textbf{76.8}\% \\ 
M4Semantic (6level) & 10.9 & 74.9\% \\ 
\hline
Original-\{DINL\} & 9.5 & 75.6\% \\ 
Original-\{Normalize\} & 9.7 & 74.1\% \\ 
Original+\{SNCV\} & 17.7 & 71.9\% \\ 
Original+\{\(S^{t-1}\)\} & 16.7 & 78.4\% \\ 
\hline
\end{tabular}
}
\end{center}
\end{table}

\chapter{Synthetic-to-Real Analysis}
\label{chap:syntoreal}
\ifpdf
    \graphicspath{{Chapter4/Figures/PNG/}{Chapter4/Figures/PDF/}{Chapter4/Figures/}}
\else
    \graphicspath{{Chapter4/Figures/EPS/}{Chapter4/Figures/}}
\fi

\section*{Summary}

In this chapter, the goal is to assess the synthetic-to-real performance of joint networks on aerial datasets. First, a detailed description of our collected \textit{TopAir} dataset is provided in section~\ref{sec:topair}. Then, a brief recounting of the methods used in the synthetic-to-real experiments is presented. The setup used to conduct the experiments is introduced. Then, multiple experiments are discussed to test the effect of different training modalities on closing the gap between the synthetic and real domains. 

To summarize, TopAir is a synthetic aerial dataset collected using AirSim in a variety of outdoor environments, comprising $\sim10K$ video frames with annotations of depth, segmentation, and camera translation-rotation data. To assess the synthetic-to-real domain shift, two networks with different backbone architectures are employed in the experiments for comparison: Co-SemDepth (U-Net backbone) and TaskPrompter (ViT backbone). The obtained results show that, generally, Co-SemDepth is more robust in the depth estimation generalization, while TaskPrompter (which has high capacity) is better in semantic segmentation. 
In the zero-shot learning scheme, changing the synthetic data used for training affects the generalization performance of the network depending on the similarity of the scenes between the synthetic and real data. In the few-shot learning scheme, both depth estimation and semantic segmentation benefit from the addition of a small percentage of real data to the training of the network.




\section{TopAir Data Collection}
\label{sec:topair}

As was discussed in Chapter~\ref{chap:related}, only a few datasets are available in the aerial field that contain both depth and semantic segmentation annotations. Such datasets do not cover all possible variations of outdoor environments, camera viewpoint, altitude, tilting angle, weather conditions, and lighting. We seek to contribute to enriching the available aerial data, especially those annotated, by collecting a new synthetic aerial dataset covering as much variations as possible. To this purpose, we use \textit{AirSim} simulator integrated with \textit{UnrealEngine4} to collect the data and call it \textit{TopAir}. 
AirSim~\cite{airsim} is an open source simulator for drones, cars, and more that is built on the game development engine UnrealEngine~\cite{unrealengine}. The simulator is well-documented, and it has multiple modes for controlling the drone or the vehicle in the simulation. In our data acquisition, we use the Computer Vision mode that simplifies the physics of the drone and treats it as a camera moving in space. By using this mode, we could navigate using only the keyboard without the need for a particular remote controller of actual drones. We adjusted the settings to collect RGB images, depth maps, and semantic segmentation maps. The setting values we tuned can be found in Table~\ref{tab:airsim_sett}. All other settings were kept as the AirSim default.

\begin{table}[h!]
\begin{center}
\caption{\centering Settings adjusted of AirSim during data acquisition}
\label{tab:airsim_sett}
\resizebox{0.6\linewidth}{!}{
\begin{tabular}{ c | c  }
\hline
 Setting & Value \\
 \hline
 SimMode & ComputerVision\\
 ImageType & 0, 3, 5 \\
 RecordOnMove & True \\
 RecordInterval  & 0.05 (Sec)\\
 Width  & 384\\
 Height  & 384\\
 FOV\_Degrees & 90\\
 AutoExposureSpeed & 100\\
 AutoExposureMaxBrightness & 0.64\\
 AutoExposureMinBrightness & 0.03\\
\hline
\end{tabular}

}
\end{center}
\end{table}

As the name \textit{TopAir} suggests, the data have been collected with a nadir (top) camera viewpoint where the tilting (pitch) angle is close to or equal to \(90\) degrees. The motivation to use this viewpoint, particularly, is the limited available synthetic data with the top view; only SkyScenes contains top view images, while MidAir, TartanAir, and SynDrone contain mostly images with forward or oblique view. The dataset is available at: \url{https://huggingface.co/datasets/yaraalaa0/TopAir}

During the collection, the camera is set at low, mid, and high altitudes varying from 10 to 100 meters above the ground. 
A multiplicity of UnrealEngine environments, rural and urban, were used for the data collection. They cover a variety of weather and daytime conditions, as well as object and building styles and layouts. These environments are:
\begin{itemize}
    \item \textit{Africa} (3 trajectories): an African forest containing mostly trees and sandy ground. The lighting is around sunset.
    \item \textit{City Park} (7 trajectories): a park in the city that has trees, grass, roads for cars and pedestrians, lakes of greenish water, kids' areas, bridges, and playgrounds. The lighting in the environment is around midday.
    \item \textit{Oak Forest} (3 trajectories): forest of mid-height trees and muddy ground. There exist several small rocks in the ground. Lighting is around the afternoon.
    \item \textit{Rural Australia} (4 trajectories): an environment representing a rural area in the deserts of Australia. The environment has roads, where cars and vehicles were placed manually, and trees of different heights on the side of the roads.
    \item \textit{Assetsville Town} (15 trajectories): an environment that represents an urban scenario where there are multiple houses, vehicles, shops, government buildings, police stations, farms, and people. Some houses, buildings, cars, and persons were inserted in the environment to enrich its diversity. The lighting here is early to midday sunlight.
    \item \textit{Accustic Cities} (3 trajectories): Modern high buildings with roads in the middle and a lake of brownish water with a bridge on it. Lighting is strong around midday.
    \item \textit{Downtown City} (7 trajectories): a small downtown city with low-height buildings and roads in the middle. Tables and decorations are placed on the sides and center of the pedestrian roads. Some cars were inserted manually in this environment. The lighting is around the afternoon. 
    \item \textit{Edith Finch} (2 trajectories): A small house with a lake and a truck in front of it and tall trees around. Lighting is around evening (moonlight
    \item \textit{Landscape Mountains} (1 trajectory): Rocky mountains and rocky ground with a lake of bluish water in the middle. Lighting is around midday.
    \item \textit{Nature Pack} (1 trajectory): a small natural environment of a waterfall and trees with grassy land. Rocks rest on the sides of the waterfall. The lighting in this environment is midday.
    \item \textit{Neighbourhood} (8 trajectories): a rural neighbourhood area with many houses of blue roofs. Some houses have swimming pools. Cars of different colors are on the roads. The lighting is afternoon sunlight.
\end{itemize}

\begin{figure}[h!]%
    \centering
    \includegraphics[width=0.9\linewidth]{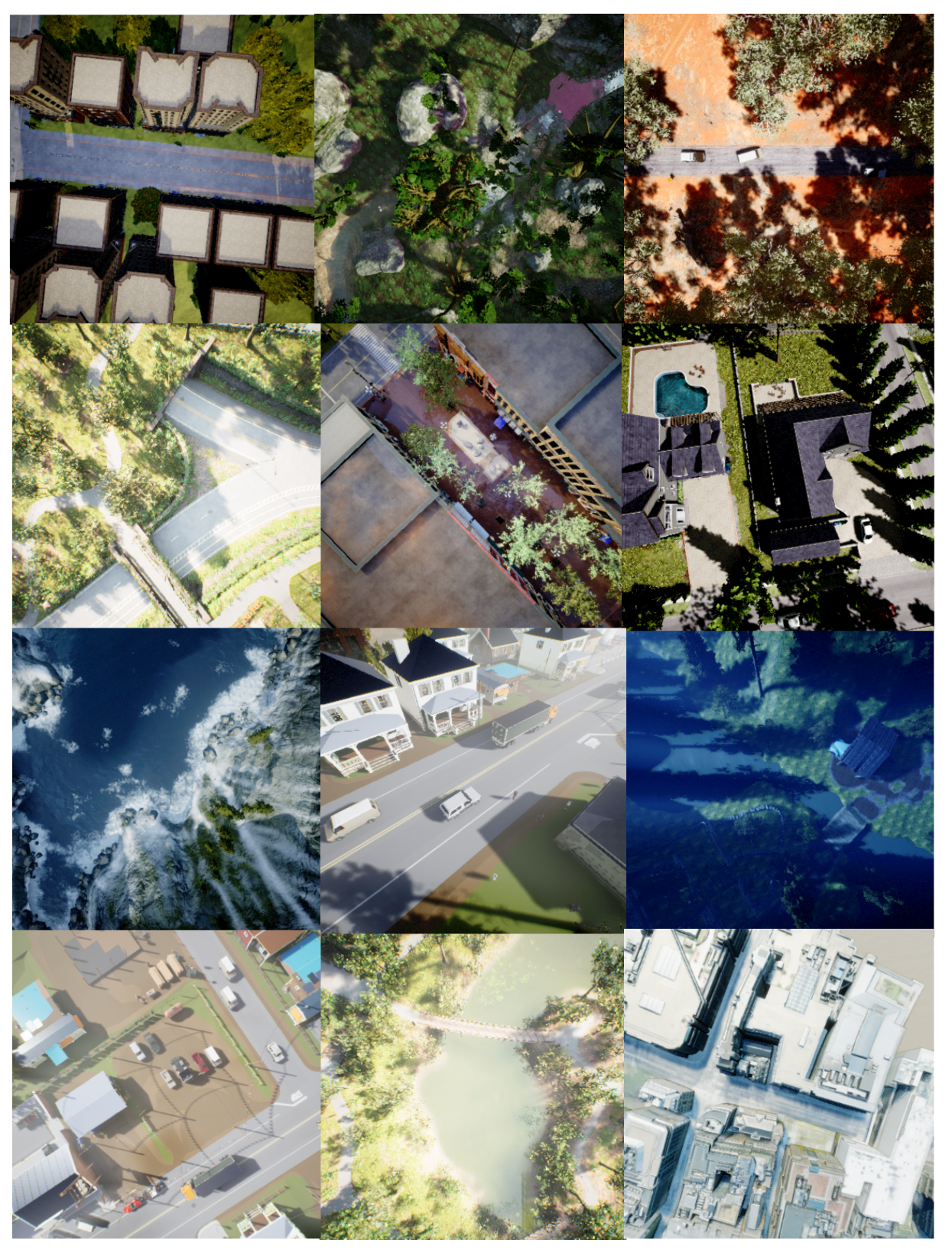}
    \caption{\centering Sample images from the collected TopAir dataset showing the variety of environments used and the variety of altitudes}%
    \label{fig:topair}%
\end{figure}

An additional advantage of TopAir is that it is a light-weight dataset that has an image resolution of 384$\times$384 and a total size of only 4GB, making it convenient for downloading and training on resource-restricted machines. Samples of RGB images from the collected TopAir data can be found in Figure~\ref{fig:topair}.

The dataset comprises 54 trajectories containing, in total, 10,385 frames. Each RGB frame is annotated with a semantic segmentation map, a depth map, and camera transformation (location$+$orientation) data. To generate the semantic segmentation annotations, the object classes in each environment were limited to a set of 9 classes by manually renaming the elements in UnrealEngine to represent these classes. While such a process required a significant effort, it had to be done only once for each environment. After that, we could collect as much data as needed from the environment. The semantic segmentation classes we used are:
\begin{enumerate}
    \item \textit{Sky}: the sky
    \item \textit{Water}: includes all water surfaces, including lakes, seas, or swimming pools
    \item \textit{Trees}: all trees and vegetation above the ground, whether short or tall, excluding grass
    \item \textit{Land}: all ground types (dirt ground, rocky ground, grass, sand, etc)
    \item \textit{Vehicle}: all types of ground vehicles (car, trucks, bus, etc) excluding bikes
    \item \textit{Rocks}: boulders and rocks (including rocky mountains)
    \item \textit{Road}: Lanes, streets, paved areas on which cars drive, or sidewalks for pedestrians
    \item \textit{Building}: buildings, residential houses, and constructions, including bridges
    \item \textit{Others}: any other object that is not included in the above classes
\end{enumerate}

\begin{figure}[h!]%
    \centering
    \includegraphics[width=0.9\linewidth]{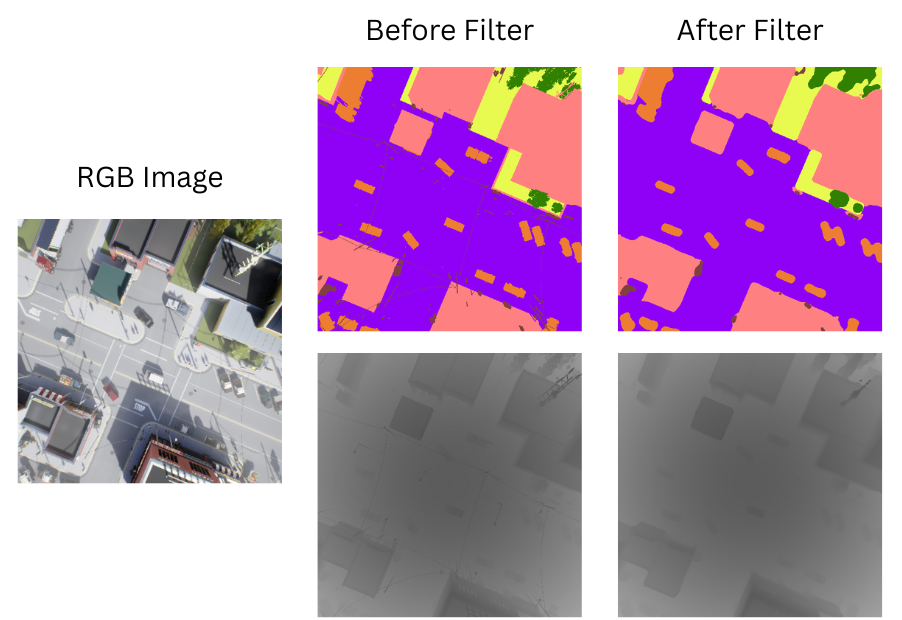}
    \caption{\centering Applying a median filter on the collected depth and semantic segmentation maps of TopAir to smooth the appearance and remove unnecessary details}%
    \label{fig:curating_data_2}%
\end{figure}

The settings of AirSim were adjusted to collect the needed annotations (depth + segmentation + location) automatically while navigating through the environments. A median filter of size 10 was applied to all the generated depth and segmentation maps to smooth their appearance and remove the noise. Such a filter computes and assigns the median value to all the pixels included in a square window of size $10\times10$, and this helps in removing the noisy non-frequent pixels from the segmentation map (they get assigned the most frequent value within the $10\times10$ window). See Figure~\ref{fig:curating_data_2} for a demonstration of the maps before and after applying the filter. 

\begin{figure}[h!]%
    \centering
    \includegraphics[width=0.85\linewidth]{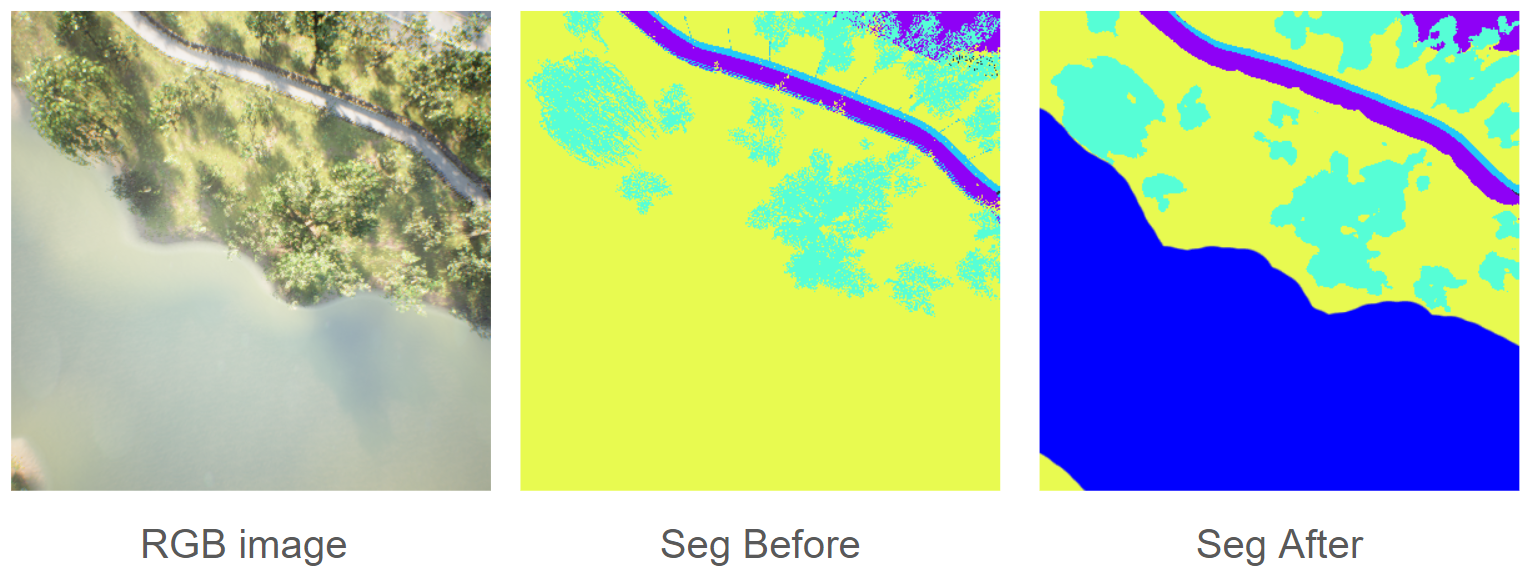}
    \caption{\centering Curating the semantic segmentation maps of the collected TopAir data by manually painting the Water region after applying a median filter}%
    \label{fig:curating_data}%
\end{figure}

There was a problem faced in the AirSim segmentation of water regions, where they were treated as transparent objects, and they were identified as "Land" instead of "Water" in the output semantic segmentation maps. For this reason, we had to manually post-process such generated maps, using a photo editor \textit{PhotoPea}~\cite{photopea}, to paint "Water" areas that were not appearing in the segmentation maps, see Figure~\ref{fig:curating_data}. 

\begin{figure}[h!]%
    \centering
    \includegraphics[width=0.85\linewidth]{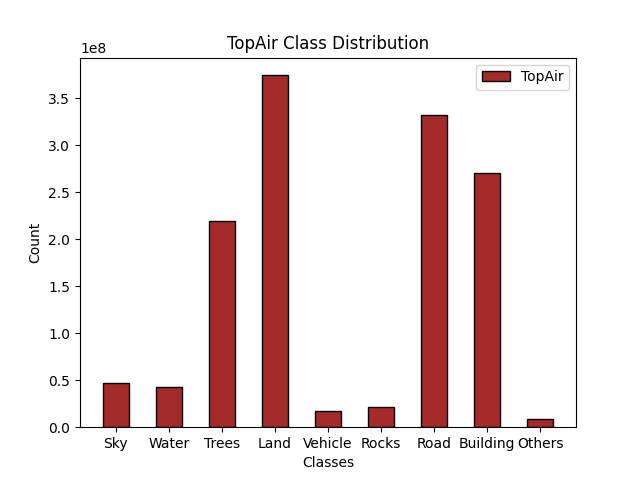}
    \caption{\centering TopAir per-pixel semantic segmentation classes distribution}%
    \label{fig:topair_seg_dist}%
\end{figure}

The per-pixel distribution of the semantic segmentation classes in TopAir is depicted in Figure~\ref{fig:topair_seg_dist}. Overall, it shows the diversity of the classes present in the dataset. Trees, Land, Roads, and Buildings have relatively high pixel count because they occupy big spaces in the images. On the other hand, vehicles, Rocks, and Others occupy small spaces, and for this reason, their pixel count is relatively low. Sky and Water appear only in a small number of images, and they have around 50 million pixels each.

\begin{figure}[h!]%
    \centering
    \includegraphics[width=0.85\linewidth]{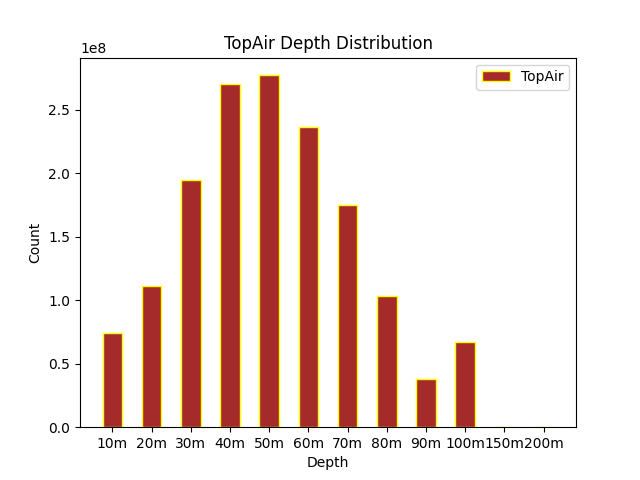}
    \caption{\centering TopAir per-pixel depth values distribution}%
    \label{fig:topair_depth_dist}%
\end{figure}

The per-pixel distribution of the depth values is depicted in Figure~\ref{fig:topair_depth_dist}. It can be noticed from the figure that there is a diversity and a good representation of all depth values ranging from 10 to 100 meters in the collected data. The most frequent depth values range from 30 to 70 meters, while the least frequent are the edge values 10, 90, and 100 meters.

\begin{figure}[h!]%
    \centering
    \includegraphics[width=0.5\linewidth]{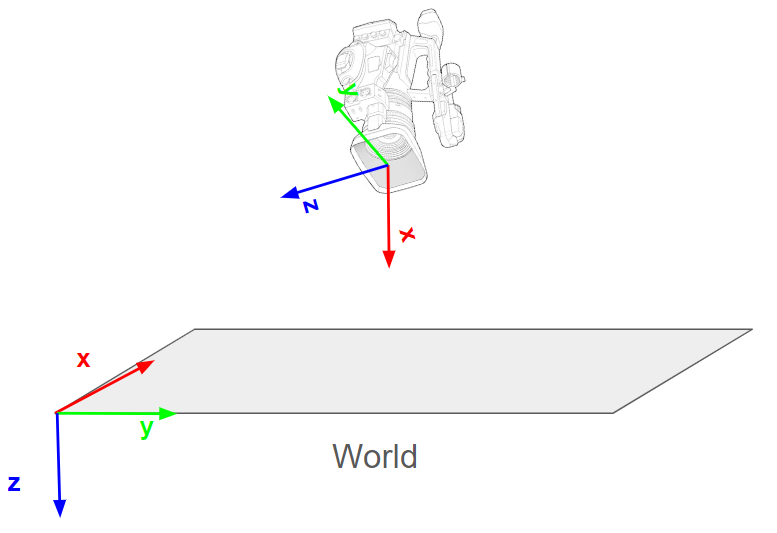}
    \caption{\centering An illustration of the convention of the camera and world reference frames used in AirSim while collecting TopAir data}%
    \label{fig:reference_frames}%
\end{figure}

The data was collected at a frame rate of 20 FPS, and in Figure~\ref{fig:reference_frames} we demonstrate the convention of the reference frames of the camera and the world used in the simulation.

\section{Adopted Methods}

In the synthetic-to-real analysis, we focus on evaluating the synthetic-to-real generalization performance of joint architectures because of their memory and time efficiency compared to single architectures. 
Specifically, we consider two joint methods: Co-SemDepth~\cite{cosemdepth} and TaskPrompter~\cite{j8}. The motivation to use these two networks, particularly, is that the first network, Co-SemDepth, is a light, small network that is more suitable for deployment on resource-limited hardware platforms,
while the second network, TaskPrompter, is relatively big and suitable for offline testing or deployment on only high-performance hardware platforms. Since the architecture of \textbf{Co-SemDepth} was previously described in the previous chapter, chapter~\ref{chap:cosemdepth}, we hereby give a brief description of the TaskPrompter's architecture. 


The main idea of \textbf{TaskPrompter}~\cite{j8} is to design a multi-task architecture that can jointly model (i) task generic, and (ii) task specific representations, and (iii) cross-task interactions. While in Co-SemDepth and several other multi-tasking works~\cite{j1,j2,j3, j4}, the task-generic representations are extracted using the shared encoder module, and the task-specific learning is done using a separate decoder module for each task, in TaskPrompter, the three objectives are learned in each network layer in an end-to-end manner. It is a Spatial-Channel Multi-task Prompting framework based on vision transformers. They design a set of spatial-channel task prompts and learn their spatial and channel interactions with the input image tokens in each transformer layer. With the help of task prompts and attention mechanisms, the network learns task-generic, task-specific representations and cross-task interactions in the same network layer without the need to add dedicated network modules for learning them. 

\begin{figure}[h!]%
    \centering

  \includegraphics[width=0.9\linewidth]{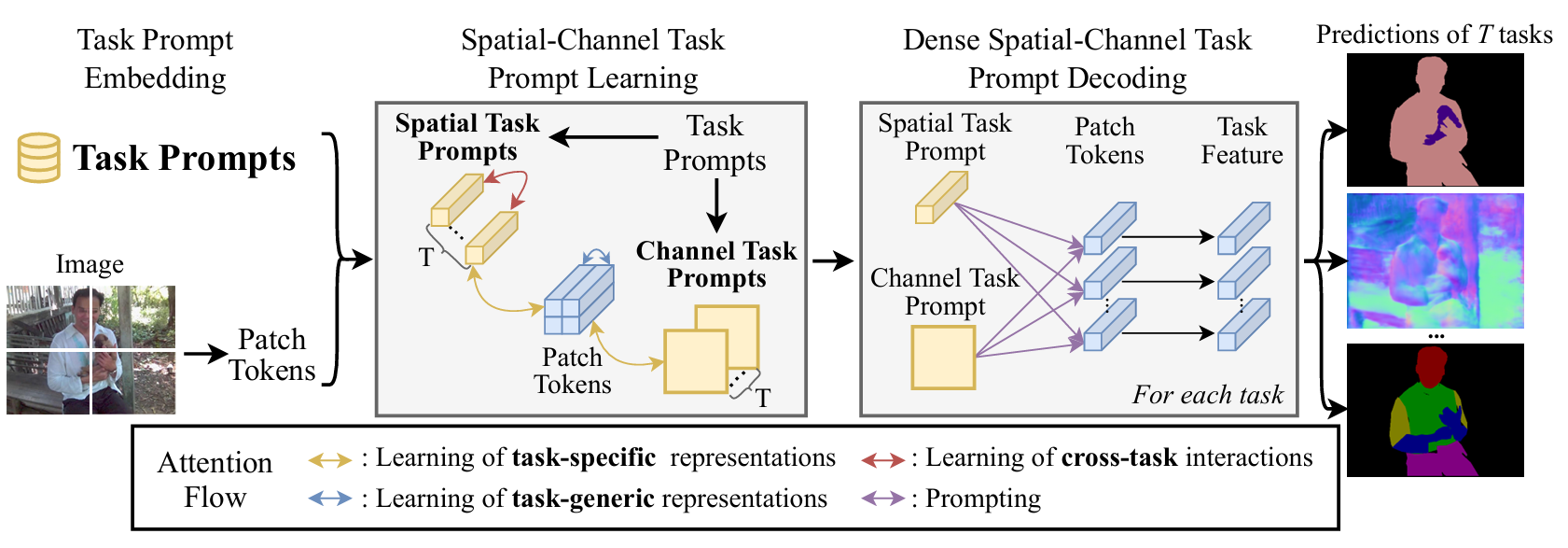}
  \caption{\centering Illustration of the spatial-channel multi-task prompt learning framework used in each layer of the transformer in TaskPrompter. Task generic representations are extracted from the image patch tokens, while task-specific representations are extracted from the relations between spatial task prompts, channel task prompts, and image patch tokens. The cross-task interactions are extracted from the spatial task prompts. Taken from~\cite{j8}, courtesy of the authors, in accordance with the policy of ICLR
  }
    \label{fig:taskprompter}%
\end{figure}

Each task prompt learns the task-specific representations of one task, while the shared image tokens are learned by the contribution of all task prompts. The interactions between each pair of task prompts contribute to the cross-task relationships. Task prompts are learnable and tuned during training. They are task-specific representations that learn both spatial and channel information of a specific task. After learning the task prompts, the task-generic image tokens, and the cross-task interactions comes the decoding phase to 
output the final predictions for each task (in our case, we are interested only in two tasks: depth estimation and semantic segmentation). This method produced state-of-the-art results on NYUD-V2 and PASCAL datasets. However, the architecture has a large number of parameters ($>$100 million parameters for the different versions) and high inference time, as will be revealed in the experiments. 

The whole process can be divided to three stages (as illustrated in Figure~\ref{fig:taskprompter}): Task Prompt Embedding, Spatial-Channel Task Prompt Learning, and the Decoding phase. 

\section{Experimental Setup}
In this section, we discuss the setup of experiments conducted to test the effect of different factors on the synthetic-to-real generalization performance of deep joint architectures in the aerial domain. In our analysis, the following factors are considered:
\begin{enumerate}
    \item Given a real dataset of interest for testing, how does changing the synthetic datasets used for training change the model performance on real data?
    \item How can changing the model architecture affect the synthetic-to-real generalization performance?
    \item Whether adding a small number of real data to the training would affect positively the synthetic-to-real generalization of the network (few-shot prediction)?
\end{enumerate}

\subsection{General Settings}
For Co-SemDepth, 5 layers are used, and all the other values of the model are kept as the default. This results in a light model that has only 5.2 million trainable parameters. 
For TaskPrompter, instead, the vision transformer "Base" model is set as the backbone with an embedding dimension of 384 and a number of channel heads equal to 8. All the other values are kept as the default. This results in a big model that has around 126 million trainable parameters. \\

\noindent
{\em The synthetic datasets considered for training are}: 
\begin{itemize}
    \item MidAir: contains natural scenes with a forward camera viewpoint.
    \item TopAir: contains both natural and urban scenes with a top camera viewpoint. The environments used for training and validation are: Neighbourhood, Assets Ville Town, DownTown city, and City Park.
    \item SkyScenes: contains mostly urban scenes with forward and top camera viewpoints.
    \item SynDrone: contains mostly urban scenes with oblique camera viewpoint.
\end{itemize}

To make the training data balanced in each experiment, regarding camera viewpoint and dataset size, MidAir and TopAir datasets are concatenated and used jointly for training the models, while SkyScenes and SynDrone datasets are used together.

For training on MidAir+TopAir, due to the difference in the number of images in the two datasets, the number of training images per epoch is set to $2\times$ the number of images in the smaller dataset (TopAir). This results in a total number of training frames per epoch = $16868$. The sampling ratio from MidAir and TopAir is set to $1:1$, meaning an equal contribution from both datasets during training. For validation, the validation set of MidAir used in~\cite{cosemdepth} is added to the validation set of TopAir. This results in a number of validation frames equal to $5381$.  

Instead, for SkyScenes+SynDrone, the number of training images per epoch is also set to $2\times$ the number of training images of the smaller dataset (SkyScenes), resulting in a total number of training frames per epoch = $12972$. The sampling ratio is set to $1:1$. For validation, we select random trajectories covering a variety of conditions from the two datasets, creating a number of validation frames = $4138$. \\


\noindent
{\em The evaluation of depth estimation is carried out on the following real data}:
\begin{itemize}
    \item WildUAV: depth values ranging from 0 to 70 meters.
    \item Dronescapes: depth values ranging from 40 to 500 meters. 
\end{itemize}

Due to the limited number of available real datasets with depth annotation, only the WildUAV and DroneScapes datasets could be used in our experiments for conducting a quantitative analysis of the depth estimation. 

\begin{figure}[h!]%
    \centering
    \includegraphics[width=0.8\linewidth]{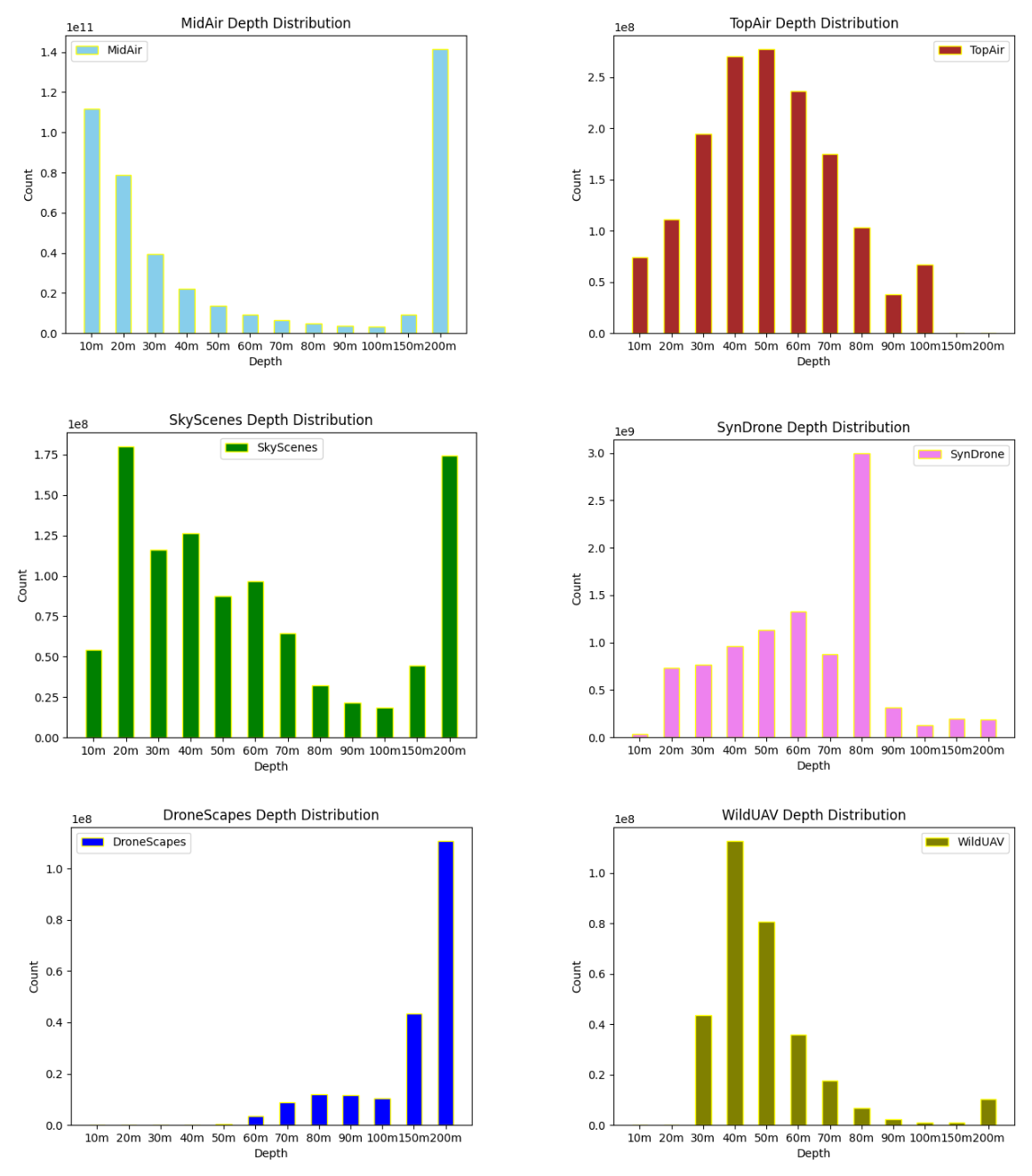}
    \caption{\centering Histograms of pixel-wise depth distribution in each dataset. Each pin on the x-axis indicates all depth values less than or equal to the pin value. It can be marked how the range of depth values in WildUAV is more relevant to the depth values of the synthetic datasets used for training, while this does not hold true for DroneScapes.}%
    \label{fig:depth_distribution}%
\end{figure}

In Figure~\ref{fig:depth_distribution}, the per-pixel distribution of the depth values in the datasets used is demonstrated. From the figure, it can be noted that the range of depth values of WilUAV is close to the range of values in the synthetic datasets used for training. However, the range of depth values of DroneScapes tends to include only large values. \\\\

\noindent
{\em The evaluation of semantic segmentation is done on the real datasets WildUAV, DroneScapes, RuralScapes, AeroScapes, FSI, UDD, and ICG. All of these datasets contain semantic segmentation annotations, and the evaluation can be done quantitatively and qualitatively. Here is a brief description of the semantic content of each of them:}
\begin{itemize}
    \item WildUAV: 1535 images resized to 528$\times$384 instead of the original 5280$\times$3956 resolution.
    \item Dronescapes: 1419 images resized to 480$\times$320 instead of the original 960$\times$540 resolution. The environments used are "Gradistei", "Herculane", "Olanesti", and "Petrova".
    \item RuralScapes: 1108 images resized to 420$\times$220 instead of the original 4096$\times$2160 resolution.
    \item Aeroscapes: 3269 images resized to 640$\times$360 instead of the original 1280$\times$720 resolution.
    \item FSI: 186 images resized to 300$\times$400 instead of the original 3000$\times$4000 resolution.
    \item UDD: 112 images resized to 400$\times$300 instead of the original 4000$\times$3000 resolution.
    \item ICG: 400 images resized to 500$\times$300 instead of the original 6000$\times$4000 resolution.
\end{itemize}

To tackle the third item in our testing factors, a small part of DroneScapes (\textit{600 images}) is added to the synthetic data during training. The percentage of real to synthetic images in the training is very small (\textit{$5\%$ for MidAir+TopAir and $4.6\%$ for SkyScenes+SynDrone}), and this reflects the typical situations for neural networks where synthetic data are abundant and a limited number of real, annotated data. However, Co-SemDepth requires as input the camera transformation between frames, and we can not use the transformation data provided in DroneScapes due to the ambiguity in the convention of its reference frames. For this reason, only TaskPrompter is used in our experiments for testing the third factor.


\subsection{Semantic Classes Mapping}
Each dataset is annotated with a unique set of segmentation classes. To unify the semantic segmentation classes used in our experiments, we define a set of 9 classes to include the most common and unique classes of interest in the aerial domain. Specifically the 9 classes are: Sky, Water, Trees, Land, Vehicle, Rocks, Road, Building, and Others. The segmentation classes of each dataset are mapped to our set as listed in Table~\ref{table_classes}. In general, the classes mapping between datasets may not be highly accurate due to the differences in the appearance and class definitions in each dataset. 

\begin{sidewaystable}
\begin{center}
\fontsize{7pt}{8pt}\selectfont
\caption{\centering The mapping of the semantic classes across different datasets. The mapping used for SkyScenes is the same as that done on the SynDrone dataset and is referred to as Sky/Syn.}
\label{table_classes}
\begin{tabular}{ c c c c c c c c c c c}

\hline
Common Set & \textit{MidAir} & \textit{TopAir} & \textit{Sky/Syn} & \textit{Dronescapes} & \textit{WildUAV} & \textit{Aeroscapes} & \textit{Ruralscapes} & \textit{UDD} &\textit{ FSI} & \textit{ICG}\\
\hline
0 Sky &  sky & sky &  sky &  sky &  sky & sky & sky  &  -  &  - & - \\
\hline
1 Water & water  & water &  water &  water &  water &  - &  water & - &  water body  &  water \\
 &   &  &   &   &   &   &   &   &   swimming pool &  pool \\
  &   &  &   &   &   &   &   &    &   flooded &   \\
\hline
2 Trees & trees & trees & vegetation & forest  &  deciduous tree  & vegetation & forest  &  vegetation  &  trees & tree \\
  &  &  &  &   &  coniferous tree &   &   &   & forest & other vegetation \\
  &  &  &  &   &  fallen trees &   &   &    &  &  \\
\hline
3 Land & dirt ground & land & terrain & land  &  dirt ground &  background & land  &  others & grass  &  grass \\
 & ground vegetation &  & other & hill &  ground vegetation &   &  hill &   & under-construction & dirt \\
  & rocky ground &  &  &  &  &   &   &    & sports arena &   \\
\hline
4 Vehicle &  - & vehicle & cars &  little-objects &  static car &  car &  car &   vehicle & vehicle  &  car \\
  &   &  & bus &   &  moving car &   &   &    &   &   \\
  &   &  & truck &   &   &   &   &    &   &  \\
  &   &  & motorcycle &   &   &   &   &    &   &  \\
  &   &  & rider &   &   &   &   &    &   &   \\
  &   &  & train &   &   &   &   &    &   &   \\

\hline
5 Rocks & boulders &  rocks  & - & - &  rocks &  - & -  &   -  &  - & rocks \\
\hline
6 Road & road & road & road &  road & road  & road  & road  &  road  &  road &  paved area \\
  &   &  & sideWalk &   &  sidewalk &   &   &    & parking  &  gravel \\
  &   &  & roadLine &   &   &   &   &    & background  &   \\
  &   &  & ground &   &   &   &   &    &   &   \\
\hline
7 Building &  construction & building & building &  residential & building & construction  &  residential &   facade  & property roof  &  roof \\
  &   & & bridge &   &  &   &  church &   roof &  chimney & door \\
  &   & & railTrack &   &  &   &   &    & industrial  &  window \\
  &   & & wall &   &   &   &  &    & solar panels  &  wall \\
  &   & &  &   &   &   &  &    & antenna  &   \\
   &   & &  &   &   &   &  &    & window  &   \\
   &   & &  &   &   &   &  &    & secondary structure  &   \\
\hline
8 Others & others & others & static & little-objects & fence &  person &  fence & - &  trampoline &  bicycle \\
  & road sign  & & dynamic &   &  people &  bike &  haystack &    & garbage bins  &  dog \\
  &  train track & & fence &   &  empty &  drone & person  &    &  boat & fence  \\
  & animals  & & pedestrian &   &   &  boat &   &    & street light  & person  \\
  &  empty & & pole &   &   & animal &   &   & water tank  & obstacle  \\
  &   & & trafficSign &   &   &  obstacle &   &    & cables & fence-pole \\
  &   & & trafficLight &   &   &   &   &    &   &   \\
  &   & & unlabeled &   &   &   &   &    &   &   \\
  &   & & other (in town10) &   &   &   &   &    &   &   \\
  &   & & bicycle &   &   &   &   &    &   &   \\
  &   & & guardRail &   &   &   &   &    &   &   \\
\hline
\end{tabular}
\end{center}
\end{sidewaystable}

\begin{figure}[h!]%
    \centering
    \includegraphics[width=0.8\linewidth, height=20cm]{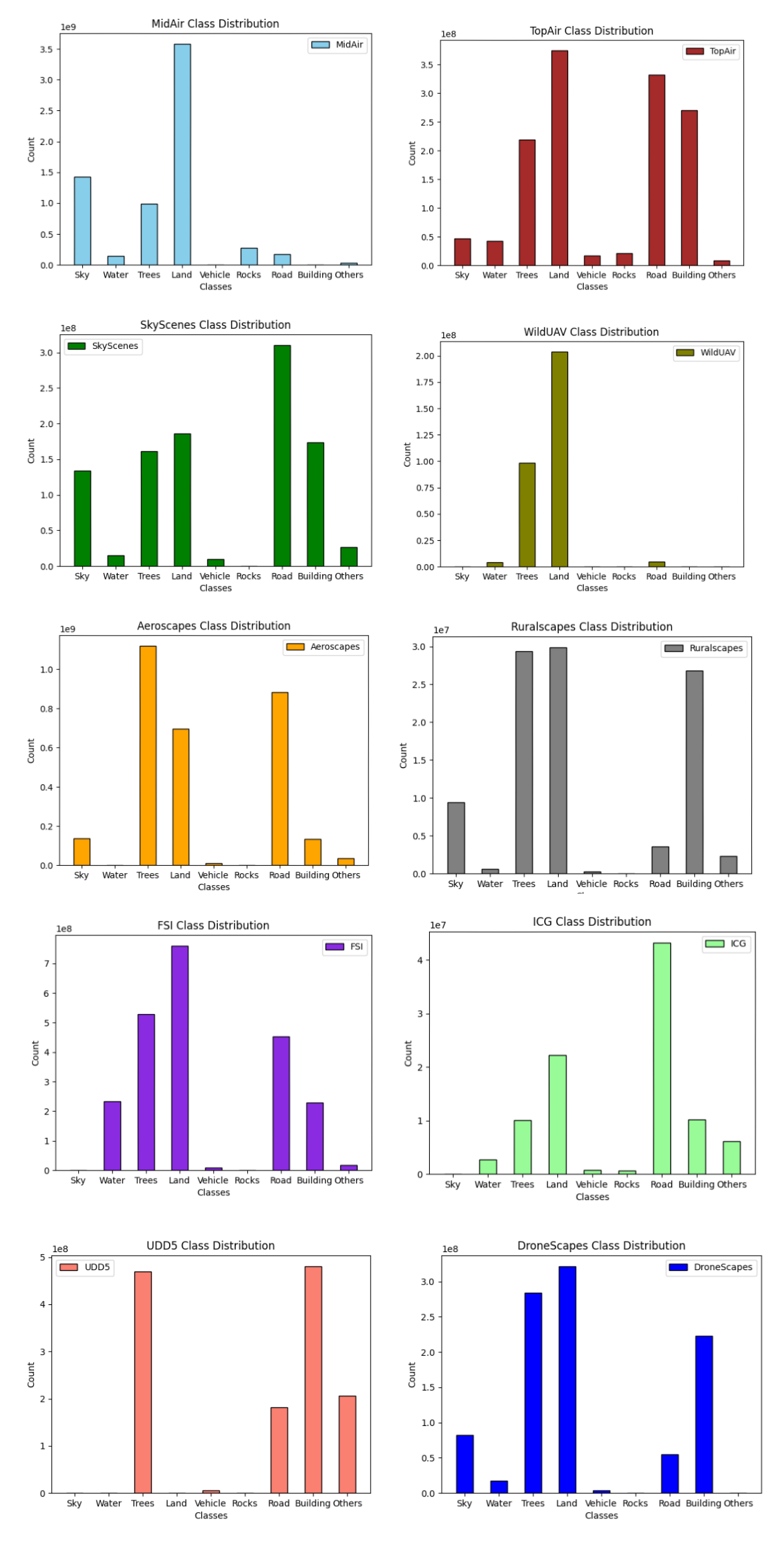}
    \caption{\centering Histograms of pixel-wise class distribution in each dataset after mapping to our common set of segmentation classes}%
    \label{fig:distribution}%
\end{figure}

The pixel-wise class distribution for each dataset after applying the mapping can be found in Figure~\ref{fig:distribution}. As can be noted from the figure, datasets differ in the semantic classes distribution; some of them have a high representation of buildings and road classes (like SkyScenes, UDD, and ICG) showing that they are mostly urban, while others contain a high representation of land and trees and do not contain buildings (like MidAir and WildUAV) suggesting that they are captured in the wild nature. It can also be noted how the class distribution of our introduced TopAir dataset is nearly representative of all the classes compared to the other datasets. The "Vehicle" class count is low in all the datasets because the vehicles (cars, bikes, etc) normally occupy a small number of pixels compared to the other classes.

\subsection{Training and Evaluation Settings}
For Co-SemDepth, the model is trained using the Adam optimizer for 60 epochs and with a fixed learning rate of $10^{-4}$. The batch size is set to 3 and the number of frames per sequence to 4. Table~\ref{table_train_cosemdepth} summarizes the training parameters of Co-SemDepth. 

For TaskPrompter, the model is trained using the Adam optimizer for a maximum of 80K iterations. The batch size is set to 3, the learning rate to $10^{-5}$, and a polynomial scheduler is used with a weight decay of $10^{-6}$. Table~\ref{table_train_taskprompter} summarizes the training parameters of TaskPrompter. 

During the training of both models, image augmentations of random rotation, flipping, and changing color (contrast, brightness, hue, and saturation) are applied. The depth values are cropped to a maximum of 200 meters.

It should be noted that when testing TaskPrompter on different datasets, the testing images have to be resized to have a resolution equal to the resolution of the training images, otherwise it gives an error. However, such a problem is not faced with Co-SemDepth, which can accept testing images of various sizes without the need to resize them.


\begin{table}[H]
\fontsize{6pt}{9pt}\selectfont
\begin{center}
\caption{\centering Training parameters used for training \textbf{Co-SemDepth} on MidAir+TopAir and SkyScenes+SynDrone}
\label{table_train_cosemdepth}
\resizebox{\linewidth}{!}{
\begin{tabular}{@{}ccccccccc@{}}
\hline
 Train Dataset  & optimizer & lr & weight decay & epochs & batch & Train size & Valid size & resolution \\
\hline
 MidAir+TopAir  & Adam & $10^{-4}$ &  - & 60 & 3 & 16868 & 5381 & $384\times384$ \\
SkyScenes+SynDrone  & Adam & $10^{-4}$ & - & 60 & 3 &  12972 & 4138  & $480\times320$ 
\\

\hline
\end{tabular}
}
\end{center}
\end{table}

\begin{table}[H]
\fontsize{6pt}{9pt}\selectfont
\begin{center}
\caption{\centering Training parameters used for training \textbf{TaskPrompter} on MidAir+TopAir and SkyScenes+SynDrone}
\label{table_train_taskprompter}
\resizebox{\linewidth}{!}{ 

\begin{tabular}{@{}ccccccccc@{}}
\hline
 Train Dataset  &  optimizer & lr & weight decay & iterations & batch & train size & valid size & resolution \\ 
\hline
MidAir+TopAir & Adam & $10^{-5}$ & $10^{-6}$ & 80k & 3 &  16868 & 5381 & $384\times384$ \\ 
SkyScenes+SynDrone & Adam & $10^{-5}$ & $10^{-6}$ & 80k & 3 & 12972 & 4138 & $480\times320$ \\

\hline
\end{tabular}
}
\end{center}
\end{table}

\begin{figure}[h!]%
    \centering
    
\begin{subfigure}[b]{\textwidth}
  \centering
  \includegraphics[width=0.85\linewidth]{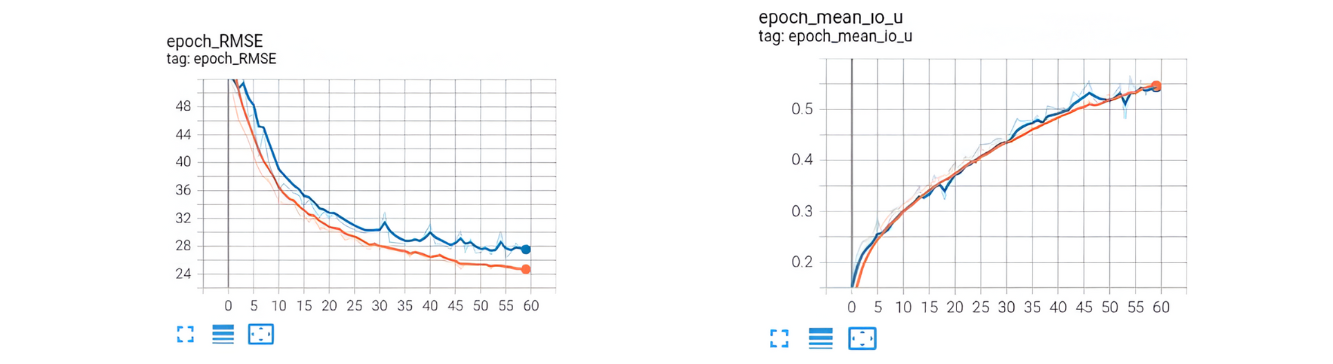}
  \caption{Co-SemDepth training plots on MidAir+TopAir}
  \label{fig:cosemdepth_midtop}
\end{subfigure}
\begin{subfigure}[b]{\textwidth}
  \centering
  \includegraphics[width=0.9\linewidth]{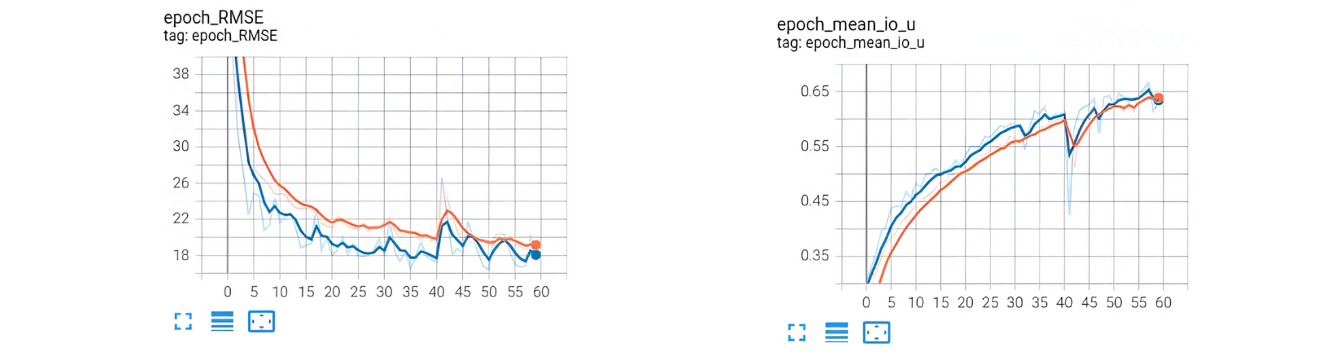}
  \caption{Co-SemDepth training plots on SkyScenes+SynDrone}
  \label{fig:cosemdepth_skysyn}
\end{subfigure}
\begin{subfigure}[b]{\textwidth}
  \centering
  \includegraphics[width=0.9\linewidth]{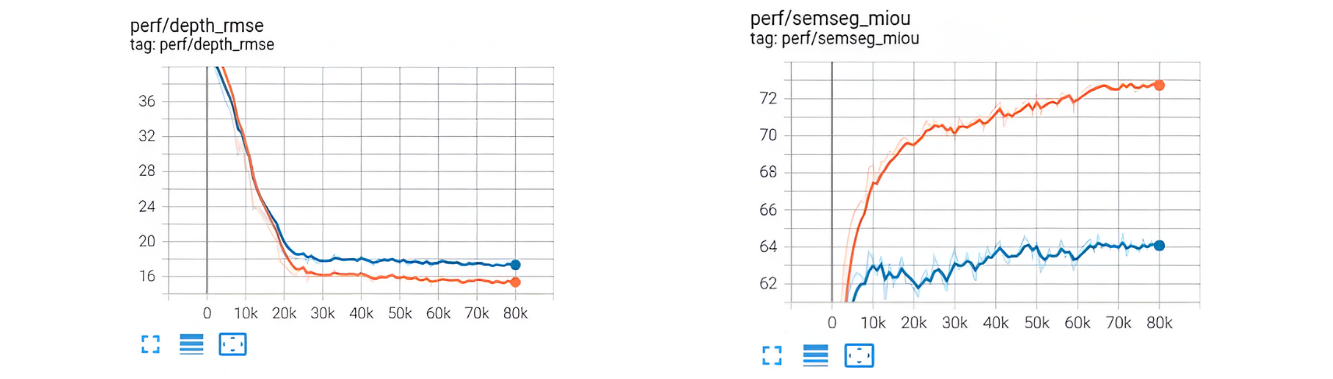}
  \caption{TaskPrompter training plots on MidAir+TopAir}
  \label{fig:taskpormpter_midtop}
\end{subfigure}
\begin{subfigure}[b]{\textwidth}
  \centering
  \includegraphics[width=0.9\linewidth]{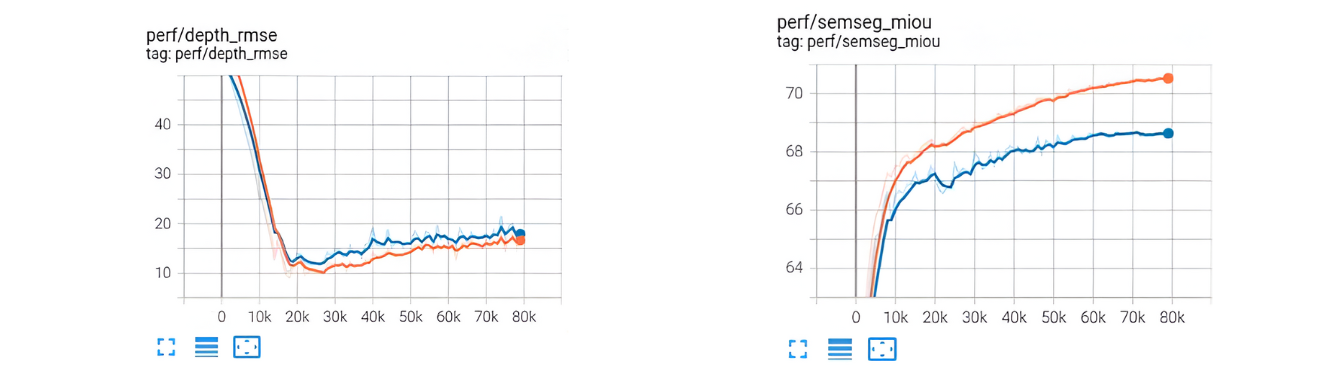}
  \caption{TaskPrompter training plots on SkyScenes+SynDrone}
  \label{fig:taskprompter_skysyn}
\end{subfigure}

    \caption{\centering Plots of the train and validation RMSE and mIoU values of the models during training on MidAir+TopAir and SkyScenes+SynDrone. While TaskPrompter tends to overfit on the training data due to its high capacity, Co-SemDepth's performance is almost equal on the train and validation data (no overfitting).  For all the models, the checkpoint that produced the best validation results was selected.}%
    \label{fig:train_curves}%
\end{figure}

The camera motion model used in our experiments with Co-SemDepth is shown in Figure~\ref{fig:camera_transf}. For all the datasets that used a camera model with a different orientation of the principal axes, the necessary transformation is applied on the camera orientation to make it similar to the model in the figure. This is specifically necessary for the training and testing of depth estimation using the Co-SemDepth architecture, which requires as input the camera motion data between every two consecutive frames. 

To quantitatively evaluate the depth prediction results, the commonly used evaluation metrics in prior works~\cite{d1, d10, d20} are used. These include the linear root mean square error (\(RMSE\)), the absolute relative error, and accuracy under threshold.

For semantic segmentation, we adopt the commonly used mean Intersection over Union \(mIoU\) metric, with a focus on the per-class \(IoU\) to assess the segmentation performance on each of the 9 classes in our set.

The training and validation plots of the \(RMSE\) and \(mIoU\) metrics for the two models can be found in Figure~\ref{fig:train_curves}. From the plots, it can be noted that while TaskPrompter tends to overfit on the training data (its performance on the training set is better than on the validation set), Co-SemDepth's performance is almost equal on the training and validation data (no overfitting). This can be due to the high capacity of TaskPrompter, which makes it tend to memorize the training data, while Co-SemDepth has a limited number of parameters making it capture only important generic features.

On average, the inference time of Co-SemDepth on a single NVIDIA Quadro P4000 GPU is 49.2 ms per frame, and for TaskPrompter, it is 120.6 ms per frame. This shows that Co-SemDepth is more convenient with regard to speed for real-time applications.

\section{Results}
The experiments of synthetic-to-Real performance are divided into two parts: \textit{Depth Estimation Evaluation}, and \textit{Semantic Segmentation Evaluation}.

\subsection{Depth Estimation Evaluation}

In Table~\ref{table_depth_wuav}, we assess the depth estimation performance of Co-SemDepth and TaskPrompter on the WildUAV dataset. In general, we can notice that the results obtained using Co-SemDepth are notably better than those of TaskPrompter despite the huge difference in the model sizes. This can be due to the parallax estimation modalities implemented in Co-SemDepth that are designed to allow for better generalization of the model in depth estimation.

It can also be noted that, for both models, using MidAir+TopAir for training produces better results than using SkyScenes+SynDrone, and this can be justified by the semantic similarity between the scenes of WildUAV (which are taken in nature) and the natural scenes of MidAir and TopAir. 

In Figure~\ref{fig:qualit_depth_wuav}, a visualization of the predicted output on sample input WildUAV images is depicted. By a qualitative analysis of the figure, it can be validated that Co-SemDepth trained on MidAir+TopAir produces the closest results to the ground truth compared to the other methods.

\begin{table*}[h!]
\begin{center}
\caption{\centering Evaluation of depth estimation performance of joint architectures on WildUAV real dataset. Best results are highlighted. }
\label{table_depth_wuav}
\resizebox{\linewidth}{!}{
\begin{tabular}{ c | c | c | c || c | c | c | c | c }
\hline
 Test Data & Train Data & Method & Params  & RMSE $\downarrow$ & AbsRelErr $\downarrow$ & \(\delta < 1.25\) $\uparrow$ &  \(\delta < 1.25^2\) $\uparrow$ &  \(\delta < 1.25^3\) $\uparrow$\\
\hline
 \parbox[t]{4mm}{\multirow{4}{*}{\rotatebox[origin=c]{90}{WildUAV}}} & \multirow{2}{*}{MidAir+TopAir} & Co-SemDepth & 5.2M  & \textbf{14.9} & \textbf{0.303} & \textbf{58.7}\% & \textbf{80.7}\% & \textbf{91.5}\%\\  
& & TaskPrompter & 126M & 21.26 & 0.459 &  18.5\% &  37.5\% &  54.5\%\\  
\cline{2-9}

 & \multirow{2}{*}{SkyScenes+SynDrone} & Co-SemDepth & 5.2M & 18.9 & 0.388 & 33.78\% & 58.6\% &  73.4\%\\  
 & & TaskPrompter & 126M & 23 & 0.48 &  10.5\% &  25.2\% &  46.2\%\\   
\cline{1-9} 
\end{tabular}
}
\end{center}
\end{table*}

\begin{figure}[h!]%
    \centering
    \includegraphics[width=\linewidth]{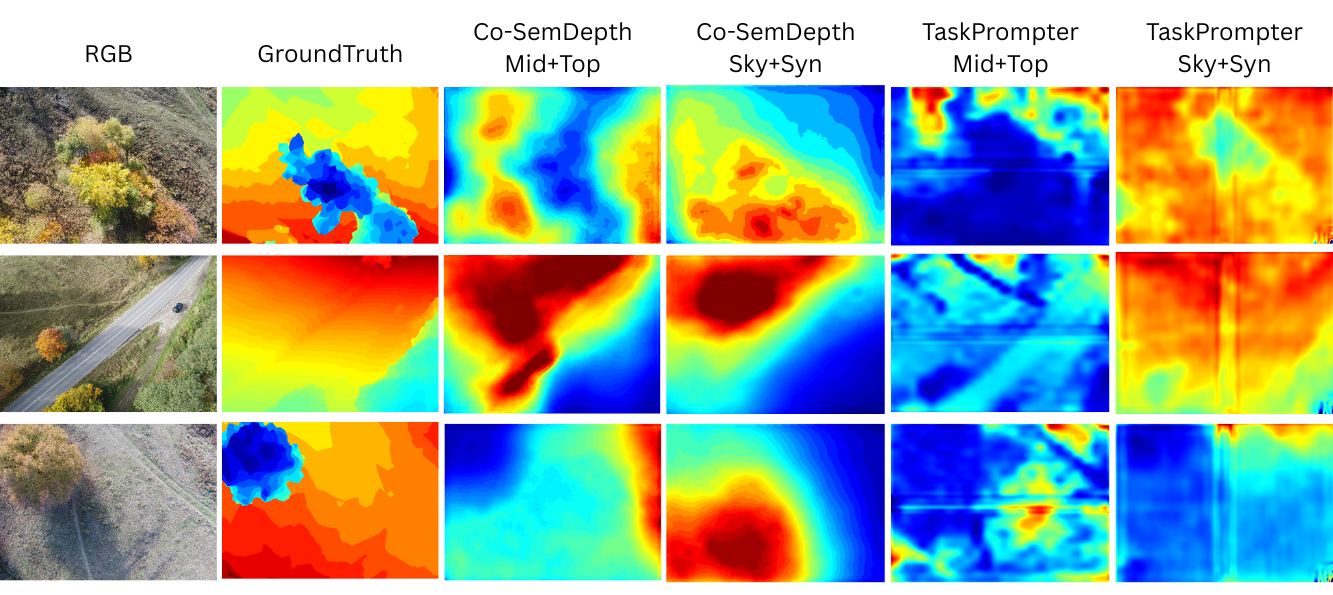}
    \caption{\centering Visualization of the estimated depth maps of sample input images from WildUAV using Co-SemDepth and TaskPrompter (displayed in relative scale where red is the highest distance and blue is the lowest). The output of Co-SemDepth is closer to the ground truth than TaskPrompter, and training on MidAir+TopAir (Mid+Top) is better than SkyScenes+SynDrone (Sky+Syn).}%
    \label{fig:qualit_depth_wuav}%
\end{figure}

\textbf{Assessment of Adding real data to the training:} As explained earlier in this chapter, this assessment is carried out using only TaskPrompter. In the top part of Table~\ref{table_depth_finetune_all}, the depth estimation performance is assessed on WildUAV before and after adding real data (DroneScapes) to the training. 

Despite the low percentage of real to synthetic training data ($4.5-5\%$), a big enhancement is achieved in the results for both MidAir+TopAir and SkyScenes+SynDrone training modalities. This strengthens the findings reported in~\cite{skyscenes,ruralsynth}, where it was demonstrated the positive effect on the model's generalization of adding different percentages of the same real test data to the training. It further adds to the previous findings that such enhancement can be achieved even if the added data is not taken from the same test data distribution (if we use a different real dataset for finetuning). Output visualization of the results can be found in Figure~\ref{fig:qualit_depth_wuav}. It can be seen from the figure that adding DroneScapes to the training makes the network predict more realistic depth maps than those predicted without adding DroneScapes.

In the bottom part of Table~\ref{table_depth_finetune_all}, the depth is evaluated on DroneScapes dataset before and after adding real data to the training. In this case, the performance of the models trained only on synthetic datasets is very low due to the big difference in the depth values between DroneScapes and the datasets used for training, see the depth distributions of the datasets in Figure~\ref{fig:depth_distribution}. 
Despite that, adding a small part of Dronescapes to the training $(4.5-5\%)$ of the model makes a big boost in its performance on the test data.

\begin{table*}[h!]
\begin{center}
\caption{\centering Evaluation of depth estimation performance of TaskPrompter on WildUAV and DroneScape real datasets without and with adding real data to the training.} 
\label{table_depth_finetune_all}
\resizebox{\linewidth}{!}{
\begin{tabular}{ c | c | c | c || c | c | c | c | c }
\hline
 Test Data & \multicolumn{1}{c|}{Method} & \multicolumn{1}{c|}{Train Data} & \multicolumn{1}{c|}{Finetuning Data} & \multicolumn{1}{c|}{RMSE $\downarrow$} & \multicolumn{1}{c|}{AbsRelErr $\downarrow$} & \multicolumn{1}{c|}{\(\delta < 1.25\) $\uparrow$} &  \multicolumn{1}{c|}{\(\delta < 1.25^2\) $\uparrow$} &  \multicolumn{1}{c}{\(\delta < 1.25^3\) $\uparrow$}\\
\hline
\parbox[t]{4mm}{\multirow{4}{*}{\rotatebox[origin=c]{90}{WildUAV}}} & \multirow{2}{*}{TaskPrompter} & \multirow{2}{*}{MidAir+TopAir} & - & 21.26 & 0.459 &  18.49\% &  37.46\% &  54.54\%\\  
 & & & Dronescapes & \textbf{15.87} & \textbf{0.32} & \textbf{31.42}\% &  \textbf{62.39}\% &  \textbf{84.27}\%\\ 
\cline{2-9}
 & \multirow{2}{*}{TaskPrompter} & \multirow{2}{*}{SkyScenes+SynDrone} & - & 23 & 0.48 &  10.5\% &  25.2\% &  46.2\%\\   
 &  & & Dronescapes & 18.05 & 0.354 &  29.4\% &  53.86\% &  73.14\%\\ 
\cline{1-9}
\parbox[t]{4mm}{\multirow{4}{*}{\rotatebox[origin=c]{90}{DroneScapes}}} & \multirow{2}{*}{TaskPrompter} & \multirow{2}{*}{MidAir+TopAir} & - & 96.77 & 0.78 &  0.1\% &  0.6\% &  2.5\%\\ 
 &  &  & Dronescapes & 76.12 & 0.485 &  8.12\% &  34.53\% &  55.15\%\\
\cline{2-9} 
 & \multirow{2}{*}{TaskPrompter} &  \multirow{2}{*}{SkyScenes+SynDrone} & - & 88.7 & 0.694 &  0\% &  0.02\% &  2.15\%\\ 
 &  &  & Dronescapes & \textbf{40.65} & \textbf{0.237} &  \textbf{50.35}\% &  \textbf{80.7}\% &  \textbf{93.49}\%\\   
\hline 

\end{tabular}
}
\end{center}
\end{table*}

\begin{figure}[h!]%
    \centering
    \includegraphics[width=\linewidth]{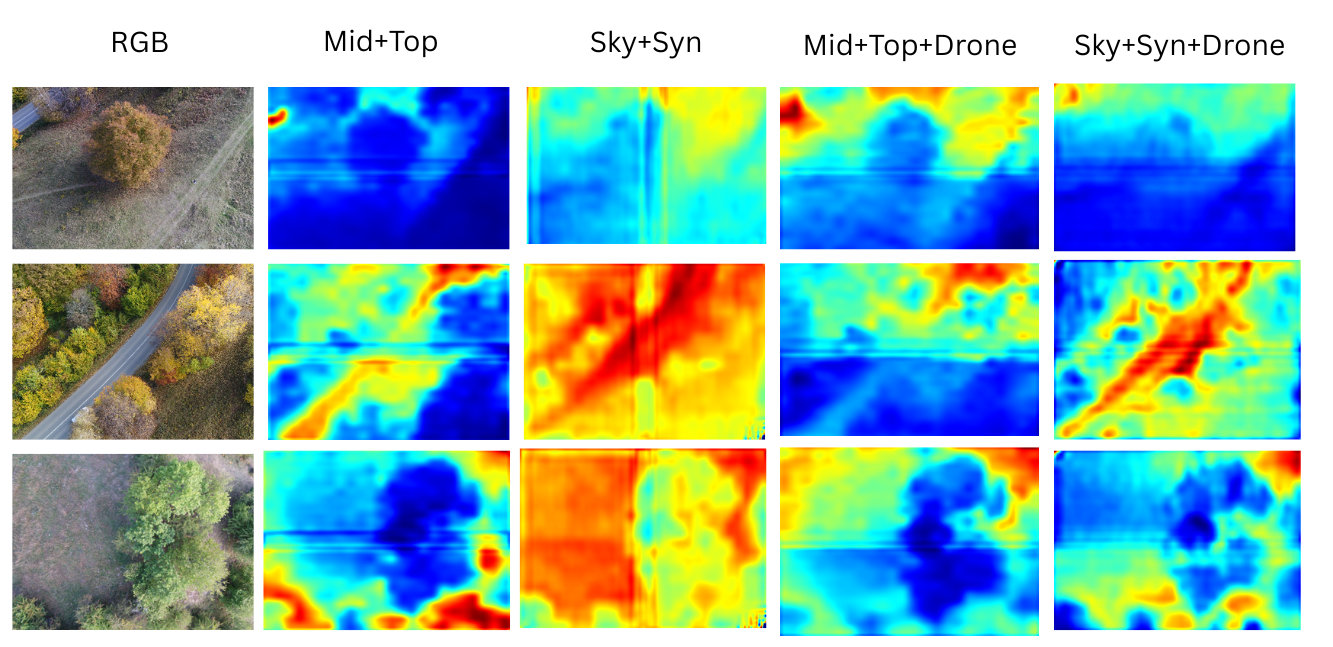}
    \caption{\centering Qualitative visualization of the estimated depth maps using TaskPrompter of sample input images from WildUAV before and after adding DroneScapes to the training (displayed in relative scale where red is the highest distance and blue is the lowest). Adding real data to the training enhanced the results.}%
    \label{fig:qualit_depth_wuav}%
\end{figure}

\begin{figure}[h!]%
    \centering
    \includegraphics[width=\linewidth]{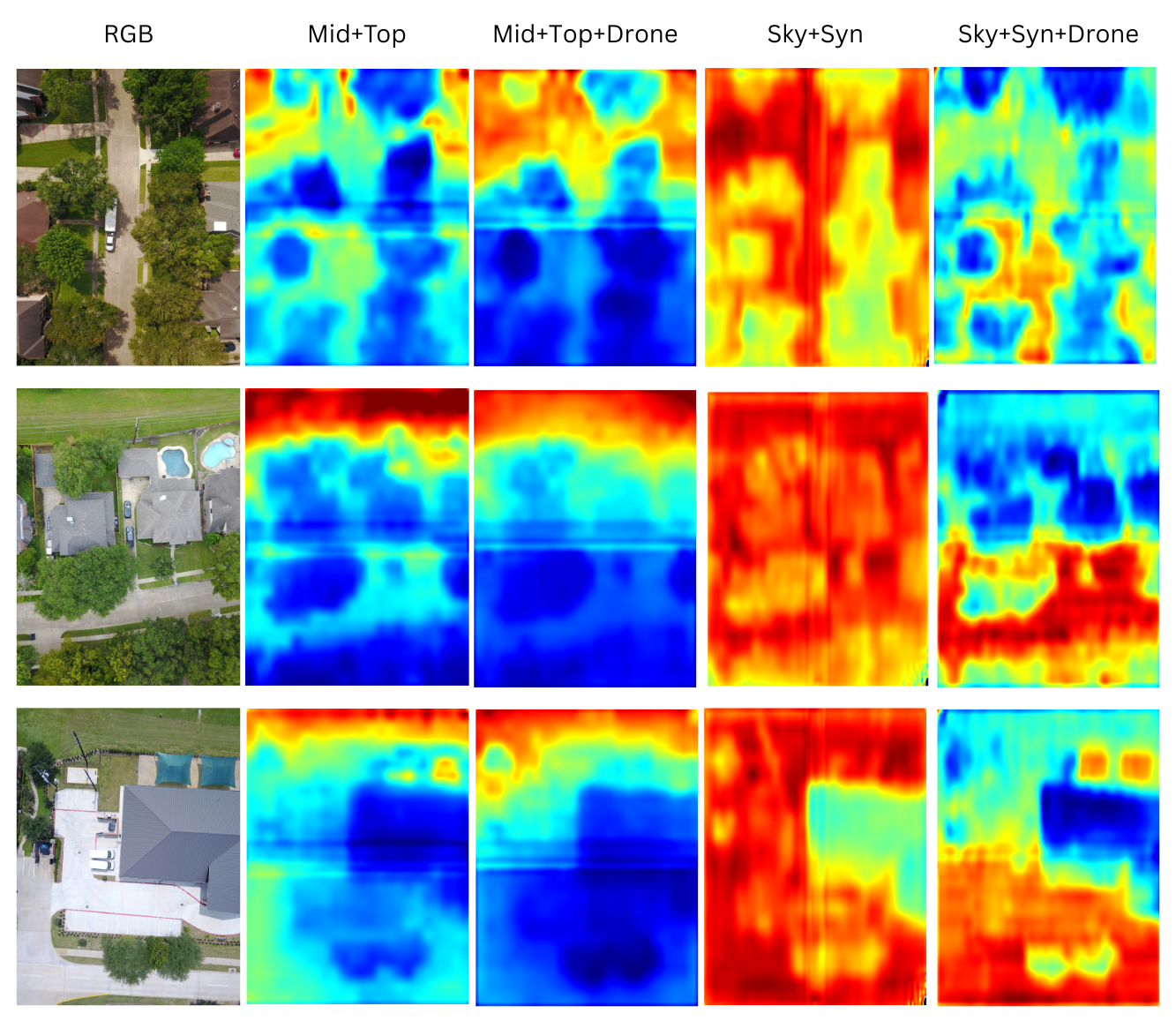}
    \caption{\centering Visualization of the estimated depth maps  (displayed in relative scale where red is the highest distance and blue is the lowest) of sample input images from FSI using TaskPrompter. The model trained on MidAir+TopAir produced the best output. Finetuning on DroneScapes enhanced the results of the model trained on SkyScenes+SynDrone}%
    \label{fig:qualit_depth_fsi}%
\end{figure}

\begin{figure}[h!]%
    \centering
    \includegraphics[width=\linewidth]{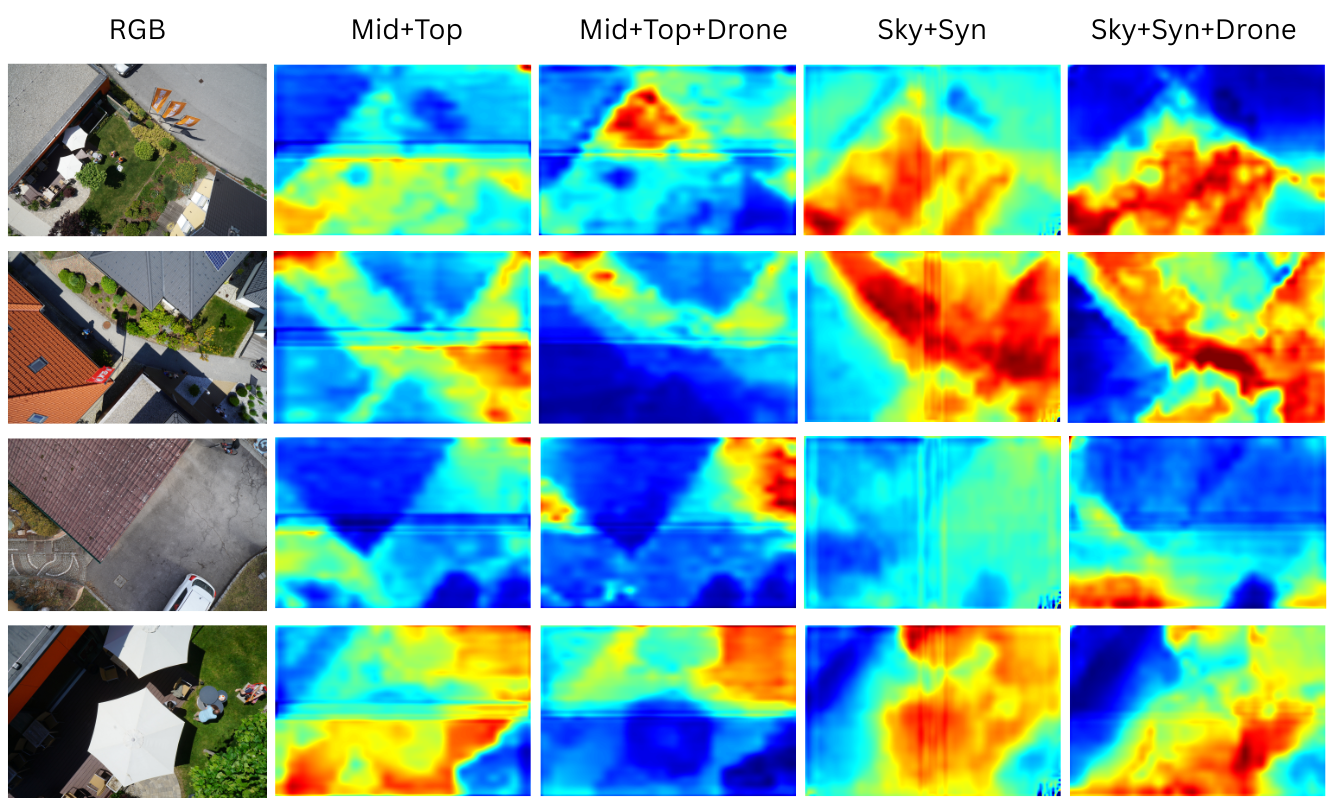}
    \caption{\centering Visualization of the estimated depth maps  (displayed in relative scale where red is the highest distance and blue is the lowest) of sample input images from ICG using TaskPrompter. The output of the model trained on MidAir+TopAir is generally better than the others. Finetuning on DroneScapes did not enhance the results except in the last row (notice the umbrellas).}%
    \label{fig:qualit_depth_icg}%
\end{figure}

The depth estimation is additionally assessed on sample images from the FSI and ICG datasets. However, due to the absence of ground truth depth maps, the analysis is carried out only qualitatively. Only TaskPrompter can be used for making predictions because Co-SemDepth requires camera location data and they are not provided in the mentioned datasets.
In Figures~\ref{fig:qualit_depth_fsi} and~\ref{fig:qualit_depth_icg} it can be found a visualization of the output on sample images from FSI and ICG, respectively. For FSI, it is observed that the model trained on MidAir+TopAir produces the best output. This can be justified by the apparent similarity between FSI scenes and the images captured in the \textit{Neighbourhood} environment in TopAir. It can also be noted that finetuning on real images from DroneScapes enhanced the results of the model trained on SkyScenes+SynDrone. For ICG, it can be observed by eye that the output of the model trained on MidAir+TopAir is generally better than the one trained on SkyScenes+SynDrone. Unfortunately, finetuning on DroneScapes does not provide considerable enhancements on the outputs in this case.

\subsection{Semantic Segmentation Evaluation}
The semantic segmentation performance of the two models is assessed in Table~\ref{table_per_class_all} on the real datasets: WildUAV, FSI, UDD, DroneScapes, RuralScapes, AeroScapes, and ICG. From the table, it can be noted that, regarding the model architecture, TaskPrompter in general is better in semantic segmentation than Co-SemDepth. This can be due to TaskPrompter's large capacity that makes it more capable of capturing the small semantic details. Specifically, it is remarkably better at segmenting the small objects ("Vehicle" and "Others"). 

Regarding the training data, depending on the appearance of the objects and the layout of the scenes in the datasets, the class segmentation in some datasets can be better using MidAir+TopAir for training or SkyScenes+SynDrone in other cases. In particular, the UDD dataset contains only urban scenes, and this makes it more similar to SkyScenes+SynDrone; hence the models trained on SkyScenes+SynDrone give higher accuracy on UDD than the ones trained on MidAir+TopAir. On the other hand, the WildUAV dataset is more similar to MidAir+TopAir due to its natural scenes that are similar to the ones present in MidAir and TopAir; hence, the models trained on MidAir+TopAir perform better in this case than the ones trained on SkyScenes+SynDrone. On FSI, the model trained on MidAir+TopAir is more capable in segmenting "Water", "Trees", and "Building", while the model trained on SkyScenes+SynDrone is better in segmenting "Land", "Vehicle", and "Road".  On RuralScapes and DroneScapes, the model trained on MidAir+TopAir achieves higher accuracy on the segmentation of "Sky", "Water", and "Land", and the model trained on SkyScenes+SynDrone performs better in the segmentation of "Vehicle", "Road", and "Building". On AeroScapes, training on MidAir+TopAir gives the highest accuracy in class "Land" and "Building", while training on SkyScenes+SynDrone is better in the other classes. The results are also class-dependent between the two training modalities for ICG. Therefore, what we can conclude from the above is that \textbf{there is no specific training dataset that gives the best results in all the cases}. Instead, the synthetic-to-real performance is dependent on the appearance of the objects belonging to different classes in the target test data and how similar they are with the objects in the synthetic training data.

\begin{figure}[h!]%
    \centering
    \includegraphics[width=\linewidth]{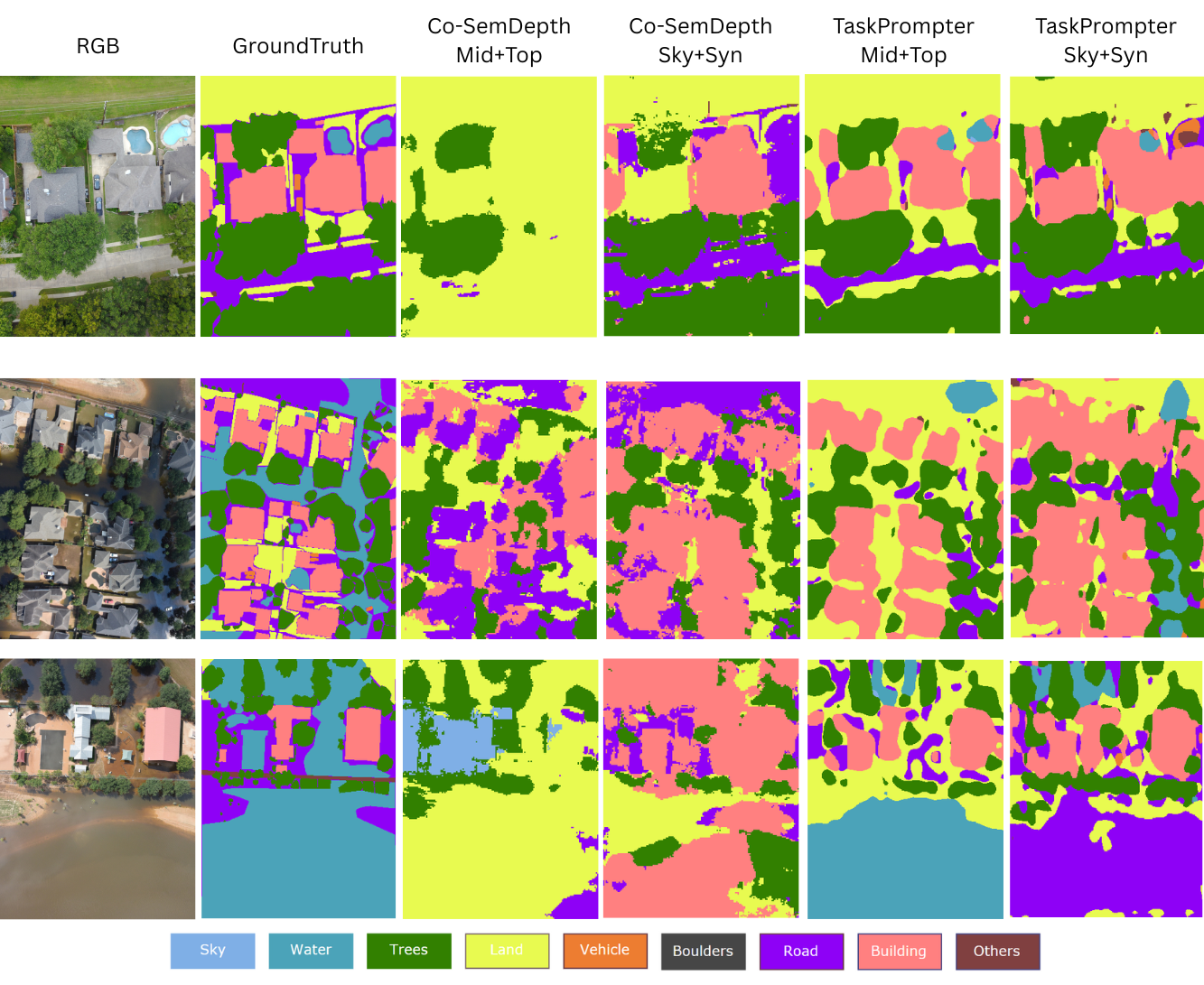}
    \caption{\centering Qualitative visualization of the estimated depth maps of sample input images from FSI using different training modalities and architectures. It can be seen that using TaskPrompter trained on MidAir+TopAir gives the closest results to the ground truth. TaskPrompter trained on SkyScenes+SynDrone is the best in the detection of cars (top row).}%
    \label{fig:sem_fsi}%
\end{figure}

A visualization of the output semantic segmentation maps on sample images from FSI can be found in Figure~\ref{fig:sem_fsi}. It can be seen that, in general, the models trained on SkyScenes+SynDrone tend to segment most of the pixels as either "Building" or "Road" (see the second and third rows) due to the urban nature of the training data. On the other hand, the models trained on MidAir+TopAir tend to the "Land" and "Trees" classes in the segmentation (see the first and third rows). It can also be noted the big performance gap in the semantic segmentation by varying the model architecture, using TaskPrompter compared to Co-SemDepth, trained on the same datasets.

\begin{figure}[h!]%
    \centering
    \includegraphics[width=\linewidth]{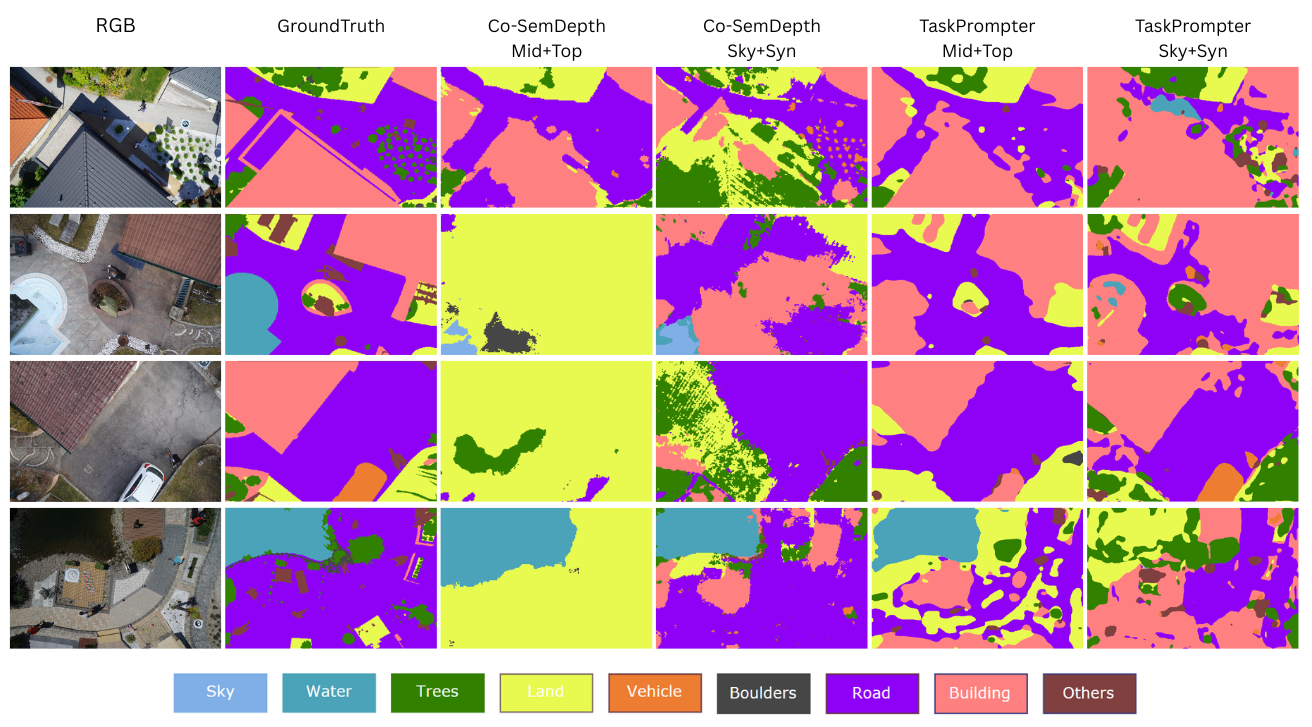}
    \caption{\centering Qualitative visualization of the estimated segmentation maps of sample input images from ICG using different training modalities. Best performing method is TaskPrompter trained on MidAir+TopAir except in the detection of cars (third row) where TaskPrompter trained on SkyScenes+SynDrone is better.}%
    \label{fig:sem_icg}%
\end{figure}

 A visualization of the output semantic segmentation maps on sample images from ICG can be found in Figure~\ref{fig:sem_icg}. It can be noted how the semantic segmentation using TaskPrompter is much better than using Co-SemDepth on this dataset. Additionally, by comparing the output of TaskPrompter trained on MidAir+TopAir and TaskPrompter trained on SkyScenes+SynDrone, we can say that using SkyScenes+SynDrone is better in the third row where the vehicle is present in the image; however, in the fourth row, using MidAir+TopAir makes the network better detect and segment the water area. In the first row, MidAir+TopAir is better at segmenting the building.

\begin{table*}[h!]
\begin{center}
\caption{\centering Per-Class IoU segmentation evaluation of the joint architectures on a variety of real data. Best results for each dataset are highlighted. }
\label{table_per_class_all}
\resizebox{\linewidth}{!}{
\begin{tabular}{ c | c | c | c || c | c | c | c | c | c | c | c | c }
\hline
  Test Data & \multicolumn{1}{c|}{Train Data} & \multicolumn{1}{c|}{Method} & \multicolumn{1}{c|}{Params} &  \multicolumn{1}{c|}{Sky} & \multicolumn{1}{c|}{Water} & \multicolumn{1}{c|}{Trees} & \multicolumn{1}{c|}{Land} & \multicolumn{1}{c|}{Vehicle} & \multicolumn{1}{c|}{Rocks}& \multicolumn{1}{c|}{Road} & \multicolumn{1}{c|}{Building} & \multicolumn{1}{c}{Others}\\
 \hline
 
\hline
\parbox[t]{4mm}{\multirow{4}{*}{\rotatebox[origin=c]{90}{WildUAV}}} & \multirow{2}{*}{MidAir+TopAir} & Co-SemDepth & 5.2M &  - &  0.93\% & 32.8\% & 34.76\% & 0\% &  - &  14.9\% &  - &  0\%  \\ 
 & & TaskPrompter & 126M  & - &  61.66\% &  \textbf{55.36}\% &  \textbf{67.63}\% &  1.65\% & - & \textbf{53.5}\% &  - &  3.1\% \\ 
 \cline{2-13} 
 & \multirow{2}{*}{SkyScenes+SynDrone} & Co-SemDepth & 5.2M  & - &  17.63\% &  4.2\% & 14.1\% & 1.2\% &  - &  1.8\% &  - &  0.04\%  \\
 & & TaskPrompter & 126M  & - &  \textbf{62.1}\% &  51\% &  62.05\% &  \textbf{46.2}\% & - & 45.9\% &  - &  \textbf{5.9}\% \\ 
\hline
\hline

\parbox[t]{4mm}{\multirow{4}{*}{\rotatebox[origin=c]{90}{FSI}}} & \multirow{2}{*}{MidAir+TopAir} & Co-SemDepth & 5.2M  & - &  10.1\% &  40.6\% &  34.67\% &  0\% & - & 9.14\% &  1.43\% &  0.01\% \\ 


& & TaskPrompter & 126M  & - &  \textbf{49.57}\% &  \textbf{67.36}\% &  63.7\% &  3.32\% & - & 34.1\% &  \textbf{59.83}\% &  1.73\% \\ 

\cline{2-13}
 & \multirow{2}{*}{SkyScenes+SynDrone} & Co-SemDepth & 5.2M  &  - &  6.31\% & 38.25\% & 36.54\% & 4.07\% &  - &  31.65\% &  17.12\% &  0.35\% \\
 & & TaskPrompter & 126M & - &  32.5\% &  63.6\% &  \textbf{64.1}\% &  \textbf{19.76}\% & - & \textbf{37.8}\% &  56.83\% &  \textbf{3.9}\% \\ 
\hline
\hline

\parbox[t]{4mm}{\multirow{4}{*}{\rotatebox[origin=c]{90}{UDD}}} & \multirow{2}{*}{MidAir+TopAir} & Co-SemDepth & 5.2M &  - &  - & 37.2\% & 17.1\% & 0\% &  - &  9.9\% &  1.7\% &  - \\ 
 &  & TaskPrompter & 126M  & - &  - &  56.52\% &  \textbf{13.02}\% &  3.16\% & - & 48.53\% &  70.75\% &  - \\
\cline{2-13}
 & \multirow{2}{*}{SkyScenes+SynDrone} & Co-SemDepth & 5.2M  & - &  - &  46.99\% & 11.24\% & 1.62\% &  - &  24.96\% &  34.01\% &  - \\ 
 &  & TaskPrompter & 126M  & - & - &  \textbf{76.32}\% &  8.9\% &  \textbf{18.4}\% & - & \textbf{49.77}\% &  \textbf{73.1}\% &  - \\
\hline
\hline

\parbox[t]{4mm}{\multirow{4}{*}{\rotatebox[origin=c]{90}{DroneScapes}}} & \multirow{2}{*}{MidAir+TopAir} & Co-SemDepth & 5.2M  &  43.88\% &  0\% &  15.66\% &  37.13\% &  0.5\% & - & 10.4\% &  10.9\% &  - \\ 
& & TaskPrompter & 126M &  \textbf{95.1}\% &  \textbf{10.15}\% &  37.25\% &  \textbf{53.7}\% &  0.8\% & - & 22.5\% &  48.5\% &  - \\ 
\cline{2-13}
 & \multirow{2}{*}{SkyScenes+SynDrone} & Co-SemDepth & 5.2M &  50.74\% &  0\% &  39.56\% &  23.87\% &  4.52\% & - & 19.3\% & 24.8\% &  - \\ 
 &  & TaskPrompter & 126M & 70.27\% &  5.77\% &  \textbf{55.79}\% &  53.5\% &  \textbf{15.04}\% & - & \textbf{43.1}\% &  \textbf{68.15}\% &  - \\ 
\hline
\hline

\parbox[t]{4mm}{\multirow{4}{*}{\rotatebox[origin=c]{90}{RuralScapes}}} & \multirow{2}{*}{MidAir+TopAir} & Co-SemDepth & 5.2M & 61.3\% &  0\% & 39.55\% & 26.7\% & 0\% &  - &  2.42\% &  0.76\% &  0\% \\ 

&  & TaskPrompter & 126M & \textbf{92.7}\% &  \textbf{9.13}\% &  40.14\% &  \textbf{42.12}\% &  0.2\% & - & 23.1\% &  31.45\% &  6.8\% \\ 

\cline{2-13}
  & \multirow{2}{*}{SkyScenes+SynDrone} & Co-SemDepth & 5.2M & 69.5\% &  0.45\% &  30.1\% & 23.84\% & 1.1\% &  - &  10.65\% &  21.76\% &  0.97\% \\ 
 &  & TaskPrompter & 126M & 86.02\% &  3.99\% &  \textbf{49.7}\% &  30.24\% &  \textbf{10.2}\% & - & \textbf{33.98}\% &  \textbf{56.96}\% &  \textbf{8.7}\% \\ 
\hline
\hline

\parbox[t]{4mm}{\multirow{4}{*}{\rotatebox[origin=c]{90}{AeroScapes}}} & \multirow{2}{*}{MidAir+TopAir} & Co-SemDepth & 5.2M &  77.04\% &  - & 33.3\% & 24.89\% & 1.3\% &  - &  11.2\% &  3.63\% &  0.3\% \\ 
 & & TaskPrompter & 126M & 92.7\% &  - &  37.84\% &  \textbf{17.75}\% &  8.86\% & - & 63.73\% &  \textbf{35.7}\% &  14.1\% \\ 
\cline{2-13}
 & \multirow{2}{*}{SkyScenes+SynDrone} & Co-SemDepth & 5.2M & 60.56\% &  - &  46.4\% & 10.63\% & 6.7\% &  - &  43.6\% &  12.85\% &  1.7\% \\ 
 & & TaskPrompter & 126M & \textbf{93.83}\% &  - &  \textbf{57.11}\% &  13.04\% &  \textbf{34.99}\% & - & \textbf{71.8}\% &  31.3\% &  \textbf{23.9}\% \\ 
\hline
\hline

\parbox[t]{4mm}{\multirow{4}{*}{\rotatebox[origin=c]{90}{ICG}}} & \multirow{2}{*}{MidAir+TopAir} & Co-SemDepth & 5.2M & - &  10.62\% &  12.7\% &  20.99\% &  0.13\% & 0.44\% & 18.6\% &  12.52\% &  0.08\% \\ 
 &  & TaskPrompter & 126M & - &  \textbf{30.15}\% &  27.8\% &  44.67\% &  3.04\% & \textbf{3.4}\% & 53.83\% &  \textbf{46.53}\% &  16.2\% \\

\cline{2-13}
 & \multirow{2}{*}{SkyScenes+SynDrone} & Co-SemDepth & 5.2M &  - & 5.2\% & 27.3\% & 20.5\% & 3.82\% & 0\% &  52.3\% &  17.65\% &  1.4\% \\ 
 &  & TaskPrompter & 126M & - &  14.54\% &  \textbf{45.52}\% &  \textbf{61.27}\% &  \textbf{45.8}\% & 0\% & \textbf{61.7}\% &  35.04\% &  \textbf{18.88}\% \\
\hline

\end{tabular}
}
\end{center}
\end{table*}

\begin{figure}[h!]%
    \centering
    \includegraphics[width=\linewidth]{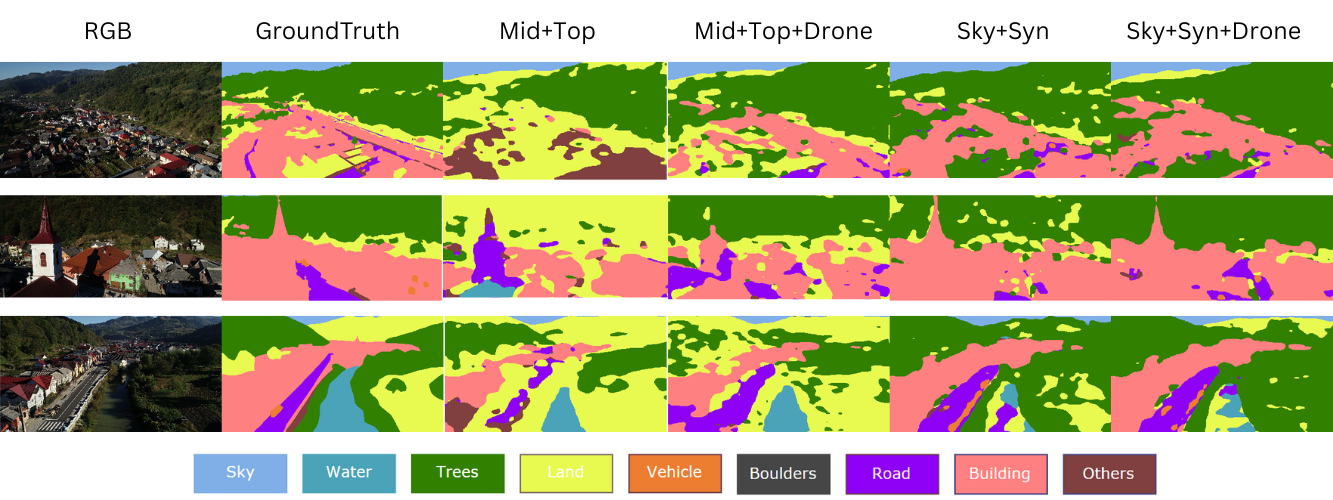}
    \caption{\centering Qualitative visualization of the estimated semantic segmentation maps of sample input images from RuralScapes using TaskPrompter before and after adding real data (DroneScapes) to the training}%
    \label{fig:qualit_rural}%
\end{figure}

\textbf{Assessment of Adding real data to the training:}
As explained earlier in this chapter, this assessment is carried out using only the TaskPrompter model. 
The effect of adding real data to the training is assessed in Table~\ref{table_per_class_all_finetune} on WildUAV, RuralScapes, Dronescapes, FSI, UDD, AeroScapes, and ICG datasets.
Here, due to the gap in the appearance of some classes between the real data used in the training (DroneScapes) and the testing data, the effect of adding real data is not always positive, as it can be observed from the table. As a positive example, the environments of RuralScapes look similar to DroneScapes, and for this reason, adding DroneScapes to the training enhanced the semantic segmentation accuracy on most of the classes in RuralScapes, see Figure~\ref{fig:qualit_rural} for a visualization of the output. In WildUAV, instead, there is an enhancement in the detection of "Water", "Trees", and "Land", while the accuracy of "Vehicle" and "Road" is decreased. Datasets like AeroScapes and ICG look completely different than DroneScapes, and this caused a drop in the accuracy after finetuning the model on DroneScapes. Instead, on DroneScapes, the accuracy on all classes is boosted after the finetuning. On FSI, the accuracy is enhanced on the segmentation of "Water", "Vehicle", and "Road", and the accuracy on the other classes is slightly lower after finetuning. On UDD, only the segmentation of "Vehicle" and "Building" was enhanced after the fine-tuning.

\begin{table*}[h!]
\begin{center}
\caption{\centering Per-Class IoU segmentation evaluation of joint architectures on WildUAV and RuralScapes without and with adding real data to the training.}
\label{table_per_class_all_finetune}
\resizebox{\linewidth}{!}{
\begin{tabular}{ c | c | c | c || c | c | c | c | c | c | c | c | c }
\hline
 Test Data & Train Data & Finetuning Data & Method & Sky & Water & Trees & Land & Vehicle & Rocks & Road & Building & Others \\
\hline
\parbox[t]{4mm}{\multirow{4}{*}{\rotatebox[origin=c]{90}{WildUAV}}} & \multirow{2}{*}{MidAir+TopAir} & - & \multirow{2}{*}{TaskPrompter} & - &  61.66\% &  55.36\% &  67.63\% &  1.65\% & - & 53.5\% &  - &  3.1\% \\
&  & Dronescapes &  & - &  59.36\% &  56.88\% & 77.54\% &  7.62\% & - & 39.12\% & - &  2.6\% \\
\cline{2-13}
& \multirow{2}{*}{SkyScenes+SynDrone} & - & \multirow{2}{*}{TaskPrompter} & - &  62.1\% &  51\% &  62.05\% &  46.2\% & - & 45.9\% &  - &  5.9\% \\
&  & Dronescapes &  & - &  63.9\% &  47.77\% &  55.66\% &  22.89\% & - & 45.26\% &  - &  5.5\% \\
\cline{1-13}

\parbox[t]{4mm}{\multirow{4}{*}{\rotatebox[origin=c]{90}{RuralScapes}}} & \multirow{2}{*}{MidAir+TopAir} & - & \multirow{2}{*}{TaskPrompter} & 92.7\% &  9.13\% &  40.14\% &  42.12\% &  0.2\% & - & 23.1\% &  31.45\% &  6.8\% \\
& & Dronescapes &  & 89.61\% &  14.07\% &  46.8\% &  47.65\% &  2.1\% & - & 33.6\% &  46.93\% &  7.8\% \\
\cline{2-13}
& \multirow{2}{*}{SkyScenes+SynDrone} & - & \multirow{2}{*}{TaskPrompter} & 86.02\% &  3.99\% &  49.7\% &  30.24\% &  10.2\% & - & 33.98\% &  56.96\% &  8.7\% \\
& & Dronescapes &  & 91.1\% &  7.75\% &  50.3\% &  30.02\% &  9.17\% & - & 39.53\% &  66.52\% &  4.5\% \\
\cline{1-13}

\parbox[t]{4mm}{\multirow{4}{*}{\rotatebox[origin=c]{90}{DroneScapes}}} & \multirow{2}{*}{MidAir+TopAir} & - & \multirow{2}{*}{TaskPrompter} &  95.1\% &  10.15\% &  37.25\% &  53.7\% &  0.8\% & - & 22.5\% &  48.5\% &  - \\
 & & Dronescapes &  & \textbf{95.16}\% &  \textbf{49.98}\% &  51.06\% &  \textbf{55.94}\% &  10.2\% & - & 42.83\% &  69.1\% &  - \\

\cline{2-13}
 & \multirow{2}{*}{SkyScenes+SynDrone} & - & \multirow{2}{*}{TaskPrompter} & 70.27\% &  5.77\% &  55.79\% &  53.5\% &  15.04\% & - & 43.1\% &  68.15\% &  - \\
 & & Dronescapes &  & 78.98\% &  35.12\% &  \textbf{56.37}\% &  52.36\% &  \textbf{26.46}\% & - & \textbf{45.55}\% &  \textbf{75.3}\% &  - \\
\cline{1-13}

\parbox[t]{4mm}{\multirow{4}{*}{\rotatebox[origin=c]{90}{FSI}}} & \multirow{2}{*}{MidAir+TopAir}  & - & \multirow{2}{*}{TaskPrompter} & - &  49.57\% &  \textbf{67.36}\% &  63.7\% &  3.32\% & - & 34.1\% &  \textbf{59.83}\% &  1.73\% \\
 &  & Dronescapes &  & - &  \textbf{52.1}\% &  58.2\% &  61.8\% &  5.3\% & - & \textbf{37.9}\% &  57.3\% &  1.95\% \\

\cline{2-13}
 & \multirow{2}{*}{SkyScenes+SynDrone} & - & \multirow{2}{*}{TaskPrompter} & - &  32.5\% &  63.6\% &  \textbf{64.1}\% &  19.76\% & - & 37.8\% &  56.83\% &  \textbf{3.9}\% \\
 & & Dronescapes &  & - &  39\% &  61.04\% &  60.8\% &  \textbf{21.3}\% & - & 35.6\% &  54.3\% &  3.01\% \\
\cline{1-13}

\parbox[t]{4mm}{\multirow{4}{*}{\rotatebox[origin=c]{90}{UDD}}} & \multirow{2}{*}{MidAir+TopAir}  & - & \multirow{2}{*}{TaskPrompter} & - &  - &  56.52\% &  \textbf{13.02}\% &  3.16\% & - & 48.53\% &  70.75\% &  - \\
 & & Dronescapes &  & - & - &  50.97\% &  11.62\% &  2.95\% & - & 41.1\% &  62.6\% & - \\
\cline{2-13}
 & \multirow{2}{*}{SkyScenes+SynDrone} & - & \multirow{2}{*}{TaskPrompter} & - & - &  \textbf{76.32}\% &  8.9\% &  18.4\% & - & \textbf{49.77}\% &  73.1\% &  - \\
 & & Dronescapes &  & - &  - &  75.55\% &  6.1\% &  \textbf{19.97}\% & - & 45.1\% &  \textbf{74.2}\% & - \\
\cline{1-13}

\parbox[t]{4mm}{\multirow{4}{*}{\rotatebox[origin=c]{90}{AeroScapes}}} & \multirow{2}{*}{MidAir+TopAir} & - & \multirow{2}{*}{TaskPrompter} & 92.7\% &  - &  37.84\% &  17.75\% &  8.86\% & - & 63.73\% &  \textbf{35.7}\% &  14.1\% \\
 &  & Dronescapes &  & 93.5\% &  - &  47.42\% &  \textbf{18.96}\% &  11.3\% & - & 63.1\% &  33.74\% &  12.6\% \\

\cline{2-13}
 & \multirow{2}{*}{SkyScenes+SynDrone} & - & \multirow{2}{*}{TaskPrompter} & \textbf{93.83}\% &  - &  \textbf{57.11}\% &  13.04\% &  \textbf{34.99}\% & - & \textbf{71.8}\% &  31.3\% &  \textbf{23.9}\% \\
 &  & Dronescapes &  & 93.34\% &  - &  55.4\% &  16.42\% &  30.3\% & - & 63.2\% &  26.34\% &  20.34\% \\
\cline{1-13}

\parbox[t]{4mm}{\multirow{4}{*}{\rotatebox[origin=c]{90}{ICG}}} & \multirow{2}{*}{MidAir+TopAir} & - & \multirow{2}{*}{TaskPrompter} & - &  30.15\% &  27.8\% &  44.67\% &  3.04\% & \textbf{3.4}\% & 53.83\% &  \textbf{46.53}\% &  16.2\% \\
 & & Dronescapes &  & - &  \textbf{31.58}\% &  30.3\% &  50.8\% &  4.6\% & 1.66\% & 56.54\% &  43\% &  8.6\% \\

\cline{2-13}
& \multirow{2}{*}{SkyScenes+SynDrone} & - & \multirow{2}{*}{TaskPrompter} & - &  14.54\% &  \textbf{45.52}\% &  \textbf{61.27}\% &  \textbf{45.8}\% & 0\% & \textbf{61.7}\% &  35.04\% &  \textbf{18.88}\% \\
 &  & Dronescapes &  & - &  12.72\% &  39.3\% &  45.86\% &  37.12\% & 0\% & 58.03\% &  34.95\% &  14.6\% \\
\cline{1-13}

\end{tabular}
}
\end{center}
\end{table*}

\subsection{3D Reconstruction}
\begin{figure}[h!]%
    \centering
    \includegraphics[width=\linewidth]{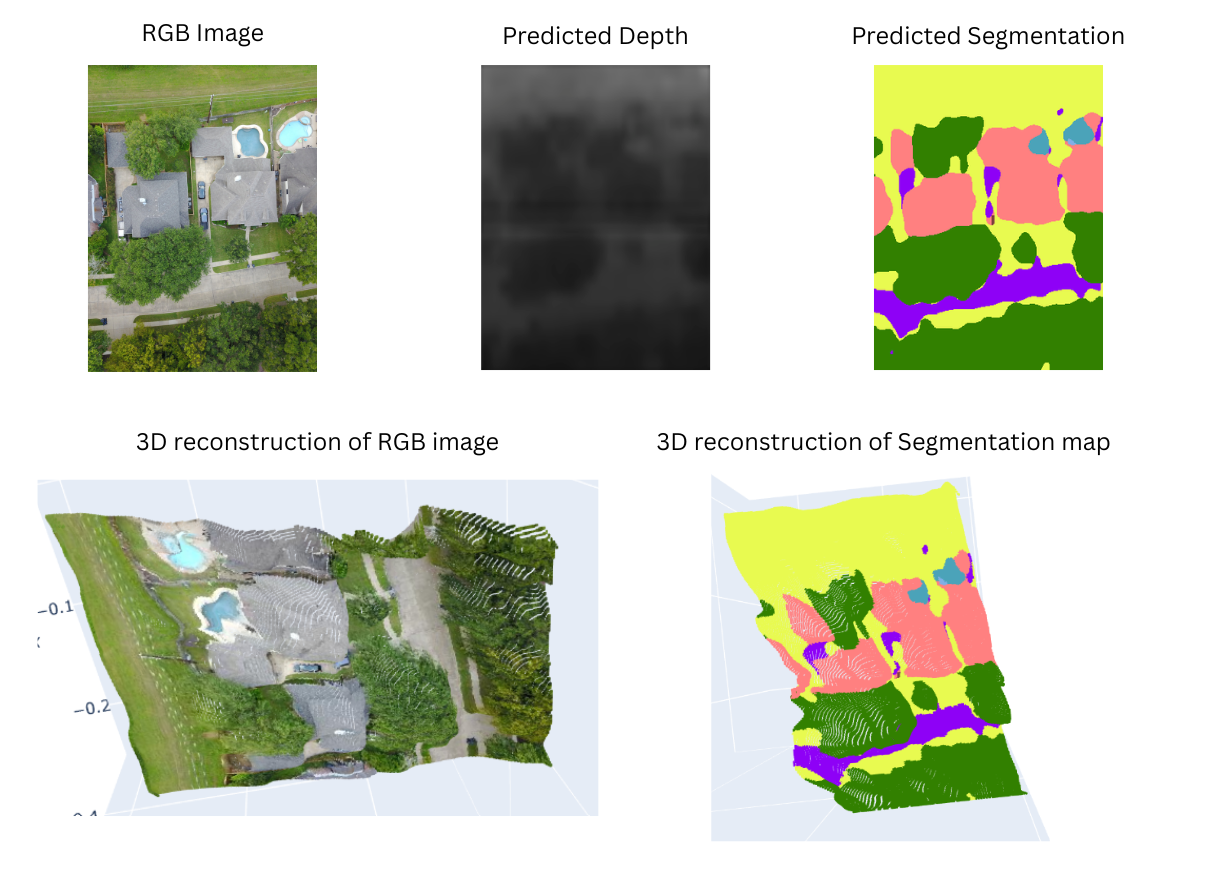}
    \caption{\centering 3D semantic map reconstruction using the predicted depth and segmentation maps using TaskPrompter (trained on MidAir $+$ TopAir) on a sample image from the FSI dataset. Depth is truncated at 200m.}%
    \label{fig:3d_reconstruct1}%
\end{figure}

\begin{figure}[h!]%
    \centering
    \includegraphics[width=\linewidth]{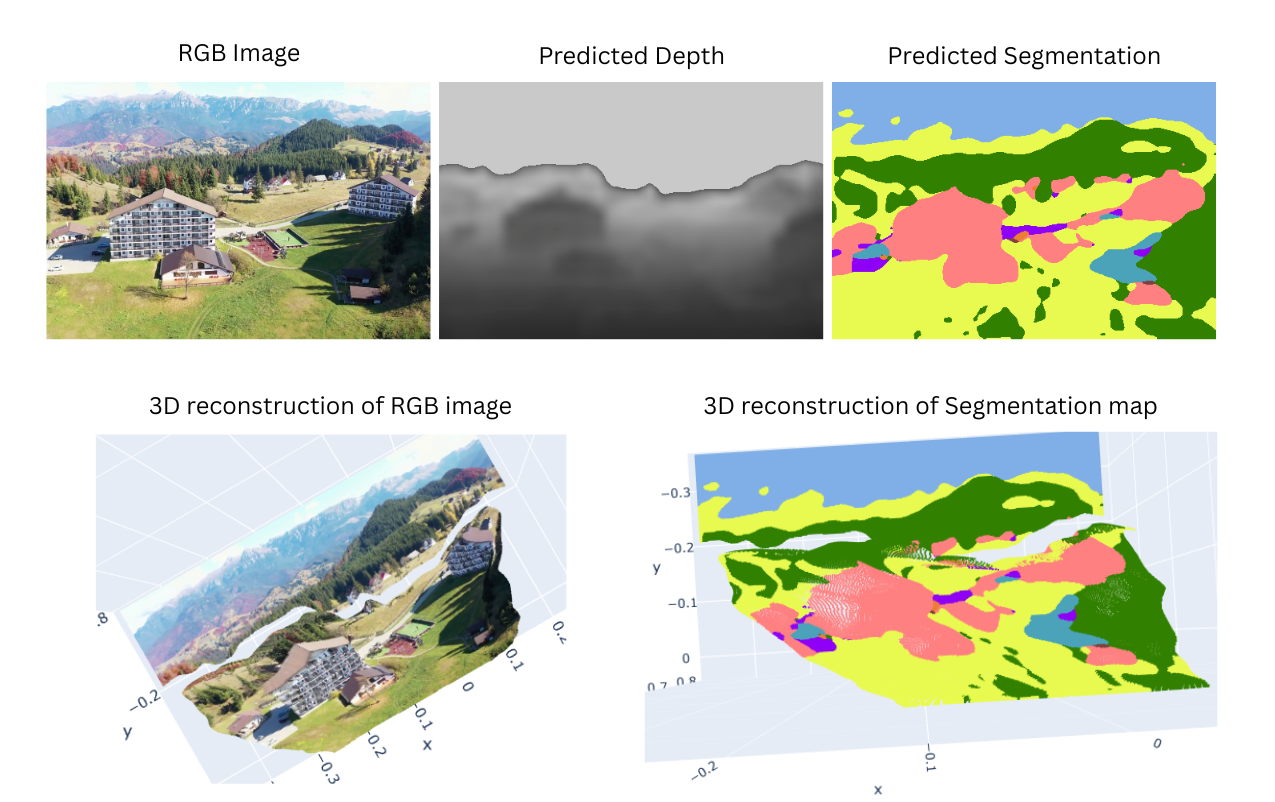}
    \caption{\centering 3D semantic map reconstruction using the predicted depth and segmentation maps using TaskPrompter (trained on SkyScenes $+$ SynDrone $+$ DroneScapes) on a sample image from DroneScapes test data. Depth is truncated at 200m}%
    \label{fig:3d_reconstruct2}%
\end{figure}

For a further qualitative assessment of the output, we resort to the Open3D library ~\cite{open3d} to construct a 3D RGB and semantic map of the scene using the predicted 2D depth and semantic maps. Figures~\ref{fig:3d_reconstruct1} and~\ref{fig:3d_reconstruct2} show sample outputs constructed using the predicted depth and semantic maps of TaskPrompter on sample input images from DroneScapes and FSI datasets. First, an RGB-D image is created using Open3D from the input segmentation map (or RGB image) and depth map. Then, a point cloud is constructed from the RGB-D image using the Pinhole camera default intrinsic parameters of Open3D. Finally, the point cloud is plotted using the visualization tool of the library. The constructed 3D maps can be particularly useful in 3D SLAM navigation of UAVs.

\chapter{Exploring Image-Style Transfer}
\label{chap:style}

\ifpdf
    \graphicspath{{Chapter5/Figures/Raster/}{Chapter5/Figures/PDF/}{Chapter5/Figures/}}
\else
    \graphicspath{{Chapter5/Figures/Vector/}{Chapter5/Figures/}}
\fi

\section*{Summary}
In this chapter, image-style transfer techniques are explored with the aim of transferring from the synthetic style to the real style for aerial images. The adopted methods are Cycle-GAN and Diffusion models. We first give a brief recount of these methods and explain the main principle behind each of them. After that, we present our experimental setup and the parameters used during the training of both methods. In the end, the preliminary results we obtained for the style transfer of images from \textit{TopAir} and \textit{SynDrone} are discussed. While Cycle-GAN focused on altering the colors of the input images to make them look realistic, the Diffusion model changed the content in the input image with realistic objects. Unfortunately, diffusion models change the semantic layout of the input images. 
However, the obtained results open the door for further research in that area.

\section{Adopted Methods}


For further exploration, we try adopting image style transfer techniques for converting the appearance of the synthetic images to a realistic style. For this purpose, we use image generative models: Cycle-GANs (based on Generative Adversarial Networks GAN)~\cite{cyclegan} and InST (based on Diffusion models)~\cite{inst}. In Figures~\ref{fig:cyclegan} and~\ref{fig:inst}, the overview of the Cycle-GAN and InST methods is depicted, respectively.

\subsection{Cycle-GAN}
\begin{figure}[h!]%
    \centering

  \includegraphics[width=\linewidth]{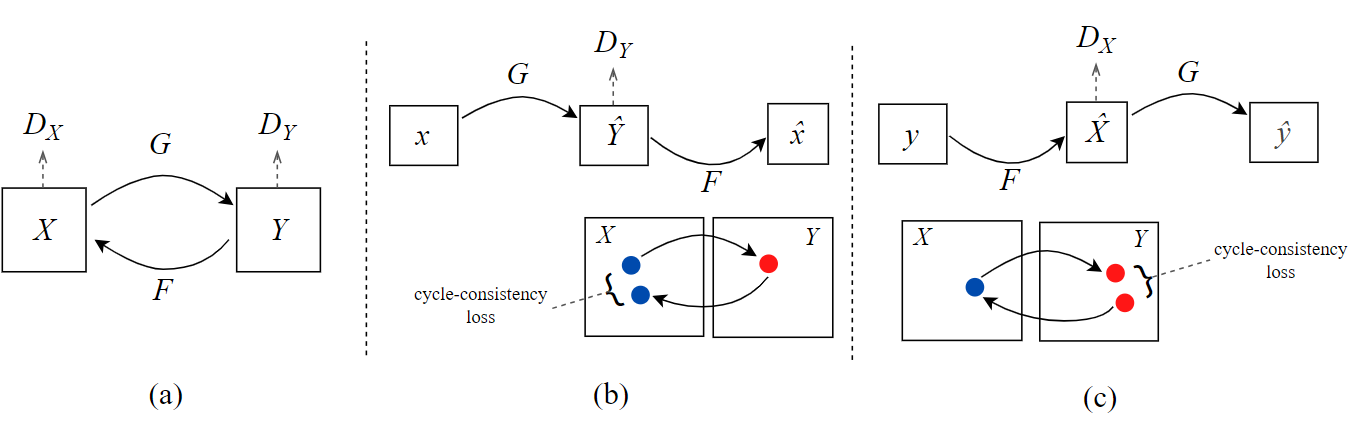}
  \caption{\centering Overview of the Cycle-GAN method for image style transfer~\cite{cyclegan}. The process is composed of two mapping functions $G$ and $F$ (a). The loss used for training of the networks is the addition of two losses: forward cycle-consistency loss (b), and backward cycle-consistency loss (c).}
    \label{fig:cyclegan}%
\end{figure}

In Cycle-GAN, the goal is to learn the mapping between one style of images and the other without the need to align image pairs for training, because it is difficult to obtain such aligned image pairs in reality. The goal is to train the network using a group of images belonging to the source domain $\{x_i\}^{N}_{i=1}$ and another group of images belonging to the target domain $\{y_j\}^{M}_{j=1}$. While the well-known pix2pix GAN model proposed in~\cite{pix2pix} produced remarkable results in image style transfer, it requires aligned image pairs for model training. Instead, Cycle-GAN overcomes such a requirement by the incorporation of a cycle-consistency loss. As depicted in Figure~\ref{fig:cyclegan}, the method is composed of the synchronous training of two generative networks $G$ and $F$. $G$ learns the mapping from domain $X$ to domain $Y$ and $F$ learns the inverse mapping from domain $Y$ to $X$. Each of them has its own discriminator. The discriminator $D_X$ aims to discriminate between images $\{x\}$ and translated images $\{F(y)\}$ while $D_Y$ aims to discriminate between images $\{y\}$ and translated images $\{G(x)\}$
The intuition is that if an image is translated from one domain to the other and back, it should arrive at the starting image. Therefore, the mapping objectives are:
\begin{enumerate}
    \item forward mapping: $x\rightarrow G(x)\rightarrow F(G(x)) \approx x $
    \item backward mapping: $y\rightarrow F(y)\rightarrow G(F(y)) \approx y $
\end{enumerate}

For this purpose, two types of losses are used for the training of the generative networks: adversarial losses for matching the translated image to the data distribution in the target domain and cycle-consistency losses to fulfill the forward and backward mapping objectives. 

\textbf{Adversarial Loss:} The adversarial loss for the mapping function $G: X\rightarrow Y$ and the discriminator $D_Y$ can be formulated as the standard formula used for GANs:
\begin{equation}
\label{eq:adv_loss}
    L_{GAN}(G,D_{Y},X,Y) = \mathbb{E}_{y\sim p_{data}(y)}[logD_Y(y)] + \mathbb{E}_{x\sim p_{data}(x)}[log(1-D_Y(G(x)))]
\end{equation}
Where the generator $G$ tries to minimize this objective while the discriminator $D$ tries to maximize it. A similar adversarial loss $L_{GAN}(F,D_{X},Y,X)$ is introduced for the mapping function $F: Y\rightarrow X$ and the discriminator $D_X$.

\textbf{Cycle-Consistency Loss:} Using only adversarial loss can guarantee that the generative networks $G$ and $F$ produce outputs belonging to the target domain distribution $Y$ and $X$, respectively. However, it is not guaranteed that such generated images are the desired outputs. For example, the output can be a random permutation of images in the target domain, and it is not guaranteed that the output holds the same semantic structure as the input image, which is a necessity for synthetic-to-real style transfer. For this reason, it is essential to incorporate another type of loss, Cycle-consistency loss, to force the mapping functions to be cycle-consistent; that is the mapping of an image from source to target domain using $G$ and back to the source domain using $F$ should produce the same input image, for visualization refer to the images in the original paper. By using such loss, the generated output would hold the style of the target domain while being as close as possible to the original image to allow for reconstruction. This is called \textit{forward cycle-consistency}. Similarly, as in Figure~\ref{fig:cyclegan}(c), \textit{backward cycle consistency} should hold true. That is, the mapping of an image from the target domain to the source domain using $F$ and back to the target domain using $G$ should lead to the original image.



The cycle-consistency loss can be formulated as:
\begin{equation}
\label{eq:cyclecon_loss}
    L_{cyc}(G,F) = \mathbb{E}_{x\sim p_{data}(x)}[\norm {F(G(x))-x}_{1} ] + \mathbb{E}_{y\sim p_{data}(y)}[\norm {G(F(y))-y}_{1} ]
\end{equation}

\textbf{Overall Objective:} The overall loss function used for training is formulated as the following:

\begin{equation}
\label{eq:cyclegan_loss}
    L(G,F,D_{x},D_{y}) = L_{GAN}(G,D_y, X, Y) + 
    L_{GAN}(F,D_x, Y, X) + 
    \lambda L_{cyc}(G, F)
\end{equation}

where $\lambda$ is a weighting factor used to control the relative importance of the two types of losses. By the use of the above loss, the two generative mapping networks $G$ and $F$ are trained to solve: $G^*,F^* = arg_{w_G,w_F}min_{G,F} max_{D_X, D_Y}L(G,F,D_{x},D_{y})$ where $w_G$ and $w_F$ are the weights of the networks.


\subsection{Diffusion models}
Diffusion models have recently gained significant attention due to their astonishing and high-quality results in image generation~\cite{diffusion1, diffusion2}. We hereby explain the principal concept behind such models and how we adopt InST~\cite{inst} for synthetic to real style transfer. 

Diffusion models try to overcome the problem of low diversity in the generated images using GANs. 
The main idea of a diffusion model is to learn the manifold of interest by the gradual addition of noise to images in the manifold until it becomes pure Gaussian noise and, then, learn to retrieve the original image (return back to the manifold of interest) by progressively removing the noise using a deep neural network~\cite{diffusion0}. 

\begin{figure}[h!]%
    \centering

  \includegraphics[width=\linewidth]{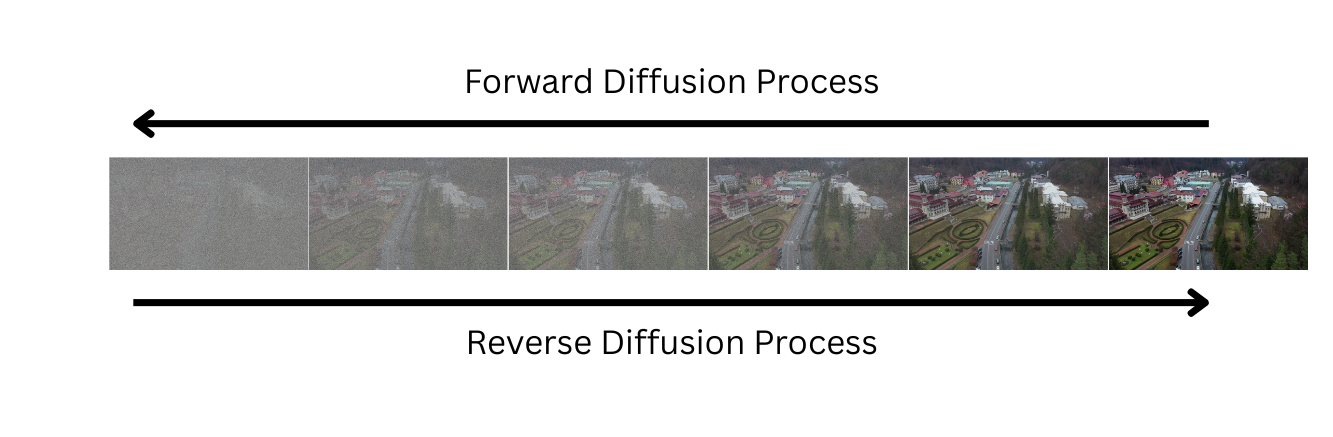}
  \caption{a simple illustration of the forward and reverse diffusion processes}
    \label{fig:diffusion}%
\end{figure}

The first step of adding Gaussian noise, also called the forward diffusion process, is a deterministic process and eventually transforms the image into pure Gaussian noise. In the second step, called the reverse sampling process, the deep model learns to remove noise gradually from a pure random noise sample to generate a new and coherent image belonging to the original data distribution. The process is illustrated in Figure~\ref{fig:diffusion}.

For formulation, the denoising model can be parameterized using a function $\epsilon_\theta(x,t)$ that tries to predict the noise component of a noisy sample $x_t$. By subtracting the predicted noise component from $x_t$ we reach $x_{t-1}$. Proceeding with the same denoising process, we can obtain gradually $x_{t-2}, x_{t-3}, ...$ until we reach the original sample $x_0$. To train the function $\epsilon_\theta(x,t)$, each sample in a minibatch is produced by selecting a random sample $x_0$, a time step $t$, and a noise $\epsilon$, and this produces a noisy sample $x_t$ such that the network tries to predict the noise component $\epsilon$ using this sample. Therefore, the training objective is \(\norm{\epsilon_\theta(x_t,t) - \epsilon}^2\). It was shown that the denoising distribution can be modelled as a Gaussian distribution $\mathcal{N}(x_{t-1}; \mu_{\theta}(x_t, t), \sum_{\theta}(x_t,t)$. It was found that fixing the variance $\sum_{\theta}(x_t,t)$ to a constant is better for sampling with fewer diffusion steps. It is suggested to parameterize the variance $\sum_{\theta}(x_t,t)$ with a neural network that is trained together with the network $\epsilon_\theta(x_t,t)$. In the end, the image generation performance (sample quality) is evaluated using the FID score~\cite{fid} metric that measures the Frechet Inception Distance between the generated sample and the real samples.

\begin{figure}[h!]%
    \centering

  \includegraphics[width=\linewidth]{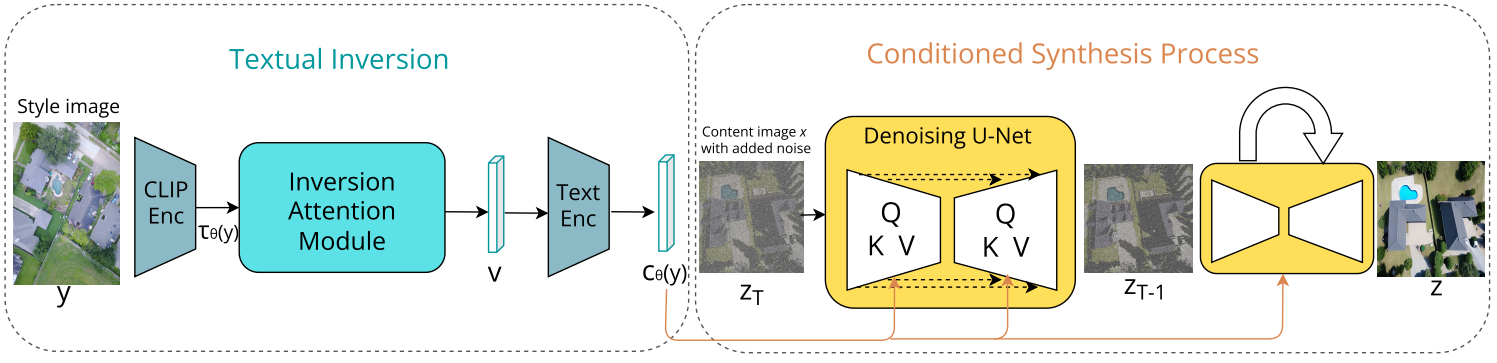}
  \caption{Overview of InST method for image style transfer using Stable Diffusion Models (SDM)~\cite{inst}. CLIP~\cite{clip} is used to give image embedding of the style image, then it is inserted as a caption conditioning in the standard form of SDM.}
    \label{fig:inst}%
\end{figure}

\textbf{InST} is a type of conditional image synthesis using Stable Diffusion Models (SDM). It is originally defined as a method to create artistic image(s) from a given example. This is done by combining the content of the input image and the style of an example image. The example styling image is converted to a short text prompt using CLIP.
Then, the diffusion-based method generates high-quality and variant artistic images conditioned on the text prompt describing the style. In Figure~\ref{fig:inst}, an overview of the method is depicted. During training, the content image $x$ is the same as the style image $y$. The image embedding of image y is obtained using CLIP image encoding. Then, it enters into an inversion attention module to convert the image embedding $\tau_\theta(y)$ into a text embedding $v$, and it is converted to the standard format $c_\theta(y)$ for caption conditioning SDM. Through a sequential denoising process, the diffusion model conditioned on the image caption $c_\theta(y)$, produces the artistically converted image $z$. During inference, the textual embedding $v$ of the style image $y$ guides the conversion process of the content image $x$.

\section{Experimental Setup}
For Cycle-GAN, the default architecture was adopted for the generative and discriminator networks\footnote{Github repository: \url{https://github.com/tensorflow/docs/blob/master/site/en/tutorials/generative/cyclegan.ipynb}}. The generative network contains 3 convolutions, 9 residual blocks, two fractionally-strided convolutions with stride $\frac{1}{2}$, and one final convolution to map features to the output RGB image. For the discriminators, the patch-level architecture was adopted~\cite{patchgan}.

For training, the weighting factor $\lambda$ is set to 10. Adam optimizer is used with a batch size of $1$. The learning rate is set to $2\times10^{-4}$ for generative and discriminator networks, and they are trained from scratch. The model is trained for a maximum of 50 epochs, and the best checkpoint is used for evaluation. The resolution of input and predicted images is $256\times256$. The implementation is done on TensorFlow. Two experiments were run: the first to train the model on the conversion from MidAir+TopAir to real images, and the second to train it on the conversion from SkyScenes+SynDrone to real images. The group of real images used during training contains 209 images randomly selected from the datasets UDD, FSI, AeroScapes, WildUAV, ICG, and RuralScapes. The first group of synthetic images contains 173 images randomly selected from TopAir, while the second group contains 145 images randomly selected from SkyScenes.

For InST, we use the default implementation done by the authors\footnote{Github repository: \url{https://github.com/zyxElsa/InST/tree/main}}. We have finetuned the model on the same group of 209 real images used above. The configuration file for finetuning is kept as the default. For testing, the checkpoint of the latent diffusion model is used along with the embeddings learned by the finetuned model. Here, during inference, a styling image has to be provided to the network to convert the input image to the target style. For converting synthetic images of the \textit{Neighbourhood} environment of \textit{TopAir}, the styling image is set as a random image from the real dataset \textit{FSI}. For converting the synthetic images of \textit{SkyScenes} and \textit{SynDrone}, the styling image is set as a random image from the real dataset \textit{VisDrone}~\cite{visdrone}.

\section{Preliminary Results}
We conduct some preliminary experiments in image style transfer to evaluate the quality and realism of the style transfer of our synthetic data used for training into the real style. This is currently an active area of research in the automotive field~\cite{atapour2018real, xiao2022transfer, zheng2018t2net, chen2019learning}, but we seek to apply it to our synthetic aerial data to see how realistic the output is. In the case of high realism of the output, the converted images used for training can help in closing the gap of synthetic-to-real generalization of the network. This can be a line of research for future improvements, God willing, on the synthetic-to-real performance of the network.

\textbf{Cycle-GAN:}

\begin{figure}[h!]%
    \centering
    \includegraphics[width=0.8\linewidth]{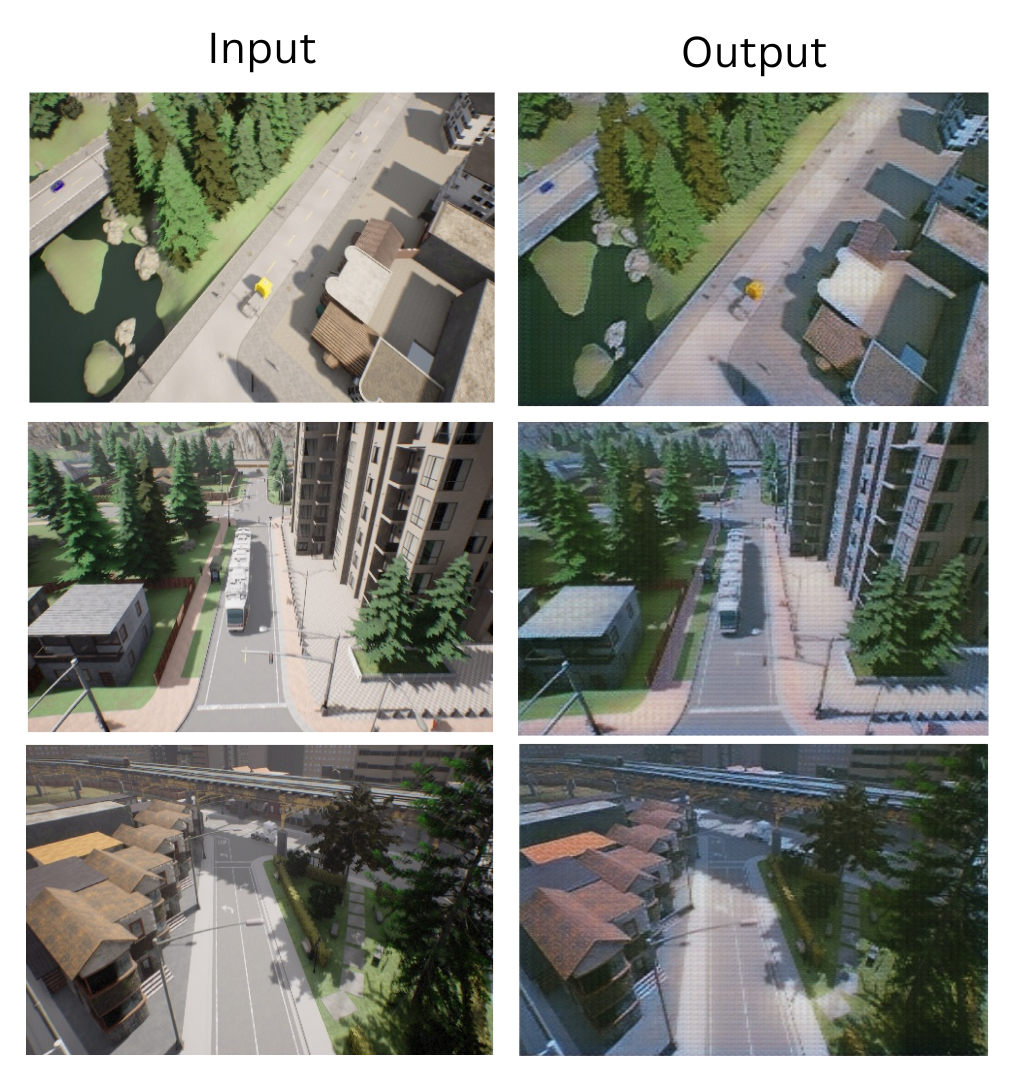}
    \caption{\centering Sample output of style transfer of images from SynDrone using Cycle-GAN}%
    \label{fig:style_tr1}%
\end{figure}

In Figure~\ref{fig:style_tr1}, sample style-converted images from SkyScenes and SynDrone are shown. It can be noticed that Cycle-GAN focuses on changing the colors in the input images to convert them to the realistic style rather than changing the appearance and content of the images. The converted output contains noise, and the quality is not satisfactory. In Figure~\ref{fig:style_tr2}, sample style-converted images from TopAir are shown. The same analysis done on SkyScenes and SynDrone holds here as well; the network focused only on changing colors, hue, jittering, brightness, and illumination conditions without content modification. The main drawbacks are the noise appearing in the output and the colors of the objects are diffused outside their boundaries.

\begin{figure}[h!]%
    \centering
    \includegraphics[width=0.8\linewidth]{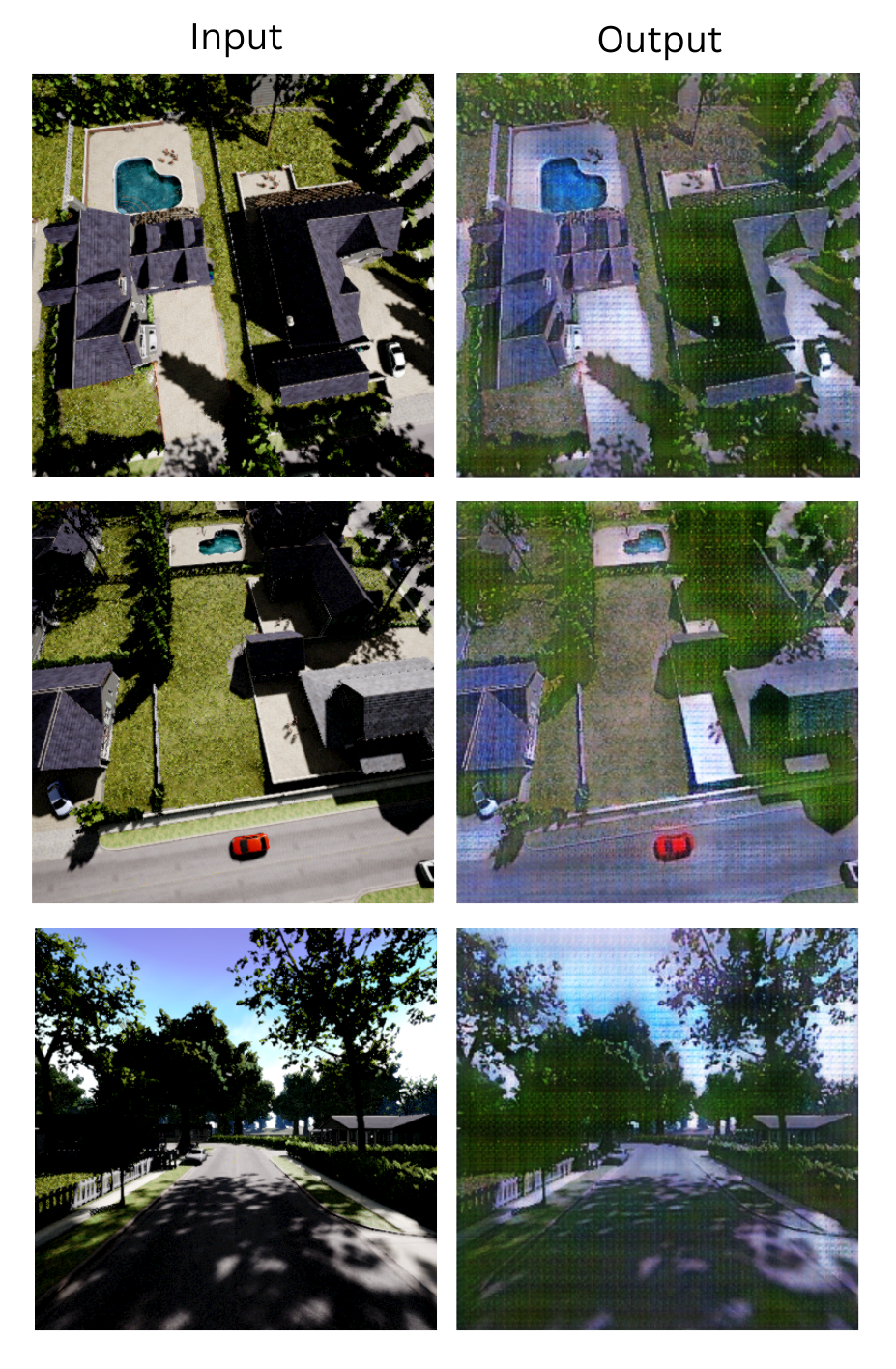}
    \caption{\centering Sample output of style transfer of images from TopAir using Cycle-GAN}%
    \label{fig:style_tr2}%
\end{figure}


\textbf{Diffusion Model:}

\begin{figure}[h!]%
    \centering
    \includegraphics[width=\linewidth]{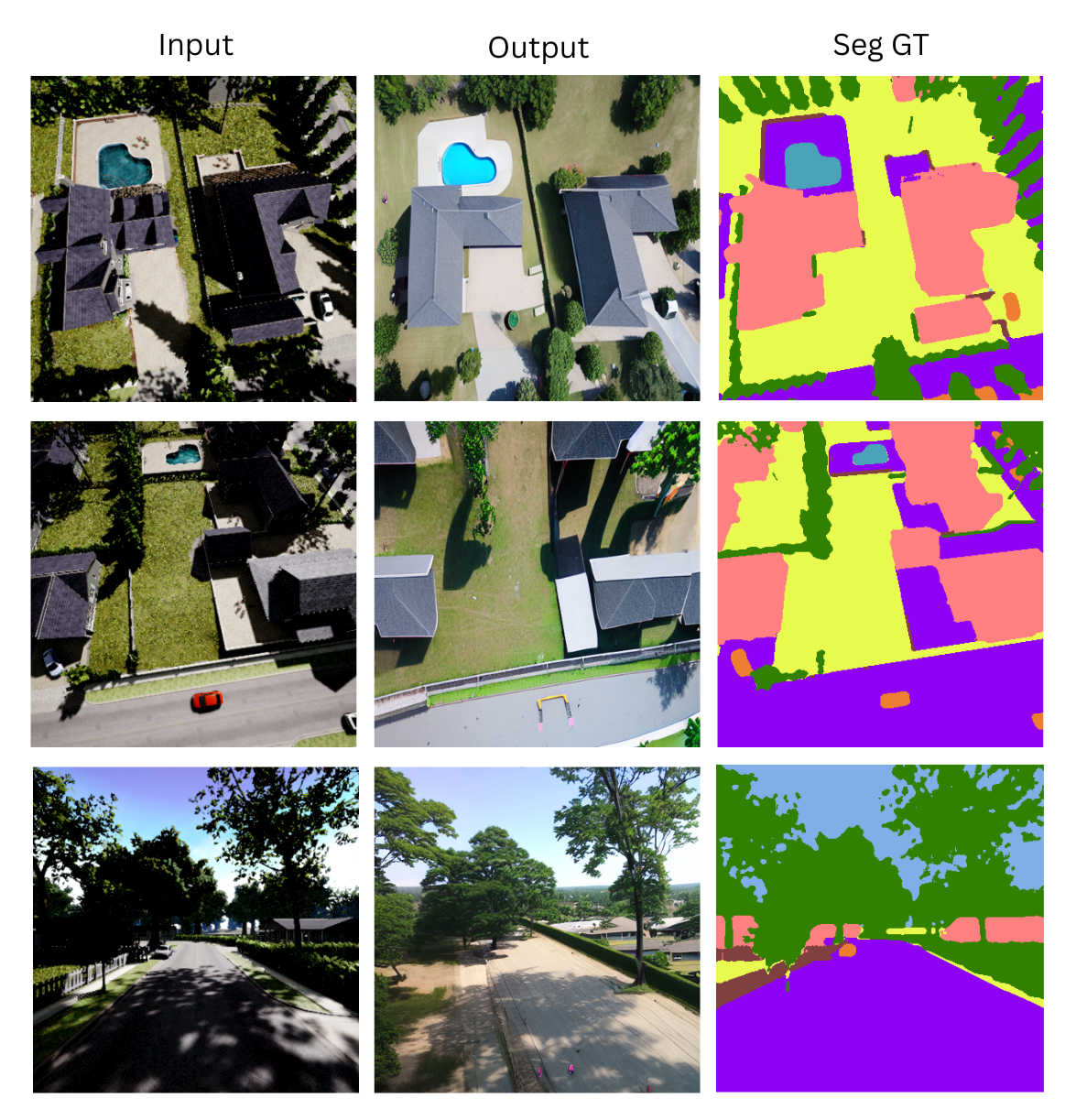}
    \caption{\centering Sample output of style transfer of images from TopAir using Diffusion Model InST. It can be noted that the output images do not hold exactly the same semantic segmentation layout.}%
    \label{fig:style_tr3}%
\end{figure}

In Figure~\ref{fig:style_tr2}, sample style-converted images from SynDrone are shown, and in Figure~\ref{fig:style_tr3}, sample style-converted images from TopAir are shown. The same input images used in Cycle-GAN are used here for testing the diffusion model. It can be noted that the network here focused on changing the content in the input images to reflect realistic elements instead of the synthetic ones. See, for example, the bus, buildings, roads, and the trees in Figure~\ref{fig:style_tr2}. However, while the output looks very realistic, it does not hold exactly the same semantic segmentation layout as the input, see the third column in Figures~\ref{fig:style_tr2} and ~\ref{fig:style_tr3}. In the second row of Figure~\ref{fig:style_tr3}, the cars and the swimming pool are removed in the output, and the position and size of the houses have been changed. In the third row, the position of the trees is changed. For this reason, the converted images cannot be used for supervised training of neural networks with the same ground truth depth and semantic segmentation maps belonging to their synthetic pairs because they are no longer valid. However, the obtained results open the door for further research in that area.

\begin{figure}[h!]%
    \centering
    \includegraphics[width=\linewidth]{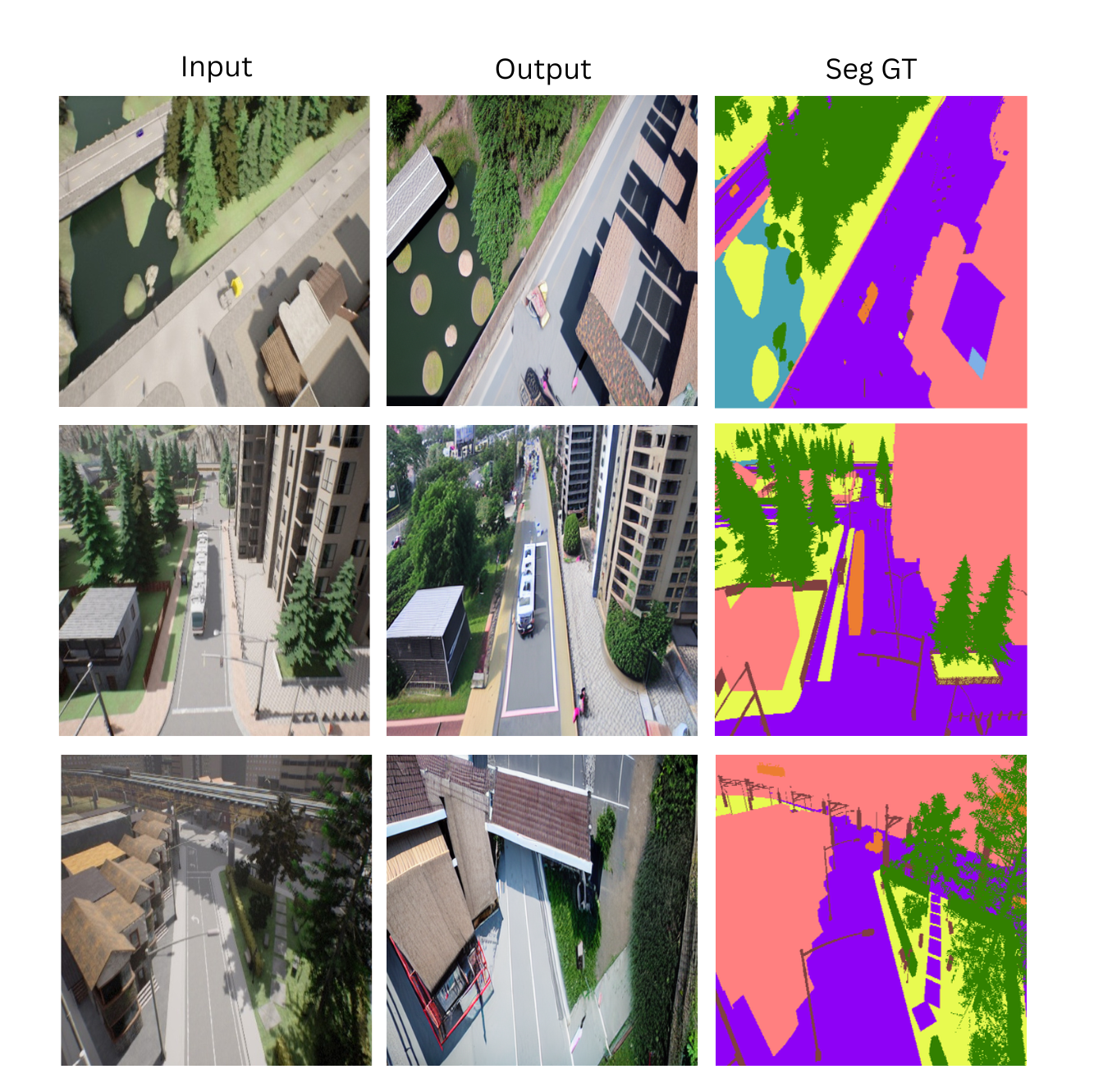}
    \caption{\centering Sample output of style transfer of images from SynDrone using Diffusion Model InST. It can be noted that the output images do not hold exactly the same semantic segmentation layout.}%
    \label{fig:style_tr2}%
\end{figure}

\chapter{Application in the Marine Domain}
\label{chap:marine}
\ifpdf
    \graphicspath{{Chapter6/Figures/PNG/}{Chapter6/Figures/PDF/}{Chapter6/Figures/}}
\else
    \graphicspath{{Chapter6/Figures/EPS/}{Chapter6/Figures/}}
\fi


\section*{Summary}

In this chapter, the methodology adopted to apply in the maritime domain is described, and the collected synthetic marine data \textit{MidSea} is presented. Then, the setup used in conducting the experiments in the maritime domain is revealed, and the experiments done on the marine datasets are analyzed. 

MidSea is a dataset (with $\sim100k$ frames) collected using UnrealEngine in a synthetic marine environment, and it is annotated with depth, segmentation maps, and camera pose information. Experiments in the maritime domain were twofold: one for validating Co-SemDepth on synthetic data (MidSea), and the other for assessing the synthetic-to-real generalization of Co-SemDepth using different techniques, including self-supervised approaches, to enhance the generalization. 
The results obtained indicate that Co-SemDepth, trained from scratch on MidSea, can perform well on both the MidSea test data and in semantic segmentation on real data from the SMD dataset. However, it did not perform well on the real MIT dataset, even after applying various techniques. 

The work in this chapter was conducted in the master's thesis of Gabriele Dellepere under the supervision of the authors of this thesis. 

\section{MidSea Data Acquisition}
\begin{figure}[h!]%
    \centering
    \includegraphics[width=\linewidth, height = 3cm]{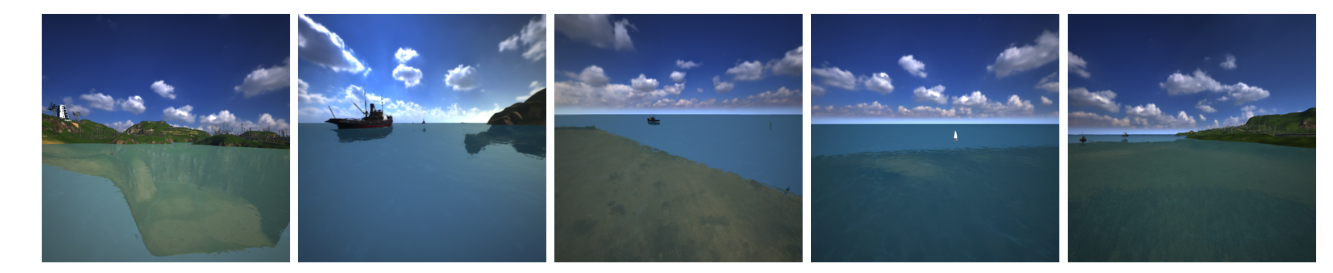}
    \caption{\centering Sample images from the collected marine data in the Tropical Island environment}%
    \label{fig:sample_marine_collc}%
\end{figure}

UnrealEngine5 was exploited to collect annotated marine data in the synthetic Tropical Island environment\footnote{\href{https://www.fab.com/listings/a8804eea-f081-4aab-97d6-bf8e031c0e46}{Available on this link} }, see Figure~\ref{fig:sample_marine_collc} for a sample of the collected data. Such a dataset was named \textit{MidSea} to reflect its content that was captured in the middle of the sea. It is annotated with depth and semantic segmentation maps as well as camera transformation data. Ships, boats, and buoys were inserted manually in the sea of the environment, and the semantic segmentation classes of objects were limited to 4 classes: Sky, Water, Ship, and Land. In this case, the class \textit{Ship} contained ships, boats, and buoys.
For depth, the sea surface in all images appeared as a linear gradient in the 2D depth maps.
The data comprises a total of 8 trajectories, containing in total around 98K frames. A train-validation split of $80\%-20\%$ was used.

\section{Coping with few annotated data}
For assessing the monocular depth estimation and semantic segmentation in marine scenes, Co-SemDepth was employed for training and evaluation. In addition to the usual supervised training of the network, self-supervised methods and training modalities were adopted on the architecture of Co-SemDepth for assessment and to help the sim-to-real transition of the network. Namely, MoCo~\cite{moco} and DINO~\cite{dino} self-supervised approaches were used. MoCo is a method for training the encoder in a self-supervised way. It consists of using two instances of the same encoder (+ a projection head for each of them) for extracting feature maps from sufficiently large unannotated datasets of images. The objective of training the first encoder, through gradient descent and backpropagation, is that the predicted features of an image should be as close as possible to the predicted features of an augmented version of the same image using the second encoder. At the same time, the predicted features of the first encoder should be as far as possible from all other images in the dataset (named "negative examples"). The second encoder is not trained using gradient descent; however, its weights are updated with an exponential moving average of the weights of the first encoder. By using MoCo, the encoder of any neural network can be trained in a self-supervised way by making it invariant to image augmentations and without the need for data labeling. In this work, MoCo is employed for the training of the encoder of Co-SemDepth, and the network performance is compared with the other modalities of training. 

DINO is another self-supervised approach for pre-training the encoder of a deep network to extract features from the image regardless of color, illumination, and rotation variations. Similar to MoCo, two versions of the same encoder are used: a student and a teacher. The predicted feature maps of the same image with and without augmentations are compared to each other. The teacher is updated using an exponential moving average of the student weights. The main differences from MoCo lie in the type of augmentations done on the images and in the mechanisms adopted to avoid collapse to constant or meaningless outputs. The augmentations done on the image are both global and local. The teacher network only sees the global augmentations, while the student network sees all types of augmentations, thus forcing the student encoder to match local features and global features and become robust against different scales. 

Avoiding collapse means avoiding the encoder student from predicting always a constant map (or maps where always one dimension dominates over the other) to match the teacher output. Such a prediction is called a trivial solution. DINO uses two operations to solve this issue: centering and sharpening of the output. In centering, the moving average of the teacher's logits is subtracted from the output before applying softmax. This step forces balance in the teacher's output. Sharpening means using a small temperature for the Softmax's output of the teacher so that its class prediction becomes more confident in one class. This pushes toward non-uniform outputs. Combining the two operations pushes the network to use all output dimensions and ensures high variability in the output feature predictions.

\section{Experimental Setup}
 
All the default settings previously described for the validation of Co-SemDepth are kept in the marine experiments\footnote{The experiments are conducted using a PyTorch re-implementation of Co-SemDepth. The code can be found here: \url{https://github.com/GabrieleDellepere/thesis}.}.
The number of epochs in each experiment depended on the convergence speed of the plotted metrics on the validation data to avoid overfitting. 

\begin{figure}[h!]%
    \centering
    
\begin{subfigure}[b]{\textwidth}
  \centering
  \includegraphics[width=0.5\linewidth]{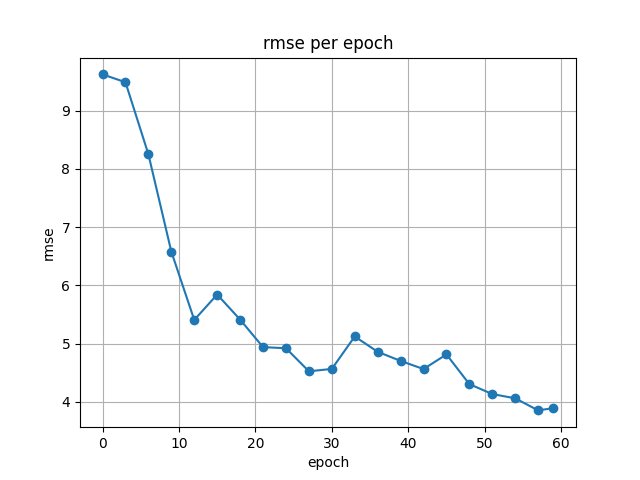}
  \caption{Validation RMSE per epoch }
  \label{fig:val_rmse}
\end{subfigure}
\begin{subfigure}[b]{\textwidth}
  \centering
  \includegraphics[width=0.5\linewidth]{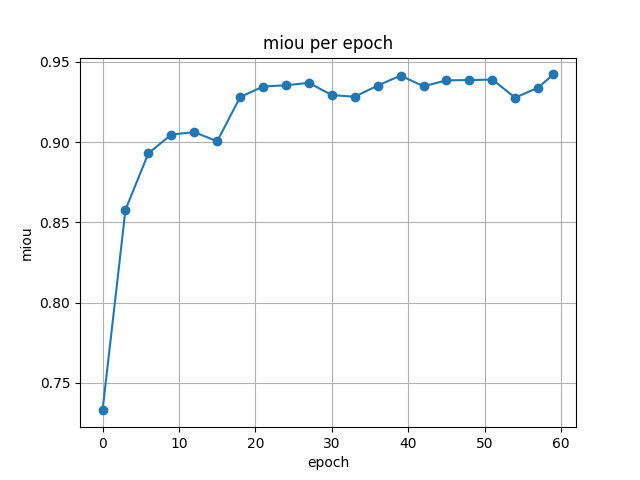}
  \caption{Validation mIoU per epoch}
  \label{fig:val_miou}
\end{subfigure}

    \caption{\centering Plots of the metrics evaluated on the validation data of MidSea while training Co-SemDepth}%
    \label{fig:plots_marine}%
\end{figure}

The dataset used for training is the collected synthetic \textbf{MidSea} dataset. Image resolution of $384\times384$ is used. 
The validation plots of $mIoU$ and $RMSE$ during training of Co-SemDepth can be viewed in Figures~\ref{fig:val_miou} and ~\ref{fig:val_rmse}.

For testing, two real datasets are used for only a qualitative evaluation due to the lack of depth and segmentation annotations:
\begin{itemize}
    \item \textbf{SMD}: contains clear scenes without water reflections or bad weather. However, the IMU measurements are not available. Hence, only semantic segmentation evaluation can be done on this dataset because the parallax decoder branch requires camera transformation data.
    \item \textbf{MIT}: it contains a variety of real-life data annotated with multiple sensor measurements, including the IMU. Thus, this dataset can be used for predicting both depth and semantic segmentation maps using the Co-SemDepth architecture.

\end{itemize}

The workstation used has 16GB RAM, an Intel Core i7 processor, and a single NVIDIA Quadro P5000 GPU card running CUDA11.4 with CuDNN 7.6.5 and Ubuntu OS.

\section{Results}
\subsection{Co-SemDepth Validation on Synthetic Data}
\begin{figure}[h]
  \centering
  \includegraphics[width=0.30\textwidth]{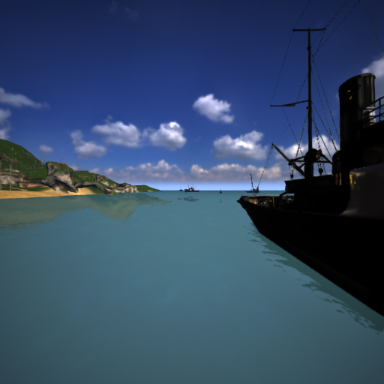}
  \includegraphics[width=0.30\textwidth]{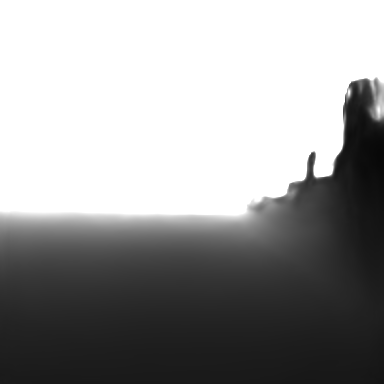}
  \includegraphics[width=0.30\textwidth]{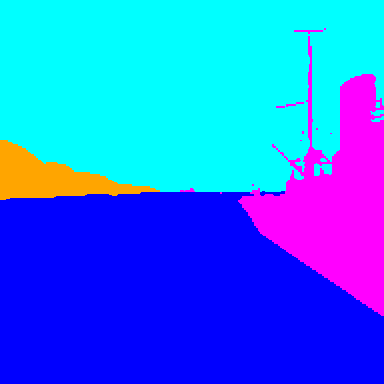}
  \caption{Sample prediction of depth and segmentation maps on MidSea. Depth is shown in a linear scale, and distance values greater than 255 meters are clamped to 255; that's why the far island disappears in the depth map. Semantic segmentation color index is: Blue$=$Sea, Cyan$=$Sky, Yellow$=$Ground/Buildings/Vegetation, Magenta$=$Other} 
  \label{fig:unreal_comp_1}
\end{figure}

For testing the application of Co-SemDepth in the marine domain, we train and test it on the collected MidSea synthetic marine data. It can be seen from the plots in Figure~\ref{fig:plots_marine} that the network shows promising results when trained on MidSea and tested on the validation split of the same dataset. Both depth and segmentation metrics continuously improve over the training epochs. In Figure ~\ref{fig:unreal_comp_1}, sample predictions on the validation data can be visualized.
It can be concluded that the network did not overfit the training data; however, it can be the case that there is little diversity in the dataset itself. For testing the network on other datasets, we choose the weights of the last epoch, as it gives the best validation performance. In Table~\ref{tab:marine_synth_methods}, the quantitative results on the validation set of MidSea can be found. 

\begin{table}[ht]
\centering
\resizebox{\textwidth}{!}{%
\begin{tabular}{ l|c|c|c|c|c|c }
\hline
Method & \multicolumn{1}{c|}{Semantic Metrics}
& \multicolumn{5}{c}{Depth Metrics} \\
\cline{2-7}
 & mIoU $\uparrow$ & RMSE $\downarrow$ & AbsRelErr $\downarrow$ & $\delta<1.25$ $\uparrow$ & $\delta<1.25^{2}$ $\uparrow$ & $\delta<1.25^{3}$ $\uparrow$ \\
\hline
Base CoSemDepth & 83.38\% & 4.24 & 0.092 & 81.7\% & 95.5\% & 98.5\% \\
\hline
CoSemDepth + custom augm. & 91.21\% & 3.59 & 0.081 & 88.0\% & 97.3\% & 98.9\% \\
\hline
Finetuned MoCo & 75.0\% & 9.03 & 0.265 & 67.3\% & 79.1\% & 87.4\% \\
\hline
ViT+DINO encoder & 79.11\% & 9.42 & 0.221 & 65.0\% & 75.2\% & 83.4\% \\
\hline
\end{tabular}
}
\caption{Comparison of different methods on the collected Synthetic dataset MidSea. Base CosemDepth includes the default augmentations, while the "custom augm." version includes more aggressive color jittering and z-axis rotation.}
\label{tab:marine_synth_methods}
\end{table}

\subsection{Synthetic-to-Real Performance}
\subsubsection{Results on SMD:} 
\begin{figure}[h!]%
    \centering
    
\begin{subfigure}[b]{\textwidth}
  \centering
  \includegraphics[width=0.45\textwidth]{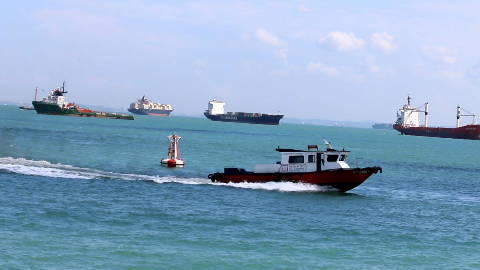}
  \includegraphics[width=0.45\textwidth]{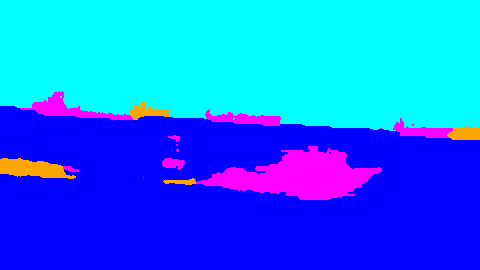}
  \caption{sample prediction 1}
  \label{fig:smd_norm1}
\end{subfigure}
\begin{subfigure}[b]{\textwidth}
  \centering
  \includegraphics[width=0.45\textwidth]{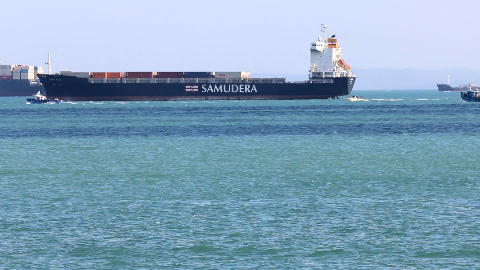}
  \includegraphics[width=0.45\textwidth]{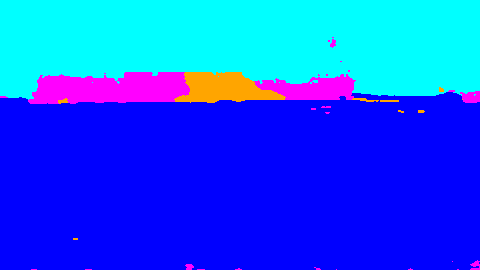}
  \caption{sample prediction 2}
  \label{fig:smd_norm2}
\end{subfigure}
    \caption{\centering Sample semantic predictions on SMD using Co-SemDepth trained on the synthetic MidSea. Semantic segmentation color index is: Blue$=$Sea, Cyan$=$Sky, Yellow$=$Ground/Buildings/Vegetation, Magenta$=$Other}
  \label{fig:smd_comp_2}
\end{figure}

\begin{figure}[h!]%
    \centering
    
\begin{subfigure}[b]{\textwidth}
  \centering
  \includegraphics[width=0.45\textwidth]{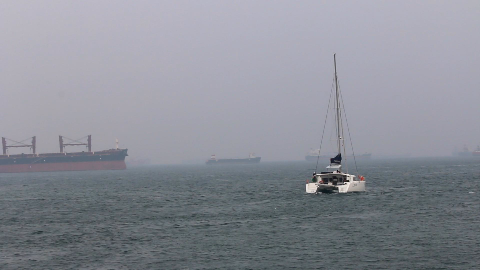}
  \includegraphics[width=0.45\textwidth]{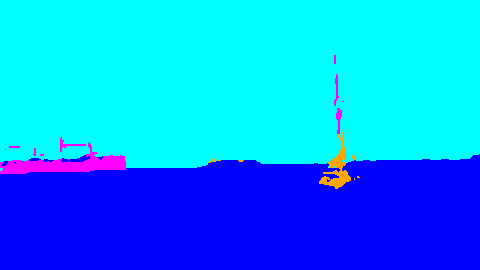}
  \caption{sample prediction 1}
  \label{fig:smd_fog1}
\end{subfigure}
\begin{subfigure}[b]{\textwidth}
  \centering
  \includegraphics[width=0.45\textwidth]{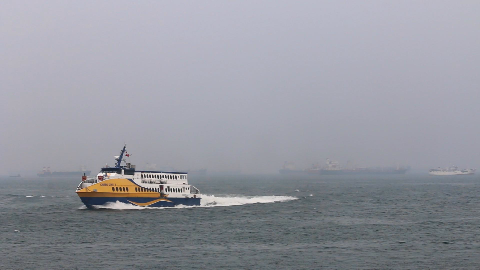}
  \includegraphics[width=0.45\textwidth]{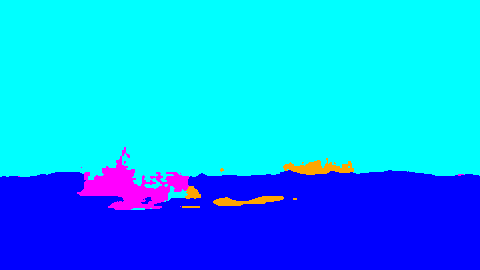}
  \caption{sample prediction 2}
  \label{fig:smd_fog2}
\end{subfigure}
    \caption{\centering Sample semantic predictions on foggy images from SMD using Co-SemDepth trained on the synthetic MidSea. Semantic segmentation color index is: Blue$=$Sea, Cyan$=$Sky, Yellow$=$Ground/Buildings/Vegetation, Magenta$=$Other}
  \label{fig:smd_comp_fog}
\end{figure}

The synthetic-to-real zero-shot performance of Co-SemDepth is evaluated by training Co-SemDepth on MidSea and testing it on images from SMD real data. Unfortunately, there are no semantic or depth annotations provided in SMD, and for this reason, the evaluation is done only qualitatively. In addition, the lack of camera pose data in SMD forces us to predict only semantic segmentation. In general, the results turn out positive. Sample predictions can be found in Figures~\ref{fig:smd_comp_2} and~\ref{fig:smd_comp_fog}. It can be seen that the network is able to identify the main elements of the scene (i.e.\ water, sky, and ships) even in hard weather conditions like fog. However, the network can sometimes be confused between ships and land, especially when the ships are far away.

\subsubsection{Results on MaSTr:} 
In addition, we run tests on the real dataset MaSTr~\cite{mastr}. In this case, we can conduct a quantitative evaluation due to the availability of semantic segmentation annotation in the dataset. The network Co-SemDepth trained on MidSea could achieve $45.84\% mIoU$ in the zero-shot testing and $49.41\% mIoU$ after fine-tuning on part of MaSTr for 20 epochs. Hence, there is an enhancement; however, it is not huge. 



    

\subsubsection{Results on MIT}
\begin{figure}[h!]%
    \centering
    
\begin{subfigure}[b]{\textwidth}
  \centering
  \includegraphics[width=0.3\textwidth]{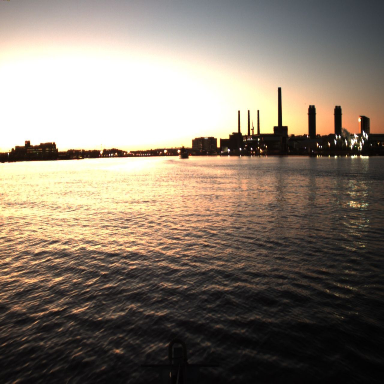}
  \includegraphics[width=0.3\textwidth]{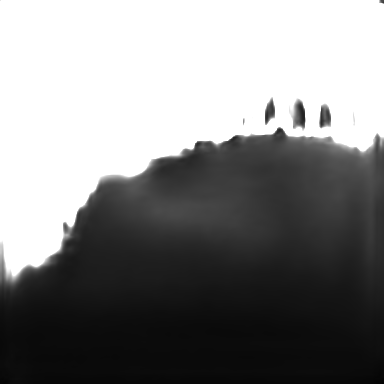}
  \includegraphics[width=0.3\textwidth]{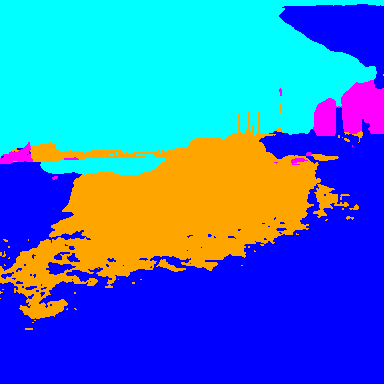}
  \caption{sample prediction 1}
  \label{fig:mit1}
\end{subfigure}
\begin{subfigure}[b]{\textwidth}
  \centering
  \includegraphics[width=0.3\textwidth]{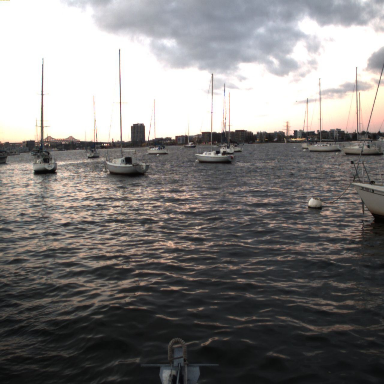}
  \includegraphics[width=0.3\textwidth]{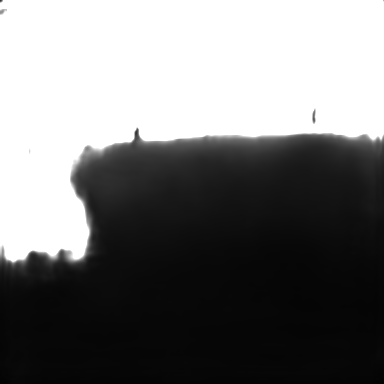}
  \includegraphics[width=0.3\textwidth]{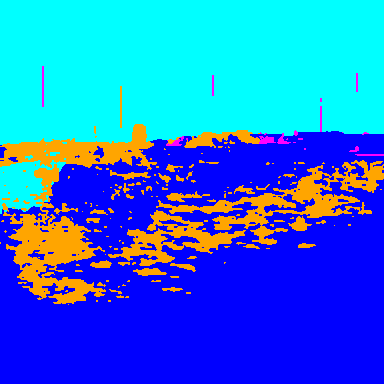}
  \caption{sample prediction 2}
  \label{fig:mit2}
\end{subfigure}
\begin{subfigure}[b]{\textwidth}
  \centering
  \includegraphics[width=0.3\textwidth]{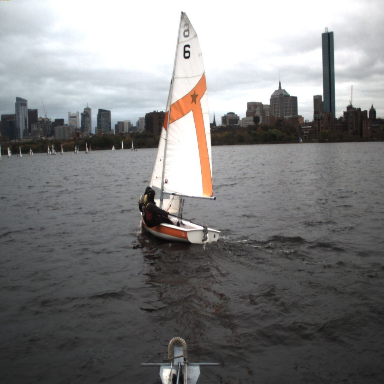}
  \includegraphics[width=0.3\textwidth]{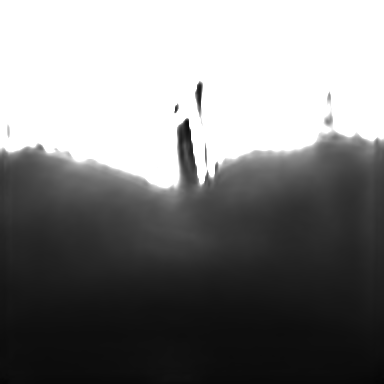}
  \includegraphics[width=0.3\textwidth]{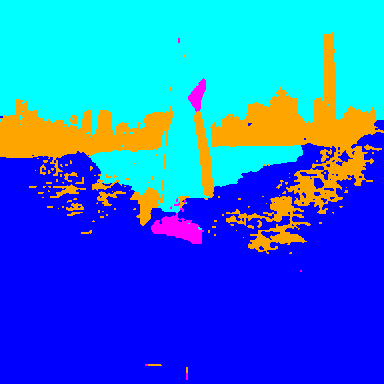}
  \caption{sample prediction 3}
  \label{fig:mit3}
\end{subfigure}
    \caption{\centering Sample zero-shot predictions on images from MIT dataset using Co-SemDepth trained on the synthetic MidSea. Depth is displayed in a linear scale and clamped at 255 meters. Semantic segmentation color index is: Blue$=$Sea, Cyan$=$Sky, Yellow$=$Ground/Buildings/Vegetation, Magenta$=$Other
}
  \label{fig:mit_cosemdepth}
\end{figure}

The MIT Sea Grant Perception dataset is considered the most diverse dataset containing marine scenarios and weather conditions. For this reason, it is selected as our high target. 
\noindent\textbf{Zero-shot prediction:} It can be seen in Figure~\ref{fig:mit_cosemdepth} sample predictions on images from MIT data using Co-SemDepth trained on MidSea. Unfortunately, the network performance here is not excellent. For semantic segmentation, it can be noted that the network can well distinguish between sky and land, especially in good weather conditions. However, when it comes to distinguishing the line between land and the sea, it struggles. 
Besides, it can be observed that Co-SemDepth is not able to properly identify the ships in the input image, and this is a problem since most of the applications in the marine domain, like exploration or obstacle avoidance, require the detection of the surrounding vehicles. For depth estimation, the situation is not good enough. The network predicts all the sea and land areas as a linear gradient, and it is not able to well detect the horizon line. It can also be noted from Figures~\ref{fig:mit1} and~\ref{fig:mit2} that the sunlight has an effect on the predicted depth maps. The areas of the sea illuminated with high sunlight are predicted with high depth values (greater than 255 meters).

\begin{figure}[h]
  \centering
  \includegraphics[width=0.4\textwidth]{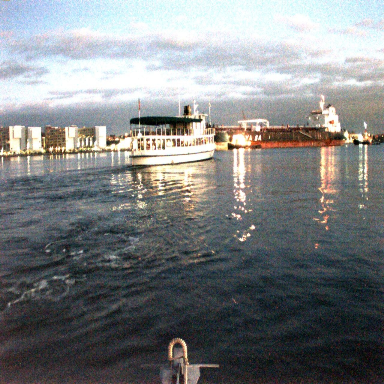}
  \includegraphics[width=0.4\textwidth]{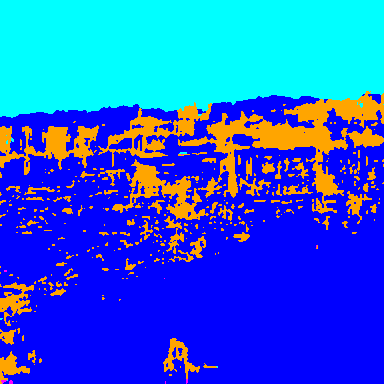}
  \caption{Sample prediction on MIT dataset after finetuning Co-SemDepth on MaSTr dataset. Color index is: Blue$=$Sea, Cyan$=$Sky, Yellow$=$Ground/Buildings/Vegetation, Magenta$=$Other}
  \label{fig:mit_with_mastr}
\end{figure}

\noindent\textbf{Finetuning on MaSTr:} To overcome the above challenges, Co-SemDepth, which is fine-tuned on the MaSTr dataset, is tried to predict semantic segmentation on the MIT dataset. In this case, no depth predictions can be performed because the MaSTr dataset does not contain camera pose data. The output results can be found in Figure~\ref{fig:mit_with_mastr}. From the figure, the network is more able to detect the sea area. However, it is prone to detecting land areas as sea. For this reason, the results of this solution are also not satisfying, and the problem of undetected vehicles persists.

\begin{figure}[h]
  \centering
\includegraphics[width=0.3\textwidth]{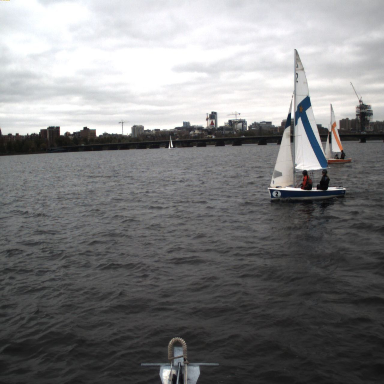}
  \includegraphics[width=0.3\textwidth]{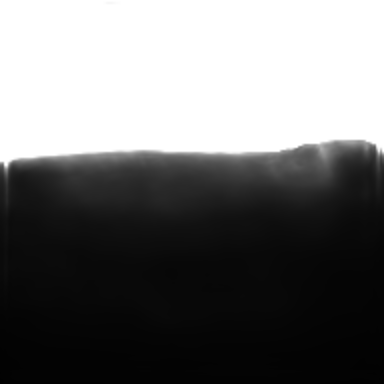}
  \includegraphics[width=0.3\textwidth]{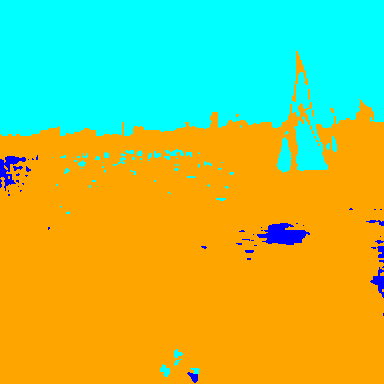}
  \caption{Sample predictions on MIT dataset using Co-SemDepth after adding custom augmentations. Depth is viewed in metric scale. Segmentation color index is: Blue$=$Sea, Cyan$=$Sky, Yellow$=$Ground/Buildings/Vegetation, Magenta$=$Other}
  \label{fig:mit_with_augm}
\end{figure}

\noindent\textbf{Augmentations:} More augmentations are added to the training of Co-SemDepth in the hope of making the images of MidSea more diverse and, thus, making the network generalize better to the unseen data of MIT. In particular, the range of color jittering is increased, and more cropping and z-axis rotations are added. In fact, the metrics obtained on the validation set of MidSea are promising compared to the values obtained with the default augmentations. The results can be found in Table~\ref{tab:marine_synth_methods}. Unexpectedly, the results on MIT got worse, see Figure~\ref{fig:mit_with_augm}. The network is completely confused between land and sea; this can be because increasing the color jittering so much makes the color of the sea and land in the training data change significantly, and that leads the network to not be able to distinguish them by color anymore. The network is still unable to detect the ships in the scene well. For depth estimation, the network is able to well detect the horizon line and this is an improvement compared to the model with default augmentations. However, it still cannot distinguish between the distances of the sea surface and vehicles in it. Such augmentations are not used in the following experiments due to the bad segmentation performance.

\begin{figure}[h]
  \centering
  \includegraphics[width=0.3\textwidth]{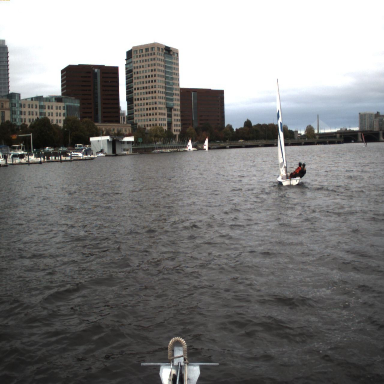}
  \includegraphics[width=0.3\textwidth]{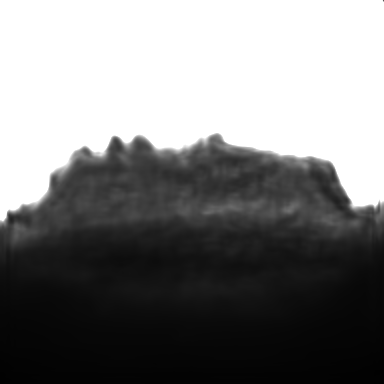}
  \includegraphics[width=0.3\textwidth]{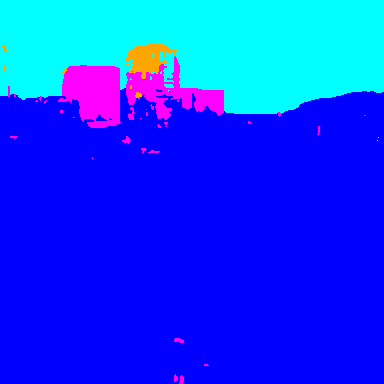}
  \includegraphics[width=0.3\textwidth]{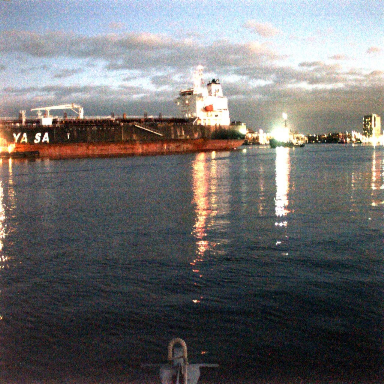}
  \includegraphics[width=0.3\textwidth]{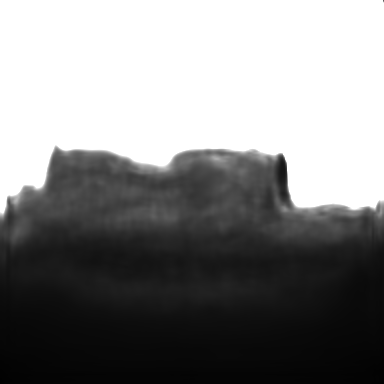}
  \includegraphics[width=0.3\textwidth]{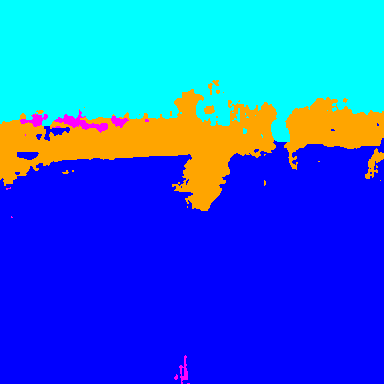}
  \caption{Sample predictions on the MIT dataset using Co-SemDepth after finetuning the weights obtained from MoCo pre-training.
  Depth is viewed in metric scale. Segmentation color index is: Blue$=$Sea, Cyan$=$Sky, Yellow$=$Ground/Buildings/Vegetation, Magenta$=$Other}
  \label{fig:mit_with_moco}
\end{figure}

\noindent\textbf{MoCo pretraining:} The MoCo self-supervised pretraining method is adopted for the training of the encoder of Co-SemDepth on both MidSea and MIT datasets. The MLP used after the encoder to aggregate the features during pretraining is composed of two layers with 2048 and 128 perceptrons with Leaky ReLU activation. The encoder is pretrained for 80 epochs. After that, the whole architecture (pre-trained encoder + decoder) is trained on MidSea with a learning rate one order of magnitude less than the one used for pre-training. The network is trained for 50 epochs. The evaluation results on MidSea validation data can be found in Table~\ref{tab:marine_synth_methods}. It can be seen that using the MoCo pretraining method makes the model perform well on MidSea (although the results are inferior to the base network). However, by looking at the output maps on images from the MIT dataset, Figure~\ref{fig:mit_with_moco}, it can be concluded that the results using the MoCo pretraining method are not satisfactory. A probable reason for that can be that MoCo needs a huge number of good negative examples to work well, and in our case, there is little diversity in the data we have. So, the difference between positive examples and negative examples is minimal, not sufficient for a proper training of the encoder. 

 \noindent\textbf{DINO pretraining:}
In this experiment, the DINO self-supervised method is adopted for evaluation. However, in this case, we load the encoder weights available on the DINOv3 HuggingFace repository\footnote{\url{https://huggingface.co/docs/transformers/main/model_doc/dinov3}}. We discard the encoder of Co-SemDepth and use instead the pretrained ViT encoder of DINO with the decoder of Co-SemDepth. The network is fine-tuned after that on MidSea. From Table~\ref{tab:marine_synth_methods}, it can be noted that the results achieved with this approach are better than MoCo. However, they are still inferior to the ones achieved by Base Co-SemDepth. 
 
\chapter{Conclusions}
\label{chap:conclusion}

\ifpdf
    \graphicspath{{Chapter7/Figures/Raster/}{Chapter7/Figures/PDF/}{Chapter7/Figures/}}
\else
    \graphicspath{{Chapter7/Figures/Vector/}{Chapter7/Figures/}}
\fi


In this thesis, we presented Co-SemDepth, a fast joint architecture for depth estimation and semantic segmentation on UAV images. 
Our architecture proved to be lightweight (containing only 5.2 M parameters) and fast (with a frame rate of 20 FPS) compared to the other state-of-the-art methods while achieving better or on-par accuracies ($RMSE$ of $6.7$ and $mIoU$ of $75.4\%$ on MidAir). This makes it suitable to be deployed on UAV hardware for conducting real-time scene analysis. 
By using our light architecture, it is possible for a UAV to analyze its surrounding scenes autonomously and independently using its onboard microcontroller or single board computer that are limited in memory; our architecture requires only $6.2 GB of GPU memory$. In this case, there will be no need for sending data to a remote server to carry out the analysis. The output depth and semantic maps can provide geometric and semantic understanding of the scene that is necessary to carry out a variety of UAV missions (including 3D SLAM).


Moreover, to help in filling the gap of the scarcity of available annotated datasets in the aerial field, the \textit{TopAir} dataset was introduced. It is a synthetic dataset collected using the AirSim simulator in a variety of UnrealEngine environments at different altitudes and lighting conditions. The dataset contains annotations of depth and segmentation maps as well as camera translation and orientation data for every frame. 

In addition, the synthetic-to-real domain shift problem was analyzed. Our architecture (light U-Net-based) and TaskPrompter (large Transformer-based architecture) were used to conduct an extensive analytical study in synthetic-to-real domain generalization of monocular depth estimation and semantic segmentation for aerial robots. The networks were trained on synthetic datasets and evaluated on real data. The study reveals that Co-SemDepth is better in the generalization of depth estimation due to the parallax estimation modalities implemented in the architecture. On the other hand, the results reveal that TaskPrompter is better in semantic segmentation, as it seems that for this task, it is more beneficial to have a complex, large network like TaskPrompter that can capture the semantic potential of the data, especially for segmenting small objects. In general, it was observed that the similarity between the environments (rural or urban) appearing in the synthetic data used for training and the real data used for testing greatly affects the domain shift network performance. Specifically, using the synthetic MidAir and TopAir datasets for training the models led to better generalization to the real data that were captured in rural areas, while using the synthetic SkyScenes and SynDrone datasets was preferable to make the model generalize to real datasets of urban nature. 

Additional experiments were conducted to test the effect of adding a small percentage of real data to the synthetic data during training (few-shot learning). The obtained results were positive on some datasets and disappointing on others. As a whole, there is a promising line for further research in that area. 3D semantic maps of the scenes 
were constructed using the predicted depth and semantic segmentation maps.

Furthermore, we have explored image style transfer techniques using Cycle-GANs and Diffusion models (InST method) for converting the style of the synthetic images to a realistic appearance. From the results, it was noted that Cycle-GAN focused on changing the colors of the pixels in the input images while maintaining the same semantic layout. On the other hand, the Diffusion model changed the elements appearing in the input image to their realistic counterparts. However, the semantic layout of the converted images changed with respect to the ground truth. The obtained preliminary results open the door for further research on closing the gap between the synthetic and real domains.

In the end, experiments were conducted in the maritime domain, trying to address the specific challenges related to this domain in depth estimation and semantic segmentation. A synthetic dataset was collected using UnrealEngine called \textit{MidSea}, and a synthetic-to-real assessment of Co-SemDepth and self-supervised approaches was carried out. 
The results obtained were good on the MidSea dataset as well as when transferring to the real SMD dataset. However, for the MIT marine dataset that contains various weather conditions, the results were not acceptable using Co-SemDepth trained from scratch. To improve the performance, MoCo and DINO self-supervised approaches were adopted to pretrain the encoder of the network, and image augmentations were increased during training. However, the results were still unsatisfactory after applying these techniques. Future work can be directed towards collecting more diverse synthetic data that covers many challenging marine conditions and use such data for training the network to enhance its synthetic-to-real generalization performance.


\begin{spacing}{0.9}


\bibliographystyle{ieeetr}
\cleardoublepage
\bibliography{References/references} 



\end{spacing}


\begin{appendices} 

\chapter{Image Warping} 
\label{chap:appendix1}

\begin{figure}%
    \centering
    {\includegraphics[width=10cm, height=6cm]{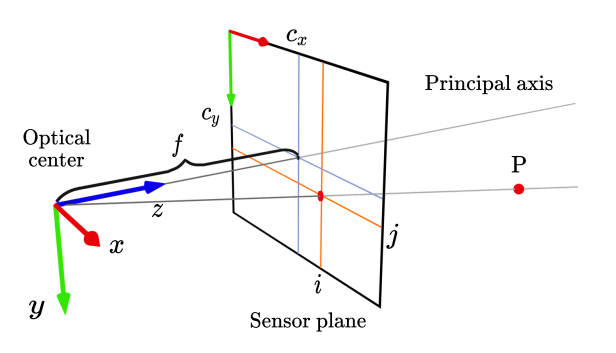} }
    \caption{A diagram of the pinhole camera model with axes and notations ~\cite{m4depth}}%
    \label{fig:camera_model}%
\end{figure}

\begin{figure}%
    \centering
    {\includegraphics[width=8cm, height=6cm]{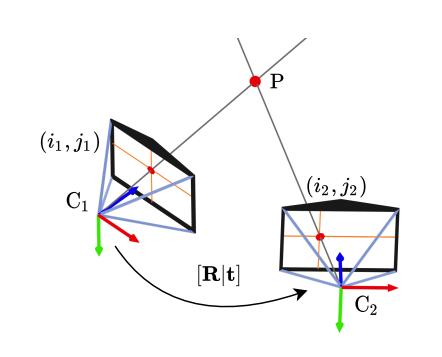} }
    \caption{Moving the camera from position C1 to C2~\cite{m4depth}}%
    \label{fig:im_conv}%
\end{figure}

Given the pinhole camera model illustrated in figure~\ref{fig:camera_model}, a camera is represented by a sensor plane and an optical center at a distance \(f\) (focal length) from the sensor plane. Such camera model can be characterized simply by 4 intrinsic parameters grouped in a matrix \(K\) called the projection matrix, as follows: 

\[K =  \begin{bmatrix} f_x & 0 & c_x\\ 0 & f_y & c_y\\ 0 & 0 & 1 \end{bmatrix}\]

A point P with coordinates \(\begin{bmatrix} x\\ y\\ z\end{bmatrix}\) in 3D space can be projected to its pixel coordinates \((i,j)\) on the sensor plane using the camera intrinsic matrix K as follows:

\begin{equation}
\begin{bmatrix} i\\ j\\ 1\end{bmatrix} = \frac{1}{z} K \begin{bmatrix} x\\ y\\ z\end{bmatrix}
\end{equation}

The previous relation can be written in homogeneous coordinates as follows:

\begin{equation}
\begin{bmatrix} z i\\ z j\\ z\\1 \end{bmatrix} = \begin{bmatrix}K & 0\\ 0 & 1\end{bmatrix} \begin{bmatrix} x\\ y\\ z\\ 1\end{bmatrix}
\end{equation}

When the camera moves from position C1 to C2 (figure ~\ref{fig:im_conv}), this movement can be expressed using a translation vector t and a rotation matrix R that are grouped in the \textit{transformation matrix} \(T\) as follows:

\[ ^1_2{T} = \begin{bmatrix} R & t\\ 0 & 1\end{bmatrix}\]

The 3D coordinates of point P expressed with respect to C1 can, then, be expressed with respect to C2 using the relation:

\begin{equation}
\begin{bmatrix} x_2\\ y_2\\ z_2\\ 1\end{bmatrix} = ^2_1{T} \begin{bmatrix} x_1\\ y_1\\ z_1\\1 \end{bmatrix}
\end{equation}

where: \[ ^2_1{T} = (^1_2{T})^{-1} \]

Given the predefined relations (equations 2 and 3), we can relate pixel coordinates \((i2, j2)\) to pixel coordinates \((i1, j1)\) using the following equation:

\begin{equation}
\begin{bmatrix} z_2 i_2\\ z_2 j_2\\ z_2\\ 1\end{bmatrix} = \begin{bmatrix}K & 0\\ 0 & 1\end{bmatrix} ^2_1{T} \begin{bmatrix}K^{-1} & 0\\ 0 & 1\end{bmatrix} \begin{bmatrix} z_1 i_1\\ z_1 j_1\\ z_1\\1 \end{bmatrix}
\end{equation}



\end{appendices}

\printthesisindex 

\end{document}